\documentclass{include/latex-class-tlp/new_tlp} 

\usepackage{times,helvet,courier}
\usepackage[utf8]{inputenc}
\usepackage{amsmath,amssymb}
\usepackage{microtype}
\usepackage{url}
\usepackage{graphicx}
\usepackage{multirow}
\usepackage{booktabs}
\usepackage{float}
\usepackage{hyperref}
\usepackage{graphicx}
\usepackage{wrapfig}
\usepackage{tikz}
\usepackage{enumerate}
\usetikzlibrary{positioning,arrows.meta}
\usepackage{listings}
\lstdefinestyle{appendixstyle}{xleftmargin=0pt}

\providecommand{\sysfont}{\textit}

\newcommand{\clingo}{\sysfont{clingo}}

\newcommand{\ginkgo}{\sysfont{ginkgo}}

\newcommand{\plasp}{\sysfont{plasp}}

\newcommand{\telingo}{\sysfont{telingo}}

\newcommand{\head}[1]{\ensuremath{\mathit{h}(#1)}} 

\newcommand{\body}[1]{\ensuremath{\mathit{B}(#1)}} 

\newcommand{\atom}[1]{\ensuremath{\mathit{A}(#1)}} 

\newcommand{\poslits}[1]{\ensuremath{{#1}^+}}
\newcommand{\neglits}[1]{\ensuremath{{#1}^-}}

\newcommand{\pbody}[1]{\poslits{\body{#1}}}
\newcommand{\nbody}[1]{\neglits{\body{#1}}}

\newcommand{\Tsign}{\ensuremath{\mathbf{T}}}
\newcommand{\Fsign}{\ensuremath{\mathbf{F}}}
\newcommand{\Vsign}{\ensuremath{\mathbf{V}}}
\newcommand{\Tlit}[1]{\ensuremath{\Tsign #1}}
\newcommand{\Flit}[1]{\ensuremath{\Fsign #1}}
\newcommand{\Vlit}[1]{\ensuremath{\Vsign #1}}
\newcommand{\Tass}[1]{\ensuremath{#1^{\Tsign}}}
\newcommand{\Fass}[1]{\ensuremath{#1^{\Fsign}}}

\newcommand{\clno}[1]{\ensuremath{\delta(#1)}}
\newcommand{\ClNo}[1]{\ensuremath{\Delta(#1)}}

\newcommand{\at}[2]{\ensuremath{#1[#2]}}
\newcommand{\prev}[1]{\ensuremath{#1'}}
\newcommand{\prevset}[1]{\ensuremath{#1'}}
\newcommand{\btw}[3]{\ensuremath{#1{[#2,#3]}}}
\newcommand{\atomat}[2]{\ensuremath{#1_#2}}
\newcommand{\tpb}[3]{\ensuremath{({#1,#2,#3})}}

\newcommand{\shift}[2]{\ensuremath{#1\langle#2\rangle}}

\newcommand{\cdnlasp}[0]{\ensuremath{\mathit{CDNL}\textnormal{-}\mathit{ASP}}}

\newcommand{\lbd}[0]{\emph{lbd}}
\newcommand{\graph}[1]{\ensuremath{\mathit{G}(#1)}}
\newcommand{\trans}[1]{\ensuremath{\mathit{trans}(#1)}}
\newcommand{\gen}[2]{\ensuremath{\mathit{gen}(#1,#2)}}
\newcommand{\normal}[1]{\ensuremath{#1^{n}}}
\newcommand{\choice}[1]{\ensuremath{#1^{c}}}
\newcommand{\integr}[1]{\ensuremath{#1^{i}}}
\newcommand{\integrprev}[1]{\ensuremath{#1^{\mathit{di}}}}

\newcommand{\normalbody}[1]{\ensuremath{\normal{\mathit{Bd}}(#1)}}
\newcommand{\pnf}[0]{PNF}

\newcommand{\ngic}[1]{\ensuremath{\mathit{ng}(#1)}}
\newcommand{\choiceprogram}[1]{\ensuremath{\mathit{Choice}(#1)}}
\newcommand{\tra}[0]{\ensuremath{\mathit{tr}^*}}
\newcommand{\trado}[1]{\ensuremath{\mathit{tr}^*(#1)}}

\newcommand{\traatom}[1]{\ensuremath{#1^*}}

\newcommand{\trb}[0]{\ensuremath{\mathit{tr}^{\lambda}}}
\newcommand{\trbdoprev}[1]{\ensuremath{\mathit{tr}^{\lambda}(#1)}}
\newcommand{\trbdo}[1]{\ensuremath{\lambda(#1)}}

\newcommand{\trbsymbol}[0]{\ensuremath{\lambda}}

\newcommand{\trbnormal}[0]{\trbsymbol{}-normal}

\newcommand{\stepsimple}[1]{\ensuremath{\mathit{step}(#1)}}

\newcommand{\simp}[1]{\ensuremath{\mathit{simp}(#1)}}
\newcommand{\steptrbsymbol}[1]{\ensuremath{\mathit{step}^{\trbsymbol}(#1)}}
\newcommand{\step}[1]{\steptrbsymbol{#1}} 

\newcommand{\assignfornogood}[1]{\ensuremath{\mathit{nogoods}(#1)}}

\newcommand{\nlit}[1]{\ensuremath{not\ #1}}

\newcommand{\cyclic}[0]{internal}
\newcommand{\stexttt}[1]{{\texttt{#1}}}

\newtheorem{theorem}{Theorem}
\newtheorem{proposition}{Proposition}

\newtheorem{lemma}{Lemma}
\newtheorem{example}{Example}

\renewcommand{\body}[1]{\ensuremath{\mathit{Bd}(#1)}} 
\renewcommand{\atom}[1]{\ensuremath{\mathit{At}(#1)}} 
\begin{document}

\title{\mbox{On the generalization of learned constraints} \mbox{for ASP solving in temporal domains}}
    
\author[J. Romero et al.]{%
JAVIER ROMERO  \and
TORSTEN SCHAUB  \and
KLAUS STRAUCH  \\
University of Potsdam, Germany
}

\submitted{[n/a]}
\revised{[n/a]}
\accepted{[n/a]}



\maketitle

\begin{abstract}
The representation of a temporal problem in ASP usually boils down to using copies of variables and constraints, one for each time stamp,
no matter whether it is directly encoded or expressed via an action or temporal language.
The multiplication of variables and constraints is commonly done during grounding and the solver is completely
ignorant about the temporal relationship among the different instances.
On the other hand,
a key factor in the performance of today's ASP solvers is conflict-driven constraint learning.
Our question in this paper is whether a constraint learned for particular time steps can be generalized and reused at
other time steps, and ultimately whether this enhances the overall solver performance on temporal problems.
Knowing well the domain of time,
we study conditions under which learned dynamic constraints can be generalized.
Notably, we identify a property of temporal representations
that enables the generalization of learned constraints across all time steps.
It turns out that most ASP planning encodings either satisfy this property 
or can be easily adapted to do so. 
%
In addition, we propose a general translation that transforms programs to fulfill this property.
Finally, 
we empirically evaluate the impact of \mbox{adding the generalized constraints to an ASP solver}.
Under consideration in Theory and Practice of Logic Programming (TPLP)
\begin{keywords} Answer Set Programming \and 
    Answer Set Solving \and 
    Temporal Reasoning
\end{keywords}

\end{abstract}


\section{Introduction}\label{sec:introduction}

Although Answer Set Programming (ASP; \cite{gellif88b}) experiences increasing popularity in academia and industry,
a closer look reveals that this concerns mostly static domains.
There is still quite a chasm between ASP's level of development for addressing static and dynamic domains.
This is because its modeling language as well as its solving machinery aim so far primarily at static knowledge,
while dynamic knowledge is mostly dealt with indirectly via reductions to the static case.
This also applies to dedicated dynamic formalisms like action and temporal languages~\cite{agcadipevi13a,gellif98a}.
In fact, their reduction to ASP or SAT usually relies on 
translations 
that introduce a
copy of each variable for each time step.
The actual dynamics of the problem is thus compiled out and a solver treats the result as any other static problem.

We address this by proposing a way to (partly) break the opaqueness of the actual dynamic problem and
equip an ASP solver with means for exploiting its temporal nature.
More precisely,
we introduce a method to strengthen the conflict-driven constraint learning framework (CDCL) of ASP solvers so that
dynamic constraints learned for specific time points can be generalized to other points in time.
These additional constraints 
can in principle reduce the search space and 
improve the performance of the ASP solvers.

We start in Section~\ref{sec:summary} showing our approach 
through an example. 
In Section~\ref{sec:background},
we review some background material. 
In Section~\ref{sec:temporal},
we introduce a
simple but general language to reason about time in ASP.
%
%
We define temporal problems, and 
characterize their solutions
in terms of completion and loop nogoods,
paralleling the approach to regular ASP solving~\cite{gekakasc12a}.
In Section~\ref{sec:approach}, 
using this language,
we study conditions under which learned constraints can be generalized to other time steps.
In Section~\ref{sec:all},
we identify a sufficient condition for the generalization of all learned constraints.
%
%
In Section~\ref{sec:cyclic:encoding}, 
we present a translation from temporal programs 
that generates programs satisfying that condition.
%
%
%
In Section~\ref{sec:experiments},
we empirically evaluate the impact of adding the 
generalized constraints to the ASP solver \clingo{}.
%
We conclude in Section~\ref{sec:conclusion}.
%

%

Our work can be seen as a continuation of the approach of \ginkgo~\cite{gekakalurosc16a},
which also aimed at generalizing temporal constraints but resorted to an external inductive proof method
(in ASP) for warranting correctness.
More generally, a lot of work has been conducted over recent years on lazy ASP solving~\cite{lebestga17a,padoporo09a,wetafr20a}.
Notably, conflict generalization was studied from a general perspective in~\cite{cowefr20a},
dealing with several variables over heterogeneous domains.
Lazy grounding via propagators was investigated in~\cite{cudorisc20a,marido22a,domari23a}.
Finally, it is worth mentioning that the usage of automata, as done in~\cite{cadihasc21a}, completely abolishes the use of time points.
A detailed formal and empirical comparative study of these approaches is interesting for future work.

This is an extended version of the conference paper~\cite{roscst22a} presented at RuleML+RR 2022.
The main new contribution is the identification in Section~\ref{sec:all} of 
a property of temporal problems that enables the generalization of learned constraints across all time points
without the need of any translation.
This significantly improves the applicability of our approach.
In fact, this property is satisfied by the planning domains that we 
considered in our empirical evaluation of~\cite{roscst22a}. 
Given this, we ran those experiments again, but this time using the original encodings,
only slightly modified to satisfy the mentioned property.
In Section~\ref{sec:cyclic:encoding} we also introduce a new translation to programs that satisfy that property. 
This translation is both more general and easier to understand than the one
presented in
our previous conference paper~\cite{roscst22a},
which can be found in~\ref{sec:translations}.
%
%
%
Finally, we have added in the~\ref{sec:proofs} the proofs of the theoretical results.


\section{An example}\label{sec:summary}

%
Our running example in this section is the \emph{Blocks World} problem, 
represented in the STRIPS subset of the PDDL planning language~\cite{mcdermott98a}.
We follow the approach of the \plasp{} system~\cite{digelurosc17a}, 
which translates a PDDL description into a set of facts, 
combines these facts with a meta-encoding implementing the PDDL semantics, 
and solves the resulting program using the ASP solver \clingo.%

Our instance consists of three blocks, \texttt{a}, \texttt{b} and \texttt{c}.
Initially, \texttt{a} is on top of \texttt{b}, that is on top of \texttt{c}, 
that is on the table. 
The goal is to rearrange the blocks in reverse order.
This is represented by the following set of facts:
\begin{lstlisting}
 block(a;b;c). init(clear(a)). init( handempty).
init(on(a,b)). init( on(b,c)). init(ontable(c)).
goal(on(b,a)). goal( on(c,b)). goal(ontable(a)).
\end{lstlisting}

There are four actions in the domain: 
\texttt{pick\_up(X)} picks up a block \texttt{X} that is on the table, 
\texttt{put\_down(X)} puts down a block \texttt{X} on the table, 
\texttt{stack(X,Y)} stacks the block \texttt{X} on top of block \texttt{Y}, and
\texttt{unstack(X,Y)} undoes that operation.
The following lines specify the action \texttt{stack(X,Y)} 
in terms of its \texttt{pre}conditions, 
\texttt{del}ete (or negative) effects 
and \texttt{add}itive (or positive) effects 
---where `\texttt{B}' is a shorthand for the body `\texttt{block(X), block(Y)}':
%
\begin{lstlisting}[basicstyle=\small\ttfamily]
       action(stack(X,Y)) :- B.  pre(stack(X,Y),holding(X)) :- B.
add(stack(X,Y), clear(X)) :- B.  pre(stack(X,Y),  clear(Y)) :- B.
add(stack(X,Y),handempty) :- B.  del(stack(X,Y),holding(X)) :- B.
add(stack(X,Y),  on(X,Y)) :- B.  del(stack(X,Y),  clear(Y)) :- B.
\end{lstlisting}
The specification of the rest of the actions is available in~\ref{appendix:blocks:action:description}.

The following meta-encoding gives meaning to the previous rules:
\begin{lstlisting}[numbers=left]
holds(F,0) :- init(F).
{ occ(A,T) : action(A) } = 1 :- T=1..n.    
:- occ(A,T), pre(A,F), not holds(F,T-1).
holds(F,T) :- occ(A,T), add(A,F).
holds(F,T) :- holds(F,T-1), T=1..n, 
              not occ(A,T) : del(A,F).
:- goal(F), not holds(F,n). 
\end{lstlisting}
Line~1 specifies which fluents \texttt{F} hold at time step \texttt{0}, 
Line~2 generates one action \texttt{A} per time step \texttt{T} between
\texttt{1} and some constant \texttt{n}, 
Line~3 forbids the occurrence of an action \texttt{A}
  if some of its preconditions do not hold,  
Line~4 defines the positive effects of an action \texttt{A}, 
Lines~5-6 define inertia for all fluents \texttt{F}, and 
Line~7 enforces that every goal fluent \texttt{F} holds at the last time point \texttt{n}.

The shortest plan for this problem requires six actions, 
one to pick-up or unstack every block, 
and another to put it down or stack it. 
Accordingly, 
when the constant \texttt{n} has the value \texttt{6}, 
our program has a unique stable model that includes the atoms
\texttt{occ(unstack(a,b),1)},  
\texttt{occ(put\_down(a),2)}, 
\texttt{occ(unstack(b,c),3)}, 
\texttt{occ(stack(b,a),4)}, 
\texttt{occ(pick\_}\newline\texttt{up(c),5)}, and 
\texttt{occ(stack(c,b),6)}. 

Such a stable model can be computed by the ASP solver \clingo, 
that is based in the \emph{ground-and-solve} approach to ASP solving.
In the first step, \clingo\ grounds the logic program 
by replacing the rules with variables with their ground (variable-free) instantiations.
The goal of this step is to generate a ground program that is 
as compact as possible, 
while preserving the stable models of the original program. 
%
%
%
%
As an example, 
consider the precondition rule in Line~3 of the meta-encoding 
for action \texttt{stack(b,a)}. 
At time point \texttt{2}, \clingo{}
generates two ground instances,
one for each precondition of the action: 
\begin{lstlisting}
:- occ(stack(b,a),2), not holds(  clear(a),1).
:- occ(stack(b,a),2), not holds(holding(b),1).
\end{lstlisting}
At time point \texttt{1}, however,
the corresponding first instance is not generated, 
as \clingo{} infers that \texttt{holds(clear(a),0)} must be true
due to the fact \texttt{init(clear(a))} 
and the rule in Line~1.
In turn, the corresponding second instance is simplified to 
\begin{lstlisting}
:- occ(stack(b,a),1).
\end{lstlisting}
since \clingo{} deduces that \texttt{holds(holding(b),0)} cannot be true, 
as no rule in the program can derive it. 

In the second step, \clingo\ uses a Conflict-Driven Nogood Learning (CDNL) 
algorithm to search for stable models of the ground program generated before. 
Whenever the algorithm backtracks, 
it learns a new constraint, also called a \emph{nogood}, 
that is satisfied by all stable models.
These learned constraints are crucial for the performance of the solver, 
as they prevent it from making the same mistakes more than once.
We explain this in more detail in Section~\ref{sec:background}.
As an example, while solving our program, \clingo\ could learn the constraint
\begin{align}\label{eq:ic:one}
\stexttt{:- holds(on(a,b),2), holds(on(b,c),2), not holds(on(b,c),4).}
\end{align}
representing that if at time point \texttt{2} 
block \texttt{a} is on top of block \texttt{b}, and \texttt{b} is on top of block \texttt{c}, 
then at time point \texttt{4} block \texttt{b} must remain on top of \texttt{c}.
Note that although 
the constraint is specific to time points \texttt{2} and \texttt{4}, 
the solver learned it using rules that are replicated at every time point. 
Then, the main question of this paper is:
\begin{itemize}
\item[] 
\emph{Can we generalize such learned constraints to other time points?} 
\end{itemize}
And if so:
\begin{itemize}
\item[] 
\emph{What impact does this have on the performance of ASP solvers?} 
\end{itemize}
We can generalize~\eqref{eq:ic:one} 
to all time steps in the interval \texttt{0..n}
with the following constraint:
\begin{align}\label{eq:genic}
\stexttt{
:- }& \stexttt{holds(on(a,b),T-2), holds(on(b,c),T-2),
} \\ \nonumber
& \stexttt{not holds(on(b,c),T), T=2..n.}
\end{align}
%
This constraint can be safely added to our program, 
since it does not eliminate any of its stable models. 
Ideally, the solver could detect this automatically and learn
this generalized constraint directly instead of~\eqref{eq:ic:one}.
This could help it to prune the search space and find solutions faster.
However, such generalization is not always easy, for two main reasons.

The first reason is that, as we have seen, 
\clingo{} does not replicate \emph{exactly} the same ground rules at every time point.
Since the constraints learned by the solver depend on these ground rules, 
a constraint learned at one time step may not apply to others.
For example, 
the solver can learn 
\begin{align}\label{eq:ic:two}
\stexttt{:- not holds(on(b,c),2).}
\end{align}
which corresponds to constraint~\eqref{eq:ic:one} for step \texttt{2},
without the atoms 
\texttt{holds(on(a,b),0)} 
and 
\texttt{holds(on(b,c),0)}.
During grounding, \clingo{} 
infers that these atoms are true
and simplifies the ground rules where they appear accordingly.
As a result, the constraints learned from these simplified rules are also simpler
---compare~\eqref{eq:ic:two} with~\eqref{eq:ic:one}.
Since these simplifications are not correct for all time points, 
the generalization of~\eqref{eq:ic:two} to all time points
\begin{lstlisting}
:- not holds(on(b,c),T), T=0..n.
\end{lstlisting}
is also not correct. 
%
%
In fact, it would eliminate all stable models. 

We address this issue manually, 
modifying the meta-encoding in such a way that
the ground rules are truly the same for all time steps.
We achieve this by replacing the rule in Line~1 by the choice rules 
\begin{lstlisting}
{ holds(F,0) } :-  init(F).
{ holds(F,0) } :- add(A,F).
\end{lstlisting}
that generate all possible initial states, and 
by eliminating the integrity constraint of Line~8. 
Once this is done, the ground instantiations at all time steps become the same.
For example, the ground instantiation of the rule 
in Line~3 for action \texttt{stack(b,a)} at time point \texttt{1} becomes
\begin{lstlisting}
:- occ(stack(b,a),1), not holds(  clear(a),0).
:- occ(stack(b,a),1), not holds(holding(b),0).
\end{lstlisting}
and instead of learning~\eqref{eq:ic:two}, 
\clingo{} would learn the constraint
\begin{align}\label{eq:ic:tri}
\stexttt{:- holds(on(a,b),0), holds(on(b,c),0), not holds(on(b,c),2).}
\end{align}

that can be safely generalized to~\eqref{eq:genic}.
Observe that the modified program no longer determines
the values of the fluents at the initial and final situations.
Instead, these values are specified by passing to \clingo{}
an additional set of so-called \emph{assumptions}.
%
%
Notably, 
these assumptions preserve the stable models of the original program
but do not interfere with the grounding process.
Hence, the ground rules remain the same across all time steps.
%
%
%
%

We formalize this approach 
in Section~\ref{sec:temporal} 
using 
\emph{temporal logic problems}, 
which consist of
\emph{temporal logic programs}
defining the ground rules that are replicated at each time step, 
and partial assignments representing the initial and final states.
Additionally, we study the \emph{temporal nogoods} associated with temporal logic programs.
Solving with assumptions is introduced in Section~\ref{sec:background}, 
while its application to temporal logic problems 
is described in Section~\ref{sec:approach}.

To illustrate the second issue, 
let us extend our example by introducing a counter
that is incremented with each action.
We modify Line~4 of the meta-encoding
to represent conditional effects of actions as follows:
%
\begin{align*}
\stexttt{holds(F,T) :- occ(A,T), add(A,F), holds(G,T-1) : cond(A,F,G).}
\end{align*}
The atom \texttt{cond(A,F,G)} expresses that for action \texttt{A} to produce effect \texttt{F}, 
the fluent \texttt{G} must have been true in the previous state. 
The counter can be represented as a conditional effect of all actions:
\begin{lstlisting}
              add(A,counter(T)) :- action(A), T=1..m.
cond(A,counter(T),counter(T-1)) :- action(A), T=2..m.
\end{lstlisting}
Actions make \texttt{counter(1)} true, 
and for \texttt{t} between \texttt{2} and \texttt{m},
they also make \texttt{counter(t)} true
if \texttt{counter(t-1)} was true in the previous state.
%
%
%
Given this extension, 
\clingo{} can learn the constraint
\begin{align}\label{eq:ic:for}
\stexttt{:- not holds(counter(2),2).}
\end{align}
as \texttt{counter(2)} always holds at time point \texttt{2} 
in all stable models.
However, the generalization to all time points 
\begin{align}\label{eq:ic:fiv}
\stexttt{:- not holds(counter(2),T), T=0..n.}
\end{align}
would be incorrect, 
as \texttt{counter(2)} does not have to hold
at time points \texttt{0} or \texttt{1}. 
This example shows that, 
even when all rules are replicated across all time steps, 
we cannot always generalize learned constraints to all time steps.
This raises the question: 
when can we generalize learned constraints to other ---or all--- 
time steps?

%

We provide a first answer in Section~\ref{sec:approach},
after formalizing the problem of generalizing learned constraints. 
Assume that 
the time steps of the rules used to learn a constraint
are known. 
%
This information allows us to to determine 
the time steps to which a learned constraint can we generalized.
For instance, 
if constraint~\eqref{eq:ic:for} 
---about fluent \texttt{counter(2)} at step \texttt{2}--- was learned 
using rules from time steps \texttt{1} and \texttt{2}, 
we could generalize it to the interval
\texttt{T=2..n}, 
because the rules used at \texttt{1} and \texttt{2}
have corresponding copies at 
\texttt{T-1} and \texttt{T}: 
\begin{align}\label{eq:ic:clocktwon}
\stexttt{:- not holds(counter(2),T), T=2..n.}
\end{align}
%
However, we could not generalize the constraint to \texttt{T=1}, 
as no copies of the rules exist for time step \texttt{0}.
%
%
%
As another example,
if~\eqref{eq:ic:for} was
learned using rules of steps \texttt{2} and \texttt{3}, 
then the generalized constraint would 
apply to the interval \texttt{T=1..n-1}, 
as the rules at \texttt{2} and \texttt{3} 
would have their corresponding copies at \texttt{T} and \texttt{T+1}, 
and there are no copies at \texttt{n+1}.
%
Clearly, a drawback of this approach is that 
it requires tracking the time steps of the rules used
to learn each constraint. 
Consequently, 
its implementation would require the modification of the inner machinery 
of~\clingo{}.


To avoid this,
in Section~\ref{sec:all}
we investigate under which conditions 
can we generalize learned constraints to all time steps.
In our previous examples, 
we could not generalize~\eqref{eq:ic:for} to~\eqref{eq:ic:fiv}
because there were 
no copies of the rules at step \texttt{0} or step \texttt{n+1}.
%
%
But what would happen if 
the program behaved \emph{as if} those copies existed?
It turns out that in such a case 
we could generalize the learned nogoods to all time steps. 
How can we capture these cases?
We define for each temporal program a \emph{transition graph}
whose edges correspond to the transitions between states. 
If every initial state has a predecessor in that graph, 
it is \emph{as if} the rule copies applied also to those initial states.
Likewise, if every state reachable from an initial state 
has a successor, 
it is \emph{as if} the rule copies applied to those final states as well. 
%
In Section~\ref{sec:all} we define temporal programs as \emph{\cyclic{}}
if their transition graph satisfies these properties, 
and we show that for \cyclic{} programs 
all learned nogoods can be generalized to \emph{all} time steps.

In practice, most planning representations
that we have encountered are either \cyclic{} or 
can be easily adapted to become \cyclic.
For instance, to adapt our meta-encoding, 
it is sufficient to modify Line~2 by replacing the 
comparison operator \texttt{=} by \texttt{<=},
allowing the non-execution of actions at any time step.
The inertia rule in Lines~5-6 ensures that fluents persist 
whenever no action occurs.
Thus, 
initial states, where no action occurs, have themselves as predecessors.
Similarly, 
reachable states have a successor state 
where no action occurs and all fluents persist.
This small adjustment makes the representation \cyclic.
Once it is applied, 
constraint~\eqref{eq:ic:for} can no longer be learned, 
as it does not hold in all stable models, 
i.e., there are stable models where no action occurs at 
time steps \texttt{1} or \texttt{2},
and \texttt{counter(2)} does not hold at \texttt{2}.
Instead, the solver can learn the following constraint:
\begin{lstlisting}
:- not holds(counter(2),2), occ(unstack(a,b),1),
   occ(putdown(a),2).
\end{lstlisting}
which can be safely generalized to 
\begin{lstlisting}
:- not holds(counter(2),T), occ(unstack(a,b),T-1),
   occ(putdown(a),T), T=1..n.
\end{lstlisting}
%
We consider this constraint, with \texttt{T=1..n}, 
a generalization across all time steps because 
its ground instances include atoms over all steps.
In this case, 
extending the generalization to additional steps 
would be pointless. 
For other constraints, it could even be incorrect 
due to the semantics of negation.

%
%
 
To handle cases where a temporal representation is not \cyclic, 
in Section~\ref{sec:cyclic:encoding}
we introduce a translation that transforms any temporal program into an \cyclic{} one.
In our example, the translation adds the following rules:
\begin{lstlisting}
{  lambda(T) } :- T=1..n.
{ holds(F,T) } :-   init(F), T=1..n, not lambda(T).
{ holds(F,T) } :-  add(A,F), T=1..n, not lambda(T).
{   occ(A,T) } :- action(A), T=1..n, not lambda(T).
\end{lstlisting}
Additionally, 
the translation inserts the atom \texttt{lambda(T)}
into the body of each rule in the meta-encoding, 
and assumes that \texttt{lambda} is false at time step \texttt{0}.
This translation ensures that initial states ---where \texttt{lambda} is false---
have themselves as predecessors, 
and reachable states have themselves 
---after erasing \texttt{lambda}---
as successors.
Once this translation is applied, 
\clingo{}, instead of learning constraint~\eqref{eq:ic:for},
could learn
\begin{align*}
\stexttt{:- not holds(counter(2),2), lambda(2), lambda(1).}
\end{align*}
that could be generalized to 
\begin{align}\label{eq:ic:eig}
\stexttt{:- not holds(counter(2),T), lambda(T), lambda(T-1), T=1..n.}
\end{align}
Furthermore, the constraints learned using the translation 
can be applied to the original program, 
after simplifying away the atoms over \texttt{lambda} 
and adapting the interval of \texttt{T} to \texttt{T=2..n}.
In our example,~\eqref{eq:ic:eig} can be simplified to~\eqref{eq:ic:clocktwon}, 
that can be safely added to our original program.

\section{Background}\label{sec:background}

We review the material from~\cite{gekanesc07a}
about solving normal logic programs, 
and adapt it for our purposes to cover 
normal logic programs with choice rules and integrity constraints
over some set $\mathcal{A}$ of atoms.

A \emph{rule} $r$ has the form $H \leftarrow B$
where $B$ is a set of literals over $\mathcal{A}$,
and $H$ is  either 
an atom $a \in \mathcal{A}$, and we call $r$ a \emph{normal rule}, or
$\{a\}$ for some atom $a \in \mathcal{A}$, 
making $r$ a \emph{choice rule}, or
$\bot$, so that $r$ is an \emph{integrity constraint}.
We usually drop braces from rule bodies $B$, and 
drop the arrow $\leftarrow$ when $B$ is empty.
We use the extended choice rule 
$\{a_1; \ldots; a_n\}\leftarrow B$ as a shorthand 
for the choice rules
$\{a_1\}\leftarrow B$, 
\ldots,
$\{a_n\}\leftarrow B$, 
and for some set of atoms $X \subseteq \mathcal{A}$ 
we denote 
by $\choiceprogram{X}$
the set of choice rules $\{ \{ a \} \leftarrow \mid a \in X \}$.
%
%
%
%
A \emph{program} $\Pi$ is a set of rules.
By \normal{\Pi}, \choice{\Pi}, and \integr{\Pi} 
we denote its normal rules, choice rules and integrity constraints,
respectively.
Semantically, 
a logic program induces a collection of \emph{stable models}, 
which are distinguished models of the program determined by the
stable models semantics (see~\cite{gekakasc12a,gellif88b} for details).

%

%
%
For a rule $r$ of the form $H \leftarrow B$, 
let $\head{r}=a$ be the \emph{head} of~$r$ if $H$
has the form $a$ or $\{a\}$ for some atom $a \in \mathcal{A}$, 
and let $\head{r}=\bot$ otherwise.
Let
\(
\body{r}=B
\)
be the \emph{body} of $r$,
\(
\pbody{r}=\{a \mid a \in \mathcal{A}, a \in B\}
\)
be the \emph{positive body} of $r$, and
\(
\nbody{r}=\{a \mid a \in \mathcal{A}, \neg a \in B\}
\)
be the \emph{negative body} of $r$.
The set of atoms occurring in a rule $r$ and in a logic program $\Pi$
are denoted by $\atom{r}$ and $\atom{\Pi}$, respectively.
The set of bodies in $\Pi$ is $\body{\Pi} = \{\body{r}\mid r\in\Pi\}$.
For regrouping rule bodies sharing the same head $a$,
we define
\(
\body{a}=\{\body{r}\mid r\in\Pi,\head{r}=a\}
\), 
and by \normalbody{a} we denote the restriction of that set to bodies of normal rules,
i.e., 
\(
\{\body{r}\mid r\in\normal{\Pi},\head{r}=a\}
\).


A Boolean \emph{assignment} $S$ over a set $\mathcal{D}$, 
called the domain of $S$,
is a set $\{\sigma_1,\ldots,\sigma_n\}$ of \emph{signed literals}
$\sigma_i$ of the form $\Tlit{a}$ or $\Flit{a}$ 
for some $a\in\mathcal{D}$ and $1 \leq i \leq n$;
%
$\Tlit{a}$ expresses that $a$ is \emph{true} and $\Flit{a}$ that it is \emph{false}.
We omit the attribute \emph{signed} for literals whenever clear from the context.
We denote the complement of a literal $\sigma$ by $\overline{\sigma}$, 
that is,
$\overline{\Tlit{a}} = \Flit{a}$ and $\overline{\Flit{a}} = \Tlit{a}$.
%
%
%
Given this, 
we access true and false propositions in $S$ via
\(
\Tass{S}
=
\{a \in \mathcal{D} \mid \Tlit{a} \in S\}
\)
and
\(
\Fass{S}
=
\{a \in \mathcal{D} \mid \Flit{a} \in S\}
\).
We say that a set of atoms $X$ is consistent with an assignment $S$ if
$\Tass{S}\subseteq X$ and $\Fass{S}\cap X = \emptyset$.
%
%
%
In our setting, a \emph{nogood} is a set $\{\sigma_1,\ldots,\sigma_n\}$ of
signed literals,
expressing a constraint violated by any assignment containing
$\sigma_1,\ldots,\sigma_n$.
%
%
%
Accordingly, 
the nogood for a body $B$, 
denoted by \ngic{B},
is $\{\Tlit a \mid a \in B^{+}\}\cup\{\Flit a \mid a \in B^{-}\}$.
%
We say that an assignment $S$ over $\mathcal{D}$ is \emph{total} if 
$\Tass{S} \cup \Fass{S} = \mathcal{D}$ 
and
$\Tass{S} \cap \Fass{S} = \emptyset$. 
A total assignment $S$ over $\mathcal{D}$
is a \emph{solution} for a set~$\Delta$ of nogoods,
if
$\delta\not\subseteq S$ for all $\delta\in\Delta$.
A set~$\Delta$ of nogoods \emph{entails} a nogood $\delta$
if $\delta\not\subseteq S$ for all solutions~$S$ over $\mathcal{D}$ for~$\Delta$, 
and it entails a set of nogoods $\nabla$ 
if it entails every nogood $\delta \in \nabla$ in the set.

We say that a nogood $\delta$ is a \emph{resolvent} of 
a set of nogoods $\Delta$ if there is a sequence of nogoods
$\delta_1, \ldots, \delta_n$ with $n \geq 1$ 
such that $\delta_n=\delta$, 
and for all $i$ such that $1 \leq i \leq n$, 
either $\delta_i \in \Delta$, or 
there are some $\delta_j$, $\delta_k$ with $1\leq j < k < i$ 
such that
$\delta_i=(\delta_j\setminus\{\sigma\})\cup(\delta_k\setminus\{\overline{\sigma}\})$
for some signed literal $\sigma$.
In this case, we say that the sequence $\delta_1, \ldots, \delta_n$
is a proof of $\delta_n$.
%
We say that a signed literal $\sigma$ is unit resulting for a nogood $\delta$ 
and an assignment $S$ if $\delta \setminus S = \{\sigma\}$
and $\overline{\sigma}\notin S$.
For a set of nogoods $\Delta$ and an assignment $S$, 
unit propagation is the process of extending $S$ with 
unit-resulting literals until no further literal 
is unit resulting for any nogood in $\Delta$.

Inferences in ASP can be expressed in terms of atoms and rule bodies.
%
%
We begin with nogoods capturing inferences from the Clark completion.
%
%
For a body $B=\{a_1,\dots,a_m,\neg a_{m+1},\dots,\neg a_n\}$,
we have that 
$
\clno{B} \!
= \!
\{\Flit{B},\Tlit{a_1},\dots,\Tlit{a_m},\Flit{a_{m+1}},\dots,$ $\Flit{a_n}\}
$
and 
$
  \ClNo{B}
  =
  \{ \, \{\Tlit{B},\Flit{a_1}\},
        \dots, $ $ 
      \{\Tlit{B},\Flit{a_m}\},
      \{\Tlit{B},\Tlit{a_{m+1}}\},
      \dots,
      \{\Tlit{B},$ $\Tlit{a_n}\} \, \}.
$
%
%
%
For an atom $a$ such that
$\normalbody{a}=\{B_1,\dots,$ $B_k\}$,
we have that 
$
\ClNo{a} = \{\,\{\Flit{a},\Tlit{B_1}\},\dots,$ $ \{\Flit{a},\Tlit{B_k}\}\,\},
$
and if $\body{a}=\{B_1,\dots,B_k\}$ then 
$
\clno{a} = \{\Tlit{a},\Flit{B_1},\dots,\Flit{B_k}\}.
$
%
%
Given this, the \emph{completion nogoods} of a logic program $\Pi$ are defined as follows:
%
\[
\label{eq:compnogoods}
\begin{array}{@{}l@{}c@{}l}
\Delta_\Pi & {} = {} &
\{\clno{B} \mid B\in\body{{\Pi}\setminus\integr{\Pi}}\} 
\cup
\{\delta\in\ClNo{B}\mid B\in\body{{\Pi}\setminus{\integr{\Pi}}}\} 
\\[2pt]
& \cup &
\{\clno{a} \mid a\in\atom{\Pi}\}
\cup
\{\delta\in\ClNo{a} \mid a\in\atom{\Pi}\}
\\[2pt]
& \cup &
\{\ngic{B} \mid B\in\body{\integr{\Pi}}\}
\end{array}
\]
Choice rules of the form $\{a\} \leftarrow B$ 
are considered by not adding the corresponding nogood $\{\Flit a, \Tlit B\}$ to $\ClNo{a}$, 
and integrity constraints from $\integr{\Pi}$ 
of the form $\bot \leftarrow B$
are considered by adding directly their corresponding nogood \ngic{B}.
%
%
The definition of the 
\emph{loop nogoods} $\Lambda_\Pi$,
capturing the inferences from loop formulas, 
is the same as in~\cite{gekanesc07a}.
We do not specify them here since they
do not pose any special challenge to our approach, 
and they are not needed in our (tight) examples.

%
%
%

To simplify the presentation,
we slightly deviate from~\cite{gekanesc07a}
and consider a version of the nogoods of a logic program 
where the occurrences of the empty body are simplified.
Note that 
$\clno{\emptyset}=\{\Flit\emptyset\}$ 
and 
$\ClNo{\emptyset}=\emptyset$.
Hence, if $\emptyset \in \body{\Pi}$ then any solution to 
the completion and loop nogoods of $\Pi$ must contain $\Tlit\emptyset$.
Based on this, we can delete from $\Delta_\Pi \cup \Lambda_\Pi$ 
the nogoods that contain $\Flit\emptyset$,
and eliminate the occurrences of $\Tlit\emptyset$ from the others.
Formally, we define the set of (simplified) nogoods for $\Pi$ 
as:
\[
\Sigma_\Pi = \{\delta\setminus\{\Tlit\emptyset\} \mid 
                \delta \in \Delta_\Pi \cup \Lambda_\Pi,
                \Flit\emptyset \notin \delta\}.
\]
To accommodate this change, for a program~$\Pi$,
we fix the domain $\mathcal{D}$ of the assignments to
the set $\atom{\Pi}\cup(\body{\Pi}\setminus\emptyset)$.
Given this, the stable models of a logic program $\Pi$ 
can be characterized by the nogoods $\Sigma_{\Pi}$ for that program.
This is made precise by the following theorem,
which is an adaptation of Theorem 3.4 from~\cite{gekanesc07a} to our setting.
%
\begin{theorem}\label{thm:nogoods:nontight}
Let $\Pi$ be a logic program.
Then,
$X\!\!\subseteq\!\!\atom{\Pi}$ is a stable model of~$\Pi$
iff
$X\!=\!\Tass{S}\cap\atom{\Pi}$ for a (unique) 
solution $S$ for $\Sigma_\Pi$.
\end{theorem}

To compute the stable models of a logic program $\Pi$, 
we apply the algorithm $\cdnlasp{}(\Pi)$ from~\cite{gekanesc07a}
implemented in the ASP solver \clingo.
The algorithm 
searches for a solution $S$ to the set of nogoods $\Sigma_\Pi$, 
and when it finds one it returns the corresponding set of atoms
$\Tass{S}\cap\atom{\Pi}$.
\cdnlasp{} maintains a current assignment $S$ and a current set of learned nogoods 
$\nabla$, both initially empty.
The main loop of the algorithm starts by applying unit propagation to $\Sigma_{\Pi}\cup\nabla$, 
possibly extending $S$.
Every derived literal is ``implied'' by some nogood 
$\delta \in \Sigma_{\Pi}\cup\nabla$, 
which is stored in association with the derived literal.
This derivation may lead to the violation of another nogood.
This situation is called \emph{conflict}.
If propagation finishes without conflict,
then a (heuristically chosen) literal can be added to $S$, 
provided that $S$ is partial,
while otherwise $S$ represents a {solution} and can be directly returned. 
%
On the other hand, if there is a conflict, there are two possibilities.
Either it is a top-level conflict, 
independent of heuristically chosen literals, 
in which case the algorithm returns \emph{unsatisfiable}.
Or, if that is not the case,
the conflict is analyzed to calculate a conflict nogood $\delta$, that is added to $\nabla$.
More in detail, $\delta$ is a resolvent of the set of nogoods associated 
with the literals derived after the last heuristic choice.
Hence, every learned nogood $\delta$ added to $\nabla$ 
is a resolvent of $\Sigma_{\Pi}\cup\nabla$ and, by induction, 
it is also a resolvent of $\Sigma_{\Pi}$.
After recording $\delta$, 
the algorithm backjumps to the earliest stage where 
the complement of some formerly assigned literal is implied by $\delta$,
thus triggering propagation and starting the loop again.

This algorithm has been extended for solving under assumptions \cite{eensor03b}.
In this setting, 
the procedure $\cdnlasp{}(\Pi,S)$
receives additionally as input a partial assignment $S$ over \atom{\Pi},
the so-called assumptions,
and returns some stable model of $\Pi$ that is consistent with $S$.
To accommodate this extension, 
the algorithm simply decides first on the literals from $S$, and
returns \emph{unsatisfiable} as soon as any of these literals is undone by backjumping.
No more changes are needed.
Notably, the learned nogoods are still resolvents of $\Delta_{\Pi}$,
that are independent of the set of assumptions $S$.

\section{Temporal programs, problems and nogoods}\label{sec:temporal}
We introduce a simple language of temporal logic programs
to represent temporal problems.
%
These programs 
represent the dynamics of a temporal domain 
by referring to two time steps: the current step and the previous step.
We refer to the former by atoms from a given set $\mathcal{A}$, 
and to the latter by atoms from the set 
$\prevset{\mathcal{A}}=\{ \prev{a} \mid a \in \mathcal{A}\}$, 
that we assume to be disjoint from $\mathcal{A}$.
We define a \emph{state} as a subset of $\mathcal{A}$. 
Following the common-sense flow of time, 
normal or choice rules define the atoms of the current step
in terms of the atoms of both the current and the previous step.
Integrity constraints forbid some states,
possibly depending on their previous state. 
%
Syntactically, a temporal logic program $\Pi$ over $\mathcal{A}$
has the form of a 
(non-temporal) logic program over $\mathcal{A}\cup\prevset{\mathcal{A}}$
such that for every rule $r \in \Pi$,
if $r \in \normal{\Pi}\cup\choice{\Pi}$ then $\head{r}\in\mathcal{A}$, 
and otherwise $(\pbody{r}\cup\nbody{r})\cap\mathcal{A}\neq\emptyset$.
Given that temporal logic programs over $\mathcal{A}$ 
can also be seen as (non-temporal) logic programs over $\mathcal{A}\cup\prevset{\mathcal{A}}$,
in what follows we may apply the notation of the latter to the former.
%

This language is designed to capture the core 
of the translations to ASP of action and temporal 
languages.
For instance, in our introductory PDDL example~\cite{mcdermott98a}, 
a temporal logic program can represent the transition
between \texttt{T-1} and \texttt{T} defined by the rules in Lines~2-6 of the meta-encoding.
In essence, this temporal program 
corresponds to the ground instantiation of 
those rules at a single time point.
Temporal programs are also closely related to the present-centered 
programs used in the implementation of the temporal ASP solver \telingo~\cite{cakamosc19a}, 
or to the programs that define the transitions in action languages~\cite{leliya13a}.


\newpage
\begin{example}
\label{example:one}
Our running example is the temporal logic program $\Pi_1$ over $\mathcal{A}_1=\{a,b,c,$ $d\}$
that consists only of choice rules and integrity constraints:
\begin{equation*}
\begin{array}{l@{\hspace{4mm}}r@{\ }c@{\ }l@{\hspace{3cm}}l@{\hspace{4mm}}r@{\ }c@{\ }l}
 & \{ a; b; c; d \} & \leftarrow &
&
 & \bot & \leftarrow & a', \neg b
\\
 & \bot & \leftarrow & \neg b', b
&
 & \bot & \leftarrow & \neg c', a
\\
 & \bot & \leftarrow & d', b
&
 & \bot & \leftarrow & c, \neg d
\\
 & \bot & \leftarrow & \neg a', \neg c
&
 & \bot & \leftarrow & \neg a', c', \neg a
\end{array}\end{equation*}%
\end{example}


%
%

Temporal logic programs $\Pi$ can be instantiated to specific time intervals.
We introduce some notation for that.
Let $m$ and $n$ be integers such that $1 \leq m \leq n$,
and $[m,n]$ denote the set of integers
$\{ i \mid m \leq i \leq n\}$.
%
For $a \in \mathcal{A}$, 
the symbol \at{a}{m} denotes the atom \atomat{a}{m}, and
for $\prev{a} \in \mathcal{A}'$, 
the symbol \at{\prev{a}}{m} denotes the atom \atomat{a}{{m-1}}.
For a set of atoms $X\subseteq\mathcal{A}\cup\prevset{\mathcal{A}}$, 
\at{X}{m} denotes the set of atoms $\{ \at{a}{m} \mid a \in X\}$, and 
\btw{X}{m}{n} denotes the set of atoms $\{\at{a}{i}\mid p \in X, i \in [m,n]\}$.
%
For a rule $r$  over $\mathcal{A}\cup\mathcal{A}'$, 
the symbol \at{r}{m}
denotes the rule that results from replacing
in $r$ every atom $a \in \mathcal{A}\cup\mathcal{A}'$ by $\at{a}{m}$,
and \btw{r}{m}{n} denotes the set of rules
$\{\at{r}{i} \mid i \in [m,n]\}$.
Finally, 
for a temporal program $\Pi$, 
\at{\Pi}{m} is $\{ \at{r}{m} \mid r \in \Pi\}$, and
\btw{\Pi}{m}{n} is 
$\{ \at{\Pi}{i} \mid i \in [m,n] \}$.

\begin{example}
The instantiation of $\Pi_1$ at $1$, denoted by $\at{\Pi_1}{1}$, is:
\begin{equation*}
\begin{array}{l@{\hspace{4mm}}r@{\ }c@{\ }l@{\hspace{3cm}}l@{\hspace{4mm}}r@{\ }c@{\ }l}
 & \{ \atomat{a}{1}; \atomat{b}{1}; \atomat{c}{1}; \atomat{d}{1} \} & \leftarrow &
&
 & \bot & \leftarrow & \atomat{a}{0}, \neg \atomat{b}{1}
\\
 & \bot & \leftarrow & \neg \atomat{b}{0}, \atomat{b}{1}
&
 & \bot & \leftarrow & \neg \atomat{c}{0}, \atomat{a}{1}
\\
 & \bot & \leftarrow & \atomat{d}{0}, \atomat{b}{1}
&
 & \bot & \leftarrow & \atomat{c}{1}, \neg \atomat{d}{1}
\\
 & \bot & \leftarrow & \neg \atomat{a}{0}, \neg \atomat{c}{1}
&
 & \bot & \leftarrow & \neg \atomat{a}{0}, \atomat{c}{0}, \neg \atomat{a}{1}
\end{array}\end{equation*}%
The programs $\at{\Pi_1}{i}$ for $i \in \{2,3,4\}$ are the same,
except that the subindex $1$ is replaced by $i$, 
and the subindex $0$ is replaced by $i-1$.
The instantiation of $\Pi_1$ at $[1,4]$, denoted by \btw{\Pi_1}{1}{4},
is $\at{\Pi_1}{1}\cup\at{\Pi_1}{2}\cup\at{\Pi_1}{3}\cup\at{\Pi_1}{4}$.
\end{example}


To represent temporal reasoning problems, 
temporal programs are complemented by
assignments $I$ and $F$
that partially or completely describe the
initial and the final state of a problem.
%
Formally, a \emph{temporal logic problem}
over some set of atoms $\mathcal{A}$ is a tuple
\tpb{\Pi}{I}{F} 
where
$\Pi$ is a temporal logic program over $\mathcal{A}$,
and $I$ and $F$ are 
assignments 
over $\mathcal{A}$.
A solution to such a problem is a sequence of states
that is consistent with the dynamics described by $\Pi$
and with the information provided by $I$ and $F$.
The possible sequences of states of length $n$, 
for some integer $n\geq 1$, 
are represented by the \emph{generator program} for $\Pi$ and $n$,
denoted by \gen{\Pi}{n},
that consists of the rules
\[\at{\choiceprogram{\mathcal{A}}}{0} \cup \btw{\Pi}{1}{n}.\]
A \emph{solution} 
to a temporal problem \tpb{\Pi}{I}{F}
is defined as a pair $(X,n)$,
where 
$n$ is an integer such that $n\geq 1$, 
and $X$ is a stable model of \gen{\Pi}{n}
consistent with $\at{I}{0}\cup\at{F}{n}$.

Temporal problems can be used to formalize planning problems, 
using a temporal logic program $\Pi$ of the form described above, 
a total assignment $I$ that assigns a value to every possible atom
(action occurrences are made false initially),
and a partial assignment $F$ to fix the goal.
The solutions of the temporal problem correspond to the plans of the planning problem.

%
%

\begin{example}\label{example:pione:solutions}
The temporal problem \tpb{\Pi_1}{\emptyset}{\emptyset} has three solutions of length $4$: 
$(X,4)$, $(X\cup\{\atomat{d}{2}\},4)$, and $(X\cup\{\atomat{b}{3}\},4)$,
where $X$ is the set of atoms
$\{\atomat{a}{0}, \atomat{b}{0}, \atomat{c}{0}, \atomat{a}{1},\atomat{b}{1},$ $ \atomat{b}{2}, \atomat{c}{3}, \atomat{d}{3}, \atomat{a}{4}, $ $\atomat{c}{4}, \atomat{d}{4}\}$.
\end{example}

To pave the way to the nogood characterization of temporal logic problems, 
we define the transition program \trans{\Pi} of a temporal logic program $\Pi$
as the (non-temporal) logic program
$\choiceprogram{\prevset{A}}\cup \Pi$
over $\mathcal{A}\cup\prevset{\mathcal{A}}$.
Each stable model of this program represents a possible transition
between a previous and a current step, 
where the former is selected by the additional choice rules
over atoms from $\prevset{\mathcal{A}}$,
and the latter is determined by the rules of $\Pi$, 
interpreted as non-temporal rules.
\begin{example} 
The transition program $\trans{\Pi_1}$ is the
(non-temporal) program $\Pi_1 \cup 
\{ \{a';b';c';d'\}\leftarrow \}$
over $\mathcal{A}_1\cup\prevset{\mathcal{A}_1}$. 
Some stable models of $\trans{\Pi_1}$ are
$\{a', b', c', a,b\}$ and
$\{c', d', $ $a, c,  d\}$, 
that correspond to the transitions 
to step $1$ and step $4$ of the solution $(X,4)$, 
respectively.
\end{example}

Next, we introduce temporal nogoods and their instantiation.
Given a temporal logic program $\Pi$ over $\mathcal{A}$, 
a temporal nogood over $\mathcal{A} \cup \body{\Pi}$ 
has the form of a (non-temporal) nogood over $\mathcal{A} \cup \prevset{\mathcal{A}} \cup \body{\Pi}$.
For a temporal nogood $\delta$ over $\mathcal{A} \cup \body{\Pi}$
and an integer $n\geq 1$, 
the instantiation of $\delta$ at $n$, 
denoted by \at{\delta}{n},
is the nogood 
that results from replacing in $\delta$ 
any signed literal $\Tlit{\alpha}$ ($\Flit{\alpha}$) by 
$\Tlit{\at{\alpha}{n}}$ (by $\Flit{\at{\alpha}{n}}$, respectively).
We extend this notation to sets of nogoods and to intervals like we did above.
For example,
$\delta_1=\{\Flit{b'},\Tlit{b}\}$ 
is a temporal nogood over
$\mathcal{A}_1\cup\body{\Pi_1}$,  
and $\btw{\delta_1}{1}{2}$ is 
$\big\{ \{\Flit{\atomat{b}{0}},\Tlit{\atomat{b}{1}}\},
    \{\Flit{\atomat{b}{1}},\Tlit{\atomat{b}{2}}\} \big\}$.
By \stepsimple{\delta} we denote the steps 
of the literals occurring in $\delta$, 
i.e., $\stepsimple{\delta} = 
\{i \mid \Tlit{\atomat{p}{i}}\in \delta\textnormal{ or }\Flit{\atomat{p}{i}}\in \delta\}$.
For example, $\stepsimple{\{\Flit{\atomat{b}{0}},\Flit{\atomat{b}{1}},\Tlit{\atomat{b}{2}}\}}=\{0,1,2\}$.



%

We are now ready to define the temporal nogoods for a temporal logic program $\Pi$ over $\mathcal{A}$. 
Recall that $\trans{\Pi}$ is a (non-temporal) logic program over 
$\mathcal{A}\cup\prevset{\mathcal{A}}$, 
whose corresponding nogoods are denoted by $\Sigma_{\trans{\Pi}}$.
Then, the set of \emph{temporal nogoods} for $\Pi$, 
denoted by $\Psi_{\Pi}$, has the form $\Sigma_{\trans{\Pi}}$, 
interpreted as a set of temporal nogoods over $\mathcal{A}\cup\body{\Pi}$,
and not as a set of (non-temporal) nogoods over $\mathcal{A}\cup\prevset{\mathcal{A}}\cup\body{\Pi}$.
%
%
\begin{example}\label{example:tempnogoods}
The set $\Psi_{\Pi_1}$ of temporal nogoods for $\Pi_1$ is
$\big\{
\{ \Tlit{a'}, \Flit{b}\},
\{ \Flit{b'}, \Tlit{b}\},
\{ \Flit{c'},$ $\Tlit{a}\},
\{ \Tlit{d'}, \Tlit{b}\}, \\
\{ \Tlit{c},  \Flit{d}\},
\{ \Flit{a'}, \Flit{c}\},
\{ \Flit{a'}, \Tlit{c'}, \Flit{a}\}
\big\}$.
\end{example}%
%

Temporal nogoods provide an alternative characterization of 
the nogoods of \gen{\Pi}{n}.
%

\begin{proposition}\label{thm:nogoods:temp:eq}
If $\Pi$ is a temporal logic program and
$n\geq 1$ 
then 
$\Sigma_{\gen{\Pi}{n}} = 
    \btw{\Psi_{\Pi}}{1}{n}$.
\end{proposition}

In words, the nogoods for \gen{\Pi}{n} 
are the same as the instantiation of the temporal nogoods for $\Pi$, 
that are nothing else than the nogoods of the logic program \trans{\Pi}
interpreted as temporal nogoods.
\begin{example}\label{example:tempnogoods:instantiation}
The set of nogoods $\Sigma_{\gen{\Pi_1}{4}}$ is equal to 
$\btw{\Psi_{\Pi_1}}{1}{4} =
\bigcup_{i \in [1,4]}\big\{
\{ \Tlit{\atomat{a}{{i-1}}}, \Flit{\atomat{b}{i}}\},
\{ \Flit{\atomat{b}{{i-1}}}, \Tlit{\atomat{b}{i}}\},$ $
\{ \Flit{\atomat{c}{{i-1}}}, \Tlit{\atomat{a}{i}}\},
\{ \Tlit{\atomat{d}{{i-1}}}, \Tlit{\atomat{b}{i}}\},
\{ \Tlit{\atomat{c}{{i}}}, \Flit{\atomat{d}{i}}\},
\{ \Flit{\atomat{a}{{i-1}}}, \Flit{\atomat{c}{i}}\},
\{ \Flit{\atomat{a}{{i-1}}}, \Tlit{\atomat{c}{{i-1}}}, \Flit{\atomat{a}{i}}\}
\big\}$.
\end{example}
By Theorem~\ref{thm:nogoods:nontight},
the temporal nogoods can be used to characterize the solutions of temporal logic problems.
%
\begin{theorem}\label{thm:nogoods:temporal}
Let \tpb{\Pi}{I}{F} be a temporal logic problem over $\mathcal{A}$. 
%
The pair $(X,n)$ is a solution to \tpb{\Pi}{I}{F}
for $n\geq 1$ and 
$X\subseteq\btw{\mathcal{A}}{0}{n}$ 
iff
$X=\Tass{S}\cap\btw{\mathcal{A}}{0}{n}$ for a (unique)
solution $S$
for $\btw{\Psi_\Pi}{1}{n}$
such that $\at{I}{0} \cup \at{F}{n} \subseteq S$.
\end{theorem}

\vspace{-3mm}
%

\section{Generalization of learned constraints}\label{sec:approach}
\vspace{-1mm}

A common software architecture to solve a temporal problem \tpb{\Pi}{I}{F}
combines a scheduler that assigns resources to different values of $n$, 
with one or many solvers that look for solutions of the assigned lengths $n$
(see~\cite{riheni06a}).
The standard approach for the solvers is to 
extend the program \gen{\Pi}{n} with facts and integrity constraints
to adequately represent $I$ and $F$, 
and call the procedure $\cdnlasp{}$ with this extended program without assumptions.
However, as we have seen in our introductory example, 
this method does not work well for our purposes, 
because it leads to a nogood representation of the initial and the final steps
that is different from 
the nogood representation of the other steps.
Hence, the constraints learned using nogoods specific to the initial and final steps
may not be generalizable to the other steps.
To overcome this issue, 
in our approach 
the solvers apply the procedure $\cdnlasp{}(\gen{\Pi}{n},$ \mbox{$\at{I}{0}\cup\at{F}{n})$}
to the generator program for $\Pi$ and $n$,
using assumptions to fix the assignments about the initial and final situations.
%
Observe that in this case, 
by Proposition \ref{thm:nogoods:temp:eq},
the solver initially 
contains exactly the nogoods $\btw{\Psi_{\Pi}}{1}{n}$,
and all the nogoods that it learns afterward are resolvents of $\btw{\Psi_{\Pi}}{1}{n}$.
%

Once this is settled, we ask ourselves:
\vspace{-1mm}
\begin{itemize}
\item[]
\emph{
What generalizations of the nogoods learned by $\cdnlasp{}$ 
can be applied to the same or other problems?
}
\vspace{-1mm}
\end{itemize}
%
%

We make the question more precise step by step.
First, instead of talking about ``the nogoods learned by the algorithm'', 
we refer to the resolvents of 
$\btw{\Psi_{\Pi}}{1}{n}$ for some temporal problem \tpb{\Pi}{I}{F}.
Or more precisely,
we refer to the resolvents of $\btw{\Psi_{\Pi}}{i}{j}$ for some $i$ and $j$ such that
$1 \leq i \leq j \leq n$,
since the learned nogoods are always the result of resolving nogoods belonging
to some interval $[i,j]$
that may be smaller than $[1,n]$.

To formalize 
the notion of the ``generalizations of nogoods'', 
we introduce some notation for shifting a non-temporal nogood an amount of $t$ time steps.
For integers $n\geq 1$ and $t$, and a non-temporal nogood $\delta$ over 
$\btw{(\mathcal{A} \cup \mathcal{A}' \cup \body{\Pi})}{1}{n}$,
the symbol \shift{\delta}{t} denotes the nogood that results from replacing in $\delta$ 
any signed literal 
$\atomat{\Tlit{\alpha}}{m}$ $(\atomat{\Flit{\alpha}}{m})$ 
by 
$\atomat{\Tlit{\alpha}}{{m+t}}$ (by $\atomat{\Flit{\alpha}}{{m+t}}$, respectively).
For example, $\shift{\delta}{0}=\delta$, and 
if $\delta=\{\Tlit{\atomat{a}{2}},\Flit{\atomat{b}{3}}\}$, then
$\shift{\delta}{1}$ is 
$\{\Tlit{\atomat{a}{3}},\Flit{\atomat{b}{4}}\}$, 
and $\shift{\delta}{-1}$ is 
$\{\Tlit{\atomat{a}{1}},\Flit{\atomat{b}{2}}\}$.
We say that \shift{\delta}{t} is a \emph{shifted version} of the nogood $\delta$, 
and that a \emph{generalization} of a nogood is a set of some of its shifted versions.
For example, 
$\{\{\Tlit{\atomat{a}{2}},\Flit{\atomat{b}{3}}\}\}$ and  
$\{ \{\Tlit{\atomat{a}{1}},\Flit{\atomat{b}{2}}\},
    \{\Tlit{\atomat{a}{2}},\Flit{\atomat{b}{3}}\},
    \{\Tlit{\atomat{a}{3}},\Flit{\atomat{b}{4}}\}\}$
are generalizations of 
$\{\Tlit{\atomat{a}{2}},\Flit{\atomat{b}{3}}\}$
and of 
$\{\Tlit{\atomat{a}{3}},\Flit{\atomat{b}{4}}\}$.


Next, by the ``other problems'' mentioned in the question, 
we refer to variations $m$ of the length of the solution,
and to variations \tpb{\Pi}{I'}{F'}
of the original problem where the initial and final situation
may change, but the temporal program remains the same.
Then, a generalization of a nogood ``can be applied'' to such problems
if it can be added to the set of nogoods used by the algorithm \cdnlasp{}
without changing the solutions to the problem.
For any variation \tpb{\Pi}{I'}{F'},
those nogoods are $\btw{\Psi_{{\Pi}}}{1}{m}$, 
and a generalization can be added to them if the generalization is entailed by them.
Hence, a generalization of a nogood ``can be applied'' to
``some problem'' \tpb{\Pi}{I'}{F'}, searching for a solution of length $m$, if the generalization is entailed by $\btw{\Psi_{{\Pi}}}{1}{m}$.
%
%
Putting all together, we can rephrase our question as follows:
\begin{itemize}
\item[]
\emph{
Given some temporal logic problem \tpb{\Pi}{I}{F},
what generalizations 
of a resolvent $\delta$ of $\btw{\Psi_{{\Pi}}}{i}{j}$
are entailed by $\btw{\Psi_{{\Pi}}}{1}{m}$?
}
\end{itemize}

\begin{figure}
    \centering
    \includegraphics[width=0.35\textwidth]{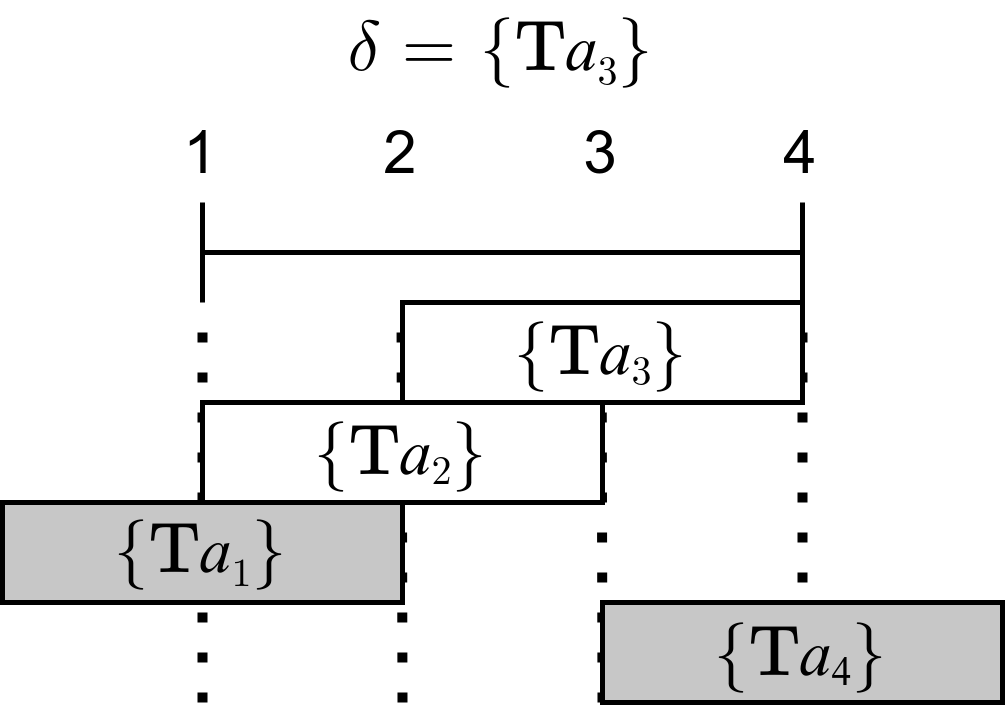}

    \caption{Representation of different shifted versions of the nogood $\delta = \{\Tlit{\atomat{a}{3}}\}$.
    The surrounding rectangles cover the interval of the nogoods needed to prove them.
    For example, the rectangle of $\{\Tlit{\atomat{a}{2}}\}$ covers the interval $[1,3]$ because
    $\{\Tlit{\atomat{a}{2}}\}$ is a resolvent of \btw{\Psi_{\Pi_1}}{1}{3}.}
    \label{fig:shifting-ex}
\end{figure}

\begin{example}\label{ex:basic}
Consider a call of $\cdnlasp{}(\gen{\Pi_1}{4},\emptyset)$
to search for a solution of length $4$ to the temporal problem 
\tpb{\Pi_1}{\emptyset}{\emptyset}. 
Initially, the solver may choose to make $\atomat{a}{3}$ 
true by adding $\Tlit{\atomat{a}{3}}$ to the initial assignment.
Then, by unit propagation, it could derive
the literal $\Tlit{\atomat{c}{2}}$ by $\{\Flit{\atomat{c}{2}}, \Tlit{\atomat{a}{3}}\}$,
the literal $\Tlit{\atomat{d}{2}}$ by $\{\Tlit{\atomat{c}{2}}, \Flit{\atomat{d}{2}}\}$,
the literal $\Flit{\atomat{b}{3}}$ by $\{\Tlit{\atomat{d}{2}}, \Tlit{\atomat{b}{3}}\}$, and
the literal $\Flit{\atomat{b}{4}}$ by $\{\Flit{\atomat{b}{3}}, \Tlit{\atomat{b}{4}}\}$,
leading to a conflict due to the violation of the nogood 
$\{ \Tlit{\atomat{a}{3}}, \Flit{\atomat{b}{4}}\}$.
At this stage, 
the solver would learn the nogood 
$\delta = \{\Tlit{\atomat{a}{3}}\}$ by resolving iteratively
$\{ \Tlit{\atomat{a}{3}}, \Flit{\atomat{b}{4}}\}$
with the nogoods 
$\{\Flit{\atomat{b}{3}}, \Tlit{\atomat{b}{4}}\}$,
$\{\Tlit{\atomat{d}{2}}, \Tlit{\atomat{b}{3}}\}$,
$\{\Tlit{\atomat{c}{2}}, \Flit{\atomat{d}{2}}\}$, and
$\{\Flit{\atomat{c}{2}}, \Tlit{\atomat{a}{3}}\}$
used for propagation.
Hence, $\delta$ is a resolvent of the set of those nogoods.
Moreover, given that those nogoods are instantiations
of some temporal nogoods of $\Psi_{\Pi_1}$ 
at the interval $[2,4]$,
$\delta$ is also a resolvent of 
\btw{\Psi_{\Pi_1}}{2}{4} and of
\btw{\Psi_{\Pi_1}}{1}{4}.
Observe that, by shifting the nogoods $1$ time point backwards, 
we obtain that 
$\shift{\delta}{-1} = \{\Tlit{\atomat{a}{2}}\}$ is a resolvent
of 
\btw{\Psi_{\Pi_1}}{1}{3}, 
and therefore also of
\btw{\Psi_{\Pi_1}}{1}{4}. 
Then, by the correctness of resolution, 
we have that the generalization
$\{\{\Tlit{\atomat{a}{2}}\},\{\Tlit{\atomat{a}{3}}\}\}$ of $\delta$
is entailed by \btw{\Psi_{\Pi_1}}{1}{4}.
%
%
On the other hand, 
$\shift{\delta}{-2} = \{\Tlit{\atomat{a}{1}}\}$ is a resolvent
of 
\btw{\Psi_{\Pi_1}}{0}{2}, but not of \btw{\Psi_{\Pi_1}}{1}{4}, 
since 
the instantiations at $0$ do not belong to \btw{\Psi_{\Pi_1}}{1}{4}.
Similarly, 
$\shift{\delta}{1} = \{\Tlit{\atomat{a}{4}}\}$ is a resolvent
of 
\btw{\Psi_{\Pi_1}}{3}{5}, but not of
\btw{\Psi_{\Pi_1}}{1}{4}, 
since 
the instantations at $5$ do not belong to
\mbox{\btw{\Psi_{\Pi_1}}{1}{4} (see Figure~\ref{fig:shifting-ex}).}
%
%
\end{example}
\vspace{-3mm}

This example suggests a sufficient condition 
for the generalization of a nogood $\delta$ learned from 
$\btw{\Psi_{\Pi}}{i}{j}$:
a shifted version $\shift{\delta}{t}$ of some generalization 
of $\delta$ is entailed by 
$\btw{\Psi_{\Pi}}{1}{n}$
if the nogoods that result from shifting 
$\btw{\Psi_{\Pi}}{i}{j}$ an amount of $t$ time points
belong to
$\btw{\Psi_{\Pi}}{1}{n}$.
%
%

\begin{theorem}\label{thm:learning:basic}
Let $\Pi$ be a temporal logic program, 
and $\delta$ be a resolvent of
$\btw{\Psi_{{\Pi}}}{i}{j}$
for some $i$ and $j$ such that $1 \leq i \leq j$.
Then,
for any $n\geq 1$, 
the set of nogoods $\btw{\Psi_{{\Pi}}}{1}{n}$ entails
the generalization 
\[\{ \shift{\delta}{t} \mid [i+t,j+t] \subseteq [1,n]\}.\]
\end{theorem}

This theorem is based on the fact that
the resolution proof that derived $\delta$ from $\btw{\Psi_{{\Pi}}}{i}{j}$
can be used to derive every 
$\shift{\delta}{t}$ from $\btw{\Psi_{{\Pi}}}{i+t}{j+t}$,
simply by shifting the nogoods $t$ time steps.
This means that $\shift{\delta}{t}$ is a resolvent of $\btw{\Psi_{{\Pi}}}{i+t}{j+t}$.
Given that $[i+t,j+t] \subseteq [1,n]$,
the nogood 
$\shift{\delta}{t}$ is also a resolvent of $\btw{\Psi_{{\Pi}}}{1}{n}$.
Then, the theorem follows from the correctness of resolution.

This result allows us to generalize the learned nogoods
to different lengths and different initial and final situations.
%
%
Following our example, 
if we were now searching for a solution of length 9
to the temporal problem \tpb{\Pi_1}{\{\Tlit{c}\}}{\{\Tlit{b}\}}, 
we could add the generalization
$\{ \{\Tlit{\atomat{a}{i}}\} \mid i \in [2,8]\}$
to 
$\cdnlasp(\gen{\Pi_1}{9},$  $\{\Tlit{\atomat{c}{0}},\Tlit{\atomat{b}{9}}\})$.
Observe that this method 
requires knowing the specific interval $[i,j]$ 
of the nogoods used to derive a learned constraint.
To obtain this information in \clingo, 
we would have to modify the implementation of the solving algorithm. 
We leave that option for future work, and 
in the next section we follow another approach
that does not require to modify the solver.

\section{When can we generalize all learned nogoods to all time steps?}\label{sec:all}

Given \emph{any temporal problem},
Theorem~\ref{thm:learning:basic} gives us
a sufficient condition for the generalization of
the nogoods learned while solving that problem.
In this section,
we investigate for \emph{what kind of temporal problems}
can we generalize all learned nogoods to all time steps. 
In other words, we would like to know when can we add the generalization 
\[\{ \shift{\delta}{t} \mid \stepsimple{\shift{\delta}{t}} \subseteq [0,n]\}\]
of a learned nogood $\delta$ to the set of nogoods used by algorithm \cdnlasp.
%

\begin{example}
In Example~\ref{ex:basic},
we saw that the nogood $\delta = \{\Tlit{\atomat{a}{3}}\}$
is a resolvent of \btw{\Psi_{\Pi_1}}{2}{4}.
By Theorem~\ref{thm:learning:basic},
it follows that the generalization
$\{\{\Tlit{\atomat{a}{2}}\},\{\Tlit{\atomat{a}{3}}\}\}$ of $\delta$
is entailed by \btw{\Psi_{\Pi_1}}{1}{n},
but that theorem does not allow us to infer that 
$\shift{\delta}{-2} = \{\Tlit{\atomat{a}{1}}\}$
is entailed by \btw{\Psi_{\Pi_1}}{1}{n}.
In fact, this would be incorrect since all the solutions
to \btw{\Psi_{\Pi_1}}{1}{n} contain the literal \atomat{\Tlit{a}}{1}.
%
But why is $\{\Tlit{\atomat{a}{1}}\}$ not entailed by \btw{\Psi_{\Pi_1}}{1}{n}?
One reason for this is that 
\btw{\Psi_{\Pi_1}}{1}{n}
does not entail the nogood
$\{ \Tlit{\atomat{c}{0}}, \Flit{\atomat{d}{0}}\}$,
that would be necessary to derive $\{\Tlit{\atomat{a}{1}}\}$. 
%
%
In fact,
since the initial state of all solutions of length $4$ to \tpb{\Pi_1}{\emptyset}{\emptyset} is $\{a,b,c\}$,
all solutions to \btw{\Psi_{\Pi_1}}{1}{4} contain the literals $\atomat{\Tlit{c}}{0}$ and $\atomat{\Flit{d}}{0}$,
violating the nogood
$\{ \Tlit{\atomat{c}{0}}, \Flit{\atomat{d}{0}}\}$.
But 
this implies that the initial state $\{a,b,c\}$ cannot have some previous state.
This is obvious looking at $\Pi_1$,
since $\{a,b,c\}$ violates the integrity constraint $\bot \leftarrow c, \neg d$.
%
On the other hand, if $\{a,b,c\}$ had some previous state, 
then it would not violate that integrity constraint, 
and therefore \btw{\Psi_{\Pi_1}}{1}{n} would entail 
$\{ \Tlit{\atomat{c}{0}}, \Flit{\atomat{d}{0}}\}$
and then $\{\Tlit{\atomat{a}{1}}\}$. 
\end{example}

This analysis suggests that we can always add a learned nogood, 
shifted backwards in time,
if the initial states
occur in some path with sufficient states before them.
A similar analysis in the other direction 
---for instance, for nogood \shift{\delta}{1} in the previous example---
suggests that 
we can add a learned nogood, 
shifted forwards in time,
if the final states occur in some path with sufficient 
states after them.
These observations lead us to this informal answer: 
we can generalize all learned nogoods 
to all time steps if the initial states have enough 
preceding states and 
the final states have enough following states.
We make this claim precise in the following.

In Section~\ref{sec:temporal} we introduced
transition programs 
and used them 
to characterize the solutions of a given temporal problem. 
Transition programs \trans{\Pi} define transitions between the states of some temporal program $\Pi$.
In turn, these transitions implicitly define a \emph{transition} graph \graph{\Pi}.
Formally, given a temporal logic program $\Pi$ over $\mathcal{A}$,
the transition graph \graph{\Pi} is the graph $( N,E )$
where $E$ is the set of edges
\[
\big\{ ( \{ a \mid \prev{a} \in X \cap \prevset{\mathcal{A}}\}, X \cap \mathcal{A}) \mid
   X \text{ is a stable model of } \trans{\Pi} \big\}
\]
and $N$ is the set of nodes occurring in some edge of $E$,
i.e., $N = \bigcup_{(X,Y)\in E}\{X,Y\}$.
Every node in $N$ is also a state of $\Pi$, so we may use both names 
interchangeably.
A \emph{path} of a transition graph $( N, E )$ is a sequence of 
states $(X_0, \ldots, X_n)$ such that
for every $i \in [1,n]$ the pair
$(X_{i-1},X_{i})$ is an edge from $E$.
We say that the \emph{length} of such a path is $n$.
We usually denote the states occurring in a path by symbols of the form $X_i$,
where the subindex $i$ represents the position of the state in the path, and 
does \emph{not} represent the instantiation of some set $X$ to time step $i$, 
which would be represented by $\at{X}{i}$.

We can characterize the solutions to a temporal problem \tpb{\Pi}{I}{F}
as the finite paths of \graph{\Pi} whose first and final nodes
are consistent with $I$ and $F$, respectively.
We state this formally in the next theorem, that extends Theorem~\ref{thm:nogoods:temporal}.

\begin{theorem}\label{thm:problemgraph}
Let 
\tpb{\Pi}{I}{F} be a temporal logic problem over $\mathcal{A}$,
$n \geq 1$, and
$X$ be a set of atoms over $\btw{\mathcal{A}}{0}{n}$.
Then, the following statements are equivalent:
\begin{itemize}
\item
The pair $(X,n)$ is a solution to \tpb{\Pi}{I}{F}.
\item
$X=\Tass{S}\cap\btw{\mathcal{A}}{0}{n}$ for a 
solution $S$
for $\btw{\Psi_\Pi}{1}{n}$
such that $\at{I}{0} \cup \at{F}{n} \subseteq S$.
\item
There is a path $( X_0, \ldots, X_n )$ 
in \graph{\Pi} such that
$X = \bigcup_{i \in [0,n]}\at{X_i}{i}$,
the state $X_0$ is consistent with $I$, and
the state $X_n$ is consistent with $F$.
\end{itemize}
\end{theorem}

\definecolor{lightgray}{gray}{0.9}
\begin{figure}
    \centering
\begin{tikzpicture}[
    set/.style={inner sep=1pt, minimum size=0.75cm, align=center, fill=lightgray, rounded corners},
    otherset/.style={inner sep=2pt, minimum size=0.75cm, align=center},
    >={Stealth[scale=0.75]}
]
\node[set] (ABC) at (0,0) {$\{a,b,c\}$};
\node[set, right=0.75cm of ABC] (AB) {$\{a,b\}$};
\node[set, above right=0.05cm and 0.75cm of AB] (BD) {$\{b,d\}$};
\node[set, below right=0.05cm and 0.75cm of AB] (B) {$\{b\}$};
\node[set, right=0.75cm of BD] (CD) {$\{c,d\}$};
\node[set, right=0.75cm of B] (BCD) {$\{b,c,d\}$};
\node[set, below right=0.05cm and 0.75cm of CD] (ACD) {$\{a,c,d\}$};

\draw[->, fill=lightgray] (ABC) -- (AB);
\draw[->] (AB) -- (BD);
\draw[->] (AB) -- (B);
\draw[->] (BD) -- (CD);
\draw[->] (B) -- (BCD);
\draw[->] (B) -- (CD);
\draw[->] (CD) -- (ACD);
\draw[->] (BCD) -- (ACD);

\draw[->, dashed, rounded corners=3mm] (AB) -- ([yshift=-0.4cm]AB.south) |- ([xshift=-0.3cm, yshift=-0.4cm]BCD.south) -- ([xshift=-0.3cm]BCD);
\draw[->, dashed, rounded corners=3mm] (ABC) -- ([yshift=-0.6cm]ABC.south) |- ([yshift=-0.6cm]BCD.south) -- (BCD);
\draw[->, dashed, rounded corners=3mm] (ABC) |- (BD); 
\draw[->, dashed, rounded corners=3mm] (ABC) |- (B); 

\node[otherset, left=0.75cm of ABC] (ABD) {$\{a,b,d\}$};
\draw[->, dashed] (ABC) -- (ABD);

\node[otherset, below=1.35cm of ACD] (BC) {$\{b,c\}$};
\draw[->, dashed] (BC) -- (ACD);

\node[otherset, left=2cm of BC] (ABCD) {$\{a,b,c,d\}$};
\draw[->, dashed] (BC) -- (ABCD);
\draw[->, dashed, rounded corners=3mm] (ABC) |- (ABCD); 

\node[otherset, above=0.85cm of ACD] (C) {$\{c\}$};
\draw[->, dashed] (C) -- (ACD);

\node[otherset, above left=0.05cm and 0.25cm of CD] (EMPTY) {$\emptyset$};
\draw[->, dashed] (EMPTY) -- (CD);
\node[otherset, above right=0.05cm and 0.25cm of CD] (D) {$\{d\}$};
\draw[->, dashed] (D) -- (CD);

\end{tikzpicture}
    \caption{
    Transition \graph{\Pi_1} of temporal program $\Pi_1$. 
    The nodes that belong to some solution of length $4$ have a gray background. 
    The transitions of those solutions are represented by normal arrows,
    while the other arrows are dashed.
    } 
    \label{fig:graph}
\end{figure}

Figure~\ref{fig:graph} represents the transition \graph{\Pi_1} of temporal program $\Pi_1$.
There are only three paths in \graph{\Pi} of length $4$: 
\begin{itemize}
\item $( \{a,b,c\}, \{a,b\}, \{b,d\}, \{c,d\}, \{a,c,d\})$, 
\item $( \{a,b,c\}, \{a,b\}, \{b\}, \{c,d\}, \{a,c,d\})$, and
\item $( \{a,b,c\}, \{a,b\}, \{b\}, \{b,c,d\}, \{a,c,d\})$.
\end{itemize}
By Theorem~\ref{thm:problemgraph}, 
each of them corresponds to one of the solutions to \tpb{\Pi_1}{\emptyset}{\emptyset} of length $4$.

Theorem~\ref{thm:problemgraph} 
establishes a relation between the solutions to the set of nogoods 
$\btw{\Psi_\Pi}{1}{n}$
and the paths of \graph{\Pi}.
This leads naturally to a relation between the nogoods entailed by 
$\btw{\Psi_\Pi}{1}{n}$
and the paths of \graph{\Pi}.

\begin{example}\label{ex:nogoodpath}
We have seen that
the set of nogoods \btw{\Psi_{\Pi_1}}{2}{4}
entails the nogood $\delta = \{\Tlit{\atomat{a}{3}}\}$.
%
%
It follows that 
\btw{\Psi_{\Pi_1}}{1}{3} entails 
$\shift{\delta}{-1} = \{\Tlit{\atomat{a}{2}}\}$. 
By Theorem~\ref{thm:problemgraph} 
we have that 
the solutions to \btw{\Psi_{\Pi_1}}{1}{3} correspond to the paths
$( X_0, X_1, X_2, X_3)$ in $\graph{\Pi_1}$.
Then, if the solutions to \btw{\Psi_{\Pi_1}}{1}{3} do not violate the nogood $\{\Tlit{\atomat{a}{2}}\}$,
it follows that the paths $( X_0, X_1, X_2, X_3)$ in $\graph{\Pi_1}$ do not violate this nogood either
---we make precise this relation between nogoods and paths below.
Hence,
the fact that \btw{\Psi_{\Pi_1}}{2}{4} entails $\{\Tlit{\atomat{a}{3}}\}$
means that,
in every path $( X_0, X_1, X_2, X_3)$ in $\graph{\Pi_1}$,
the state $X_2$ cannot include the atom $a$.
We can check this in $\graph{\Pi_1}$, 
where the only states that appear in such a position 
are $\{b,d\}$, $\{c,d\}$, $\{b\}$ and $\{b,c,d\}$, 
and neither of them contains $a$.
On the other hand, the states
$\{a,b,c\}$ and $\{a,b\}$ can contain $a$ 
because they are not in the third position of any path, 
and the same holds for the final state $\{a,c,d\}$, 
since it does not occur in the penultimate position of any path.
\end{example}

We formalize the relation between nogoods and paths.
For simplicity,
we only discuss the case where learned nogoods consist of normal atoms,
but the extension to body atoms does not pose any special challenge,
since body atoms can be seen as a conjunction of normal atoms.
Let $\Pi$ be a temporal program over $\mathcal{A}$, 
let $( X_0, \ldots, X_n )$ be some path in $\graph{\Pi}$,
and $\delta$ be some (non-temporal) nogood over 
$\btw{\mathcal{A}}{0}{n}$.
We say that the path $( X_0, \ldots, X_n )$ violates $\delta$ 
if 
\[\delta \subseteq \bigcup_{i \in [0,n]}\big( 
 \{ \Tlit{a_i} \mid a \in X_i \} \cup
 \{ \Flit{a_i} \mid a \in \mathcal{A}\setminus X_i \}
\big).\]
The right-hand side of the equation represents the path as an assignment.
%

\begin{proposition}\label{prop:nogoodpath}
Let $\Pi$ be a temporal logic program over $\mathcal{A}$, 
$n \geq 1$, and 
let $\delta$ be a (non-temporal) nogood over $\btw{\mathcal{A}}{0}{n}$.
Then, the following two statements are equivalent:
\begin{itemize}
\item The set of nogoods $\btw{\Psi_{{\Pi}}}{1}{n}$ entails $\delta$.
\item Every path 
$( X_{0}, \ldots, X_{n})$ of length $n$ in $\graph{\Pi}$
does not violate 
$\delta$. 
\end{itemize}
\end{proposition}

Our next theorem answers \emph{formally} the question of this section.
To introduce it, 
we define a node in a given graph as 
a \emph{source} if it has no incoming edges, 
as a \emph{sink} if it has no outgoing edges, 
and as an \emph{internal} node otherwise.
Then, we say that a temporal program $\Pi$ is \emph{\cyclic{}}
if every node in \graph{\Pi} is internal.
The theorem states that we can generalize all learned nogoods 
to all time steps when the temporal program is \emph{\cyclic{}}.

Observe that if $\Pi$ is \cyclic{}, then every path can be extended arbitrarily from both ends.
This means that every state has always ``enough preceding states''
and ``enough following states'', as we said earlier. 
\begin{proposition}\label{prop:extendpath:program}
Let $\Pi$ be a \cyclic{} temporal program. 
If $( X_0, \ldots, X_n )$ is a path of length $n$  in \graph{\Pi}, 
then for 
any $i,j \geq 0$
there is a path 
$( Y_{0}, \ldots, Y_{n+i+j} )$ of length $n+i+j$ in \graph{\Pi}
such that $( X_0, \ldots, X_n ) = (Y_i, \ldots, Y_{n+i})$.
\end{proposition}
%
%

\begin{example}
Program $\Pi_1$ is not \cyclic{}. 
For example, the node $\{a,b,c\}$ has no incoming edge, 
and the node $\{a,c,d\}$ has no outgoing edge.
\end{example}

\begin{example}\label{example:cyclic:one}
Let $\Pi_2$ the program that results from replacing in $\Pi_1$
the rules 
$\bot \leftarrow c, \neg d$ 
and 
$\bot \leftarrow \neg a', c', \neg a$
by 
$\bot \leftarrow b, d$ 
and 
$\bot \leftarrow c, \neg b$.
One can check that program $\Pi_2$ is cyclic, 
and that the 
nogood $\{\atomat{\Flit{b}}{1}, \atomat{\Flit{a}}{2}\}$ 
is a resolvent of $\btw{\Psi_{{\Pi_2}}}{2}{3}$. 
By Theorem~\ref{thm:learning:basic}, we can conclude that 
$\{\atomat{\Flit{b}}{0}, \atomat{\Flit{a}}{1}\}$ and
$\{\atomat{\Flit{b}}{1}, \atomat{\Flit{a}}{2}\}$
are entailed by $\btw{\Psi_{{\Pi_2}}}{1}{3}$, 
but we cannot do the same about 
$\{\atomat{\Flit{b}}{2}, \atomat{\Flit{a}}{3}\}$.
On the other hand, by Proposition~\ref{prop:nogoodpath}, 
the fact that 
$\btw{\Psi_{{\Pi_2}}}{1}{3}$ entails  
$\{\atomat{\Flit{b}}{1}, \atomat{\Flit{a}}{2}\}$ 
implies that every path 
$ ( X_0, \ldots, X_3 )$ of length $3$ in \graph{\Pi_2} 
does not violate $\{\atomat{\Flit{b}}{1}, \atomat{\Flit{a}}{2}\}$.
%
Given that $\Pi$ is \cyclic{}, 
it follows that 
every path $( Y_0, Y_1 )$ of length $1$ in \graph{\Pi_2} 
does not violate $\{\atomat{\Flit{b}}{0}, \atomat{\Flit{a}}{1}\}$.
Otherwise, 
by Proposition~\ref{prop:extendpath:program} 
there would be some path $(X_0, \ldots, X_3)$ of length $3$ in $\graph{\Pi}$ 
such that $(Y_0, Y_1) = (X_1, X_2)$,
and this path would violate $\{\atomat{\Flit{b}}{1}, \atomat{\Flit{a}}{2}\}$
---which would be a contradiction.
%
Given that the paths $( Y_0, Y_1 )$ in \graph{\Pi_2} 
do not violate $\{\atomat{\Flit{b}}{0}, \atomat{\Flit{a}}{1}\}$,
Proposition~\ref{prop:nogoodpath}
implies that $\btw{\Psi_{{\Pi_2}}}{1}{1}$ entails 
$\{\atomat{\Flit{b}}{0}, \atomat{\Flit{a}}{1}\}$.
Therefore, 
$\btw{\Psi_{{\Pi_2}}}{i}{i}$ entails 
$\{\atomat{\Flit{b}}{{i-1}}, \atomat{\Flit{a}}{i}\}$
for $i \in [1,4]$, and 
$\btw{\Psi_{{\Pi_2}}}{1}{4}$ entails 
all those nogoods together.
In other words, 
$\btw{\Psi_{{\Pi_2}}}{1}{4}$
entails all shifted versions of 
$\{\atomat{\Flit{b}}{1}, \atomat{\Flit{a}}{2}\}$
that fit in the interval $[0,4]$.
\end{example}

\begin{theorem}\label{thm:cyclic:program}
Let $\Pi$ be a temporal logic program,
and $\delta$ be a resolvent of
$\btw{\Psi_{{\Pi}}}{i}{j}$
for $1 \leq i \leq j$.
If $\Pi$ is \cyclic{},
for any $n\geq 1$, 
the set of nogoods $\btw{\Psi_{{\Pi}}}{1}{n}$
entails the generalization
\[\{ \shift{\delta}{t} \mid \stepsimple{\shift{\delta}{t}} \subseteq [0,n]\}.\]
\end{theorem}

Theorem~\ref{thm:cyclic:program}
allows the generalization of learned nogoods to all time steps 
when the temporal program is \cyclic{}. 
Unfortunately, 
\cyclic{} temporal programs are not very useful for representing planning problems. 
To see this, consider some planning problem and any of its transitions, 
where an action $a$ in state $X$ results in state $Y$.
If actions occur at the same time point as their effects
---as in our introductory example---
the transition graph 
must have an edge between $X$ and $Y \cup \{a\}$.
%
Since actions at the previous step can be chosen freely, 
for any action $b$ there will also be an edge between 
$X \cup \{b\}$ and $Y \cup \{a\}$.
%
%
But if the temporal program 
is \cyclic{}, then
$X \cup \{b\}$ must have an incoming edge. 
Hence, $X$ must result from executing action $b$ in some state.
Overall, this means that every state $X$ that leads to another state $Y$ 
must be the result of \emph{every possible} action under some previous state,
which clearly does not apply to many planning problems.
%
%
%
Something similar happens 
if action occurrences are placed before 
their effects.

To address this issue, 
we extend our study by taking into account 
the initial states associated with temporal logic problems.
We refine the definition of being internal 
by considering only the initial states and 
the states that are reachable from them.
This modification resolves the problem observed in our previous example: 
if the state $X \cup \{b\}$ has no incoming edges, 
then it is not reachable from any initial state, and no condition is imposed on it. 


Let $\Pi$ be a temporal logic program and $I$ be a partial assignment.
A state $X$ of $\Pi$ is
\begin{itemize}
\item \emph{initial wrt} $I$
if it is consistent with $I$ and has some outgoing edge in $\graph{\Pi}$; 
\item 
\emph{reachable wrt} $I$ if it can be reached 
from some initial state of $\Pi$ wrt $I$,
i.e.,
if there is some path $( Y, \ldots, X )$ in $\graph{\Pi}$ 
starting at some initial state 
$Y$ of $\Pi$ wrt $I$; and  
\item 
\emph{loop-reachable} if it can be reached
from some loop in \graph{\Pi}, 
i.e., 
if there is some path of the form $(Y, \ldots, Y, \ldots, X)$ in \graph{\Pi} for some state $Y$ of $\Pi$.
\end{itemize}
The temporal program $\Pi$ is \emph{\cyclic{} wrt} $I$
if these conditions hold: 
\begin{enumerate}[(i)]
\item 
Every initial state wrt $I$ is loop-reachable. 
\item
Every reachable state wrt $I$ is internal. 
\end{enumerate}
%
%
%
Condition (i) guarantees that initial states have ``enough previous states'', 
and condition (ii) guarantees that reachable states have
``enough successor states''. 
Together, they imply that 
any path starting at some initial or reachable state can be extended indefinitely in both directions.
\begin{proposition}\label{prop:extendpath:problem}
Let $\Pi$ be a temporal logic program and 
$I$ be a partial assignment such that $\Pi$ is \cyclic{} wrt $I$.
If $( X_0, \ldots, X_n )$ is a path of length $n$ in \graph{\Pi} and $X_0$ is 
initial or reachable wrt $I$,
then for 
any $i,j \geq 0$
there is a path 
$( Y_{0}, \ldots, Y_{n+i+j} )$ of length $n+i+j$ in \graph{\Pi}
such that $( X_0, \ldots, X_n ) = (Y_i, \ldots, Y_{n+i})$.
\end{proposition}
%

Temporal programs that are \cyclic{} with respect to a partial assignment
generalize the notion of \cyclic{} temporal programs. 

\begin{proposition}
A temporal logic program $\Pi$ is \cyclic{} iff it is \cyclic{} wrt the empty assignment.
\end{proposition}

If a temporal program is \cyclic{} wrt an assignment $I$, 
we can generalize all learned nogoods to all time steps
provided that $I$ \emph{holds initially}. 
We enforce this condition by adding the set of nogoods 
$
 \big\{ \{ \Flit{\atomat{a}{0}} \} \mid \Tlit{a} \in I \big\} \cup
 \big\{ \{ \Tlit{\atomat{a}{0}} \} \mid \Flit{a} \in I \big\}
$
that we denote by 
$\assignfornogood{I}$.
\begin{theorem}\label{theorem:cyclic:problem}
Let $\Pi$ be a temporal logic program,
$I$ be a partial assignment,
and $\delta$ be a resolvent of
$\btw{\Psi_{{\Pi}}}{i}{j}$
for $1 \leq i \leq j$.
%
If $\Pi$ is \cyclic{} wrt $I$, then 
for any $n\geq 1$, 
the set of nogoods $\btw{\Psi_{{\Pi}}}{1}{n}\cup \assignfornogood{I}$
entails the generalization
\[\{ \shift{\delta}{t} \mid \stepsimple{\shift{\delta}{t}} \subseteq [0,n]\}.\]
\end{theorem}

In theory, the inclusion of $\assignfornogood{I}$ 
appears to conflict with our goal of applying the learned nogoods
to variations of the original problem with different initial states.
But in practice 
this need not pose a problem, 
since the assignment $I$ may 
reflect a form of representing problems in general
---such as placing actions at the step where their effects occur---
and be independent of any specific instance.  
As a result, the learned nogoods may remain applicable to all intended variations
of the original problem.

This observation holds for 
most representations of planning problems that we have encountered, 
which can be captured by temporal logic programs \cyclic{} wrt a generic assignment $I$.
%
%
Such an assignment ensures that
\begin{enumerate}[(a)]
\item
initial states have no action occurrences.
\end{enumerate}
To achieve this, 
actions must occur at the same time step as their effects. 
In ASP, this is usually a matter of convenience.
To satisfy condition (i) of being \cyclic{} wrt $I$, it suffices if:
\begin{enumerate}[(a)]
\setcounter{enumi}{1}
\item
whenever no action occurs in a state, that state remains unchanged.
\end{enumerate}
Conditions (a) and (b) create a loop in all initial states, making them loop-reachable.
To meet condition (b), 
the representation should allow 
the non-execution of actions, 
and the law of inertia should make all fluents persist. 
Most representations of planning problem conform to this,
or can be adapted with minor changes. 
Condition (ii) of being \cyclic{} wrt $I$ can be satisfied directly if
after the execution of each action, another action can always be applied.
%
An alternative and more generic condition is the following: 
\begin{enumerate}[(a)]
\setcounter{enumi}{2}
\item
from every state 
with an action occurrence,
there is a transition 
to the same state without action occurrences.
\end{enumerate}
In addition to the representation of inertia, 
this condition only requires that the representation always allows the non-execution of actions. 
Based on our experience, most representations of planning problems satisfy conditions (a-c)
or, as we have seen in our introductory example, can be easily modified to do that.
In fact, the encodings used in our conference paper~\cite{roscst22a}
required only minor changes to fit within this approach.

In summary, 
our methodology represents a planning problem as a temporal logic problem
$\tpb{\Pi}{I}{F}$ 
such that 
the assignment $I$ is divided into:
\begin{itemize}
\item
A generic assignment $I_1$ that 
ensures 
condition (a).
\item
A specific assignment $I_2$ that describes the initial situation of the planning problem.
\end{itemize}
%
Moreover, the temporal program $\Pi$ 
is construed to satisfy conditions (b) and (c), ensuring that it is \cyclic{} wrt $I_1$.
%
Given this, to solve
$\tpb{\Pi}{I}{F}$ 
we run the procedure 
$\cdnlasp{}(\gen{\Pi}{n},\at{I_1}{0}\cup\at{I_2}{0}\cup\at{F}{n})$ for different lengths $n$.
For each run, 
any nogood $\delta$ learned by the solver is a resolvent of $\btw{\Psi_{{\Pi}}}{1}{n}$.
By Theorem~\ref{theorem:cyclic:problem},
we can generalize $\delta$ across all time steps,
also in other runs with different lengths, 
and in other variations of the original problem.
%
%
Such variatons can be represented as 
temporal logic problems
$\tpb{\Pi}{I_1 \cup I_2'}{F'}$
where $\Pi$ and $I_1$ remain unchanged. 
As before, we can solve them using 
$\cdnlasp{}(\gen{\Pi}{n},\at{I_1}{0}\cup\at{I_2'}{0}\cup\at{F'}{n})$
for different lengths $n$.
Since the solutions computed by such calls are solutions of
$\btw{\Psi_{{\Pi}}}{1}{n} \cup \assignfornogood{I_1}$, 
by Theorem~\ref{theorem:cyclic:problem}
the generalization of $\delta$ to all time steps can also be added to them.

\section{From non-\cyclic{} to \cyclic{} temporal programs}\label{sec:cyclic:encoding}
If we want to solve the temporal problem 
$\tpb{\Pi_1}{\emptyset}{\emptyset}$ 
and generalize the nogoods learned during this process, 
the results from the previous section are not applicable
because $\Pi_1$ is not \cyclic{}, 
and the initial state is empty. 
To address this limitation, 
we introduce a method to translate any temporal program into an \cyclic{} one 
that preserves the same solutions 
modulo the original atoms. 
%
%
We can solve the original problem 
using the new program, 
and 
given that this program is \cyclic{}, 
we can generalize all learned nogoods to all time steps.
Moreover, 
we can apply these nogoods
to the original program, after eliminating the auxiliary atoms
introduced by the translation. 

Let $\Pi$ be a temporal logic program over $\mathcal{A}$, 
and let $\trbsymbol$ be a fresh atom not included in $\mathcal{A}$.
We define $\trbdo{\Pi}$ as the temporal logic program
\begin{align}
\{ \{ \trbsymbol{} \} \leftarrow \} \ & \cup \label{eq:lambda:one}\\
\{ \head{r} \leftarrow \body{r} \cup \{ \trbsymbol{}\} \mid r \in \Pi \} \ & \cup \label{eq:lambda:two}\\
\{ \{ a \} \leftarrow \neg\trbsymbol{} \mid a \in \mathcal{A} \}. \ & \label{eq:lambda:tri}
\end{align}
This translation
extends $\Pi$ by introducing a choice rule for $\trbsymbol{}$,
tagging the rules of $\Pi$ with $\trbsymbol{}$, and
allowing the selection of any subset of $\mathcal{A}$ when $\trbsymbol$ does not hold.

To understand the translation, 
let us study the relationship between the transition graphs defined by $\Pi$ and $\trbdo{\Pi}$. 
These graphs are defined by the corresponding transition programs: 
\begin{itemize}
\item $\trans{\Pi} = \Pi \cup \choiceprogram{\mathcal{A}'}$, and
\item $\trans{\trbdo{\Pi}} = \trbdo{\Pi}\cup \choiceprogram{\mathcal{A}'\cup\{\trbsymbol{}'\}}$. 
\end{itemize}
%
The behavior of $\trans{\trbdo{\Pi}}$ 
depends on whether $\trbsymbol{}$ is 
chosen at~\eqref{eq:lambda:one} or not.

If $\trbsymbol{}$ is chosen,
then  $\trans{\trbdo{\Pi}}$ has the same stable models as the program 
\[\{\trbsymbol\leftarrow\}\cup\trans{\Pi}\cup\{\{\trbsymbol'\}\leftarrow\}\]
as the rules in~\eqref{eq:lambda:two} 
can be simplified to the original rules from $\Pi$, 
and the rules in~\eqref{eq:lambda:tri} become irrelevant.
Then, for every stable model $X$ of $\trans{\Pi}$,
the sets $X \cup \{\trbsymbol{}\}$ 
and $X \cup \{\trbsymbol{},\trbsymbol{}'\}$
are 
stable models of $\trans{\trbdo{\Pi}}$.
%
%
In terms of transition graphs, 
this means that for every edge $(X,Y)$ in \graph{\Pi} 
there are two corresponding edges $(X,Y\cup\{\trbsymbol\})$ and $(X\cup\{\trbsymbol\},Y\cup\{\trbsymbol\})$
in \graph{\trbdo{\Pi}}.

If $\trbsymbol{}$ is not chosen, 
then  $\trans{\trbdo{\Pi}}$ has the same stable models as the program 
\[\{ \{ a \} \leftarrow \mid a \in \mathcal{A}\} \cup \choiceprogram{\mathcal{A}'\cup\{\trbsymbol{}'\}}\]
as the rules in~\eqref{eq:lambda:two} have no effect, 
and the body of~\eqref{eq:lambda:tri} becomes true. 
Then, for every $X,Y \subseteq \mathcal{A}$, 
if $X'$ denotes the set $\{ a' \mid a \in X\}$, 
the sets $X' \cup Y$ and
$X' \cup Y \cup \{\trbsymbol{}'\}$ are stable models of $\trans{\trbdo{\Pi}}$.
Accordingly, 
\graph{\trbdo{\Pi}} contains the
edges $(X,Y)$ and $(X\cup\{\trbsymbol\},Y)$.

\newpage
\begin{proposition}\label{prop:labmda:graph}
Let $\Pi$ be a temporal logic program, and
let $E$ be the set of edges in its transition graph $\graph{\Pi}$. 
%
The set of edges in \graph{\trbdo{\Pi}} is 
the union of the following four sets:
\begin{itemize}
    \item
    $\{(X,Y\cup\{\trbsymbol\}) \mid (X,Y) \in E\}$ 
    \item
    $\{(X\cup\{\trbsymbol\},Y\cup\{\trbsymbol\}) \mid (X,Y) \in E\}$ 
    \item
    $\{(X,Y) \mid X,Y \subseteq \mathcal{A}\}$ 
    \item
    $\{(X\cup\{\trbsymbol\},Y) \mid X,Y \subseteq \mathcal{A}\}$ 
\end{itemize}
\end{proposition}
Given these sets, particularly the second, 
there is a one-to-one correspondence 
between the paths in $\graph{\Pi}$ 
and the paths in $\graph{\trbdo{\Pi}}$
where $\trbsymbol{}$ belongs to all states. 
This correspondence holds even
if 
$\trbsymbol{}$ does not occur in the initial state of the path, 
since the first edge can also be taken from the first set of 
the previous proposition. 
By Theorem~\ref{thm:problemgraph},  
this correspondence between paths induces 
a correspondence 
between the solutions to $\Pi$ and the solutions to 
$\trbdo{\Pi}$ where $\trbsymbol{}$ is true always after the initial step.
To formalize this, 
we say that a solution $(X,n)$ to a given temporal problem is \emph{\trbnormal{}} if
$\atomat{\trbsymbol}{0}\notin X$ and 
$\atomat{\trbsymbol}{i}\in X$ for all $i \in [1,n]$. 
\begin{proposition}\label{prop:cyclic:trb}
Let $\tpb{\Pi}{I}{F}$ be a temporal logic problem. 
There is a one-to-one correspondence between the 
solutions to $\tpb{\Pi}{I}{F}$ 
and the \trbnormal{} solutions to 
$\tpb{\trbdo{\Pi}}{I}{F}$.
\end{proposition}

\begin{example}
The temporal problem \tpb{\Pi_1}{\emptyset}{\emptyset} has three solutions of length $4$: 
$(X,4)$, $(X\cup\{\atomat{d}{2}\},4)$, and $(X\cup\{\atomat{b}{3}\},4)$,
where $X$ is defined as in Example~\ref{example:pione:solutions}. 
These solutions  correspond one-to-one to 
the 
three \trbnormal{} solutions of the same length to
\tpb{\trbdo{\Pi_1}}{\emptyset}{\emptyset}:
$(X\cup Y ,4)$, $(X\cup\{\atomat{d}{2}\}\cup Y,4)$, and $(X\cup\{\atomat{b}{3}\} \cup Y,4)$,
where $Y$ is $\{\atomat{\trbsymbol}{1},\atomat{\trbsymbol}{2},
                \atomat{\trbsymbol}{3},\atomat{\trbsymbol}{4}\}$.
\end{example}
The call 
$\cdnlasp{}(\gen{\trbdo{\Pi}}{n},$ $
            \at{I}{0}\cup\at{F}{n}\cup
            \{\Flit{\atomat{\trbsymbol{}}{0}},
              \Tlit{\atomat{\trbsymbol{}}{1}},\ldots,
              \Tlit{\atomat{\trbsymbol{}}{n}}\})$
computes \trbnormal{} solutions to $(\trbdo{\Pi},I,F)$,
enforcing the correct value for $\trbsymbol$ at every time point using assumptions.
The solutions to the original problem $(\trbdo{\Pi},I,F)$ 
can be obtained from these \trbnormal{} solutions
by removing the atoms $\{\atomat{\trbsymbol}{1}, \ldots, \atomat{\trbsymbol}{n}\}$.

The following proposition establishes that 
the program $\trbdo{\Pi}$ is \cyclic{} wrt $\{\Flit{\atomat{\trbsymbol}}{0}\}$.
To see why, 
observe that 
the third and fourth sets of Proposition~\ref{prop:labmda:graph}
show that,
in $\graph{\trbdo{\Pi}}$,
every state $X$ --whether it contains $\trbsymbol{}$ or not--- 
is connected to the state $X \setminus \{\trbsymbol{}\}$. 
This implies that:
\begin{enumerate}[(a)]
\item every state without $\trbsymbol{}$ is connected to itself,
and
\item every state has a successor state.
\end{enumerate}
%
Condition (a) provides condition (i) for 
$\Pi$ being \cyclic{} wrt $\{\Flit{\atomat{\trbsymbol}}{0}\}$,
since initial states do not have $\trbsymbol$; 
and condition (b) ensures the corresponding condition (ii).
%

\begin{proposition}\label{prop:cyclic:translation}
For any temporal program $\Pi$, 
the program $\trbdo{\Pi}$ is \cyclic{} wrt $\{\Flit{\atomat{\trbsymbol}}{0}\}$.
\end{proposition}

Since 
$\trbdo{\Pi}$ is \cyclic{} wrt $\{\Flit{\atomat{\trbsymbol}}{0}\}$, 
by Theorem~\ref{theorem:cyclic:problem} we obtain the next result,
that allows us to generalize all nogoods learned using
$\Psi_{{\trbdo{\Pi}}}$
to all time steps, 
as long as $\trbsymbol$ is initially false.

\begin{theorem}\label{thm:translation:generalization}
Let $\Pi$ be a temporal logic program,
and $\delta$ be a resolvent of
$\btw{\Psi_{{\trbdo{\Pi}}}}{i}{j}$
for some $i$ and $j$ such that $1 \leq i \leq j$.
%
For any $n\geq 1$, 
the set of nogoods $\btw{\Psi_{{\trbdo{\Pi}}}}{1}{n}\cup \big\{\{\Tlit{\atomat{\trbsymbol}{0}}\}\big\}$
entails the generalization
\[\{ \shift{\delta}{t} \mid \stepsimple{\shift{\delta}{t}} \subseteq [0,n]\}.\]
\end{theorem}

\begin{example}
The set $\Psi_{{\trbdo{\Pi_1}}}$ of temporal nogoods of $\trbdo{\Pi}$ 
is 
$\big\{ \delta \cup \{\Tlit{\trbsymbol}\} \mid \delta \in \Psi_{\Pi_1}\big\}$
($\Psi_{\Pi_1}$ is given in Example~\ref{example:tempnogoods}).
We showed
in Example~\ref{ex:basic}
that $\{\Tlit{\atomat{a}{3}}\}$ is a resolvent of 
\btw{\Psi_{\Pi_1}}{2}{4}.
Similarly, the nogood
$\delta = \{\Tlit{\atomat{a}{3}}, 
            \Tlit{\atomat{\trbsymbol}{2}}, 
            \Tlit{\atomat{\trbsymbol}{3}}, 
            \Tlit{\atomat{\trbsymbol}{4}}\}$
is a resolvent of
\btw{\Psi_{\trbdo{\Pi_1}}}{2}{4}, 
derived by iteratively resolving
$\{ \Tlit{\atomat{a}{3}}, \Flit{\atomat{b}{4}}, \Tlit{\atomat{\trbsymbol}{4}}\}$
with
$\{\Flit{\atomat{b}{3}}, \Tlit{\atomat{b}{4}}, \Tlit{\atomat{\trbsymbol}{4}}\}$,
$\{\Tlit{\atomat{d}{2}}, \Tlit{\atomat{b}{3}}, \Tlit{\atomat{\trbsymbol}{3}}\}$,
$\{\Tlit{\atomat{c}{2}}, \Flit{\atomat{d}{2}}, \Tlit{\atomat{\trbsymbol}{2}}\}$, and
$\{\Flit{\atomat{c}{2}}, \Tlit{\atomat{a}{3}}, \Tlit{\atomat{\trbsymbol}{3}}\}$.
Note that the additional $\Tlit{\atomat{\trbsymbol}{i}}$ literals in $\delta$ 
indicate the time steps of the nogoods used in the resolution proof.
The nogood $\delta$ and its shifted versions 
$\shift{\delta}{-1} = \{\Tlit{\atomat{a}{2}}, 
            \Tlit{\atomat{\trbsymbol}{1}}, 
            \Tlit{\atomat{\trbsymbol}{2}}, 
            \Tlit{\atomat{\trbsymbol}{3}}\}$
and 
$\shift{\delta}{-2} = \{\Tlit{\atomat{a}{1}}, 
            \Tlit{\atomat{\trbsymbol}{0}}, 
            \Tlit{\atomat{\trbsymbol}{1}}, 
            \Tlit{\atomat{\trbsymbol}{2}}\}$ 
fall within the interval $[0,4]$.
Hence,
by Theorem~\ref{thm:translation:generalization},
they 
are entailed by 
$\btw{\Psi_{{\trbdo{\Pi_1}}}}{1}{4} \cup \{\Tlit{\atomat{\trbsymbol}{0}}\}$.
In particular, 
$\shift{\delta}{-2}$ is trivially entailed, 
since the solutions to 
$\btw{\Psi_{{\trbdo{\Pi_1}}}}{1}{4} \cup \{\Tlit{\atomat{\trbsymbol}{0}}\}$
cannot include $\Tlit{\atomat{\trbsymbol}{0}}$, 
which belongs to $\shift{\delta}{-2}$. 
\end{example}

The nogoods learned from the program 
$\trbdo{{\Pi}}$
can be applied to 
the original program $\Pi$
after removing
all auxiliary atoms.
%
We define the \emph{simplification} of a nogood $\delta$, denoted by $\simp{\delta}$, 
as the nogood obtained from $\delta$ after removing 
all literals of the form $\Flit{\atomat{\trbsymbol}{i}}$ or $\Tlit{\atomat{\trbsymbol}{i}}$.
for any integer $i$.

\begin{theorem}\label{thm:translation:generalization:original}
Let $\Pi$ be a temporal logic program,
and let $\delta$ be a resolvent of
$\btw{\Psi_{{\trbdo{\Pi}}}}{i}{j}$
for some $i$ and $j$ such that $1 \leq i \leq j$.
For any $n\geq 1$, 
the set of nogoods $\btw{\Psi_{{\Pi}}}{1}{n}$ 
entails the generalization
\[\{ \simp{\shift{\delta}{t}} \mid \stepsimple{\shift{\delta}{t}} \subseteq [0,n], 
        \Tlit{\atomat{\trbsymbol}{0}}\notin \shift{\delta}{t} \}.\]
\end{theorem}

\begin{example}
The nogood 
$
\delta = \{\Tlit{\atomat{a}{3}},
\Tlit{\atomat{\trbsymbol}{2}}, 
\Tlit{\atomat{\trbsymbol}{3}}, 
\Tlit{\atomat{\trbsymbol}{4}} 
\}$ 
is a resolvent of
$\btw{\Psi_{{\trbdo{\Pi_1}}}}{2}{4}$.
Both $\delta$ and its shifted versions 
$\shift{\delta}{-1}$ and 
$\shift{\delta}{-2}$ 
fall within the interval $[0,4]$, 
but  
$\shift{\delta}{-2}$ includes 
$\Tlit{\atomat{\trbsymbol}{0}}$. 
Hence, by Theorem~\ref{thm:translation:generalization:original}, 
the set of nogoods 
$\btw{\Psi_{{\Pi_1}}}{1}{4}$ 
entails 
$\simp{\delta} = \{\Tlit{\atomat{a}{3}}\}$ 
and 
$\simp{\shift{\delta}{-1}} = \{\Tlit{\atomat{a}{2}}\}$.
Note that the condition 
$\Tlit{\atomat{\trbsymbol}{0}}\notin \shift{\delta}{t}$
of Theorem~\ref{thm:translation:generalization:original} is necessary, 
as without it, we would \emph{incorrectly} infer that 
$\simp{\shift{\delta}{-2}} = \{\Tlit{\atomat{a}{1}}\}$ 
is entailed by
$\btw{\Psi_{{\Pi_1}}}{1}{4}$.
In the end, we arrive at the same conclusion as in Example~\ref{ex:basic},
but in that case
our reasoning relied on prior knowledge 
---not available to the \cdnlasp{} algorithm---
about the time steps $[2,4]$ 
of the nogoods used to derive $\{\Tlit{\atomat{a}{3}}\}$, 
while in this case 
the information about the time steps 
is encoded directly in $\delta$  
through the $\Tlit{\atomat{\trbsymbol}{i}}$ literals, 
making this additional knowledge unnecessary. 
\end{example}

\section{Experiments}\label{sec:experiments}

We experimentally evaluate 
the generalization of learned nogoods in ASP planning 
using the solver \clingo.
The goal of the experiments is to study the performance of 
\clingo{} when the planning encodings are extended by 
the generalizations of some constraints learned by \clingo{} itself.
We are interested only in the solving time and not in the grounding time,
but in any case we have observed no
differences between grounding times among the different configurations. 
We did experiments in two different settings, single shot and multi shot,
that we detail below.
Following the approach of~\cite{gekakalurosc16a},
in all experiments we disregarded the learned nogoods of size greater than $50$ and
of degree greater than $10$, 
where the degree of a nogood is defined as 
the difference between the maximum and minimum step of the literals of the nogood\footnote{These values are experimentally chosen. The higher the size and degree of the nogoods, the less useful they are as they become more specific.}.
In all the experiments, the learned nogoods are always sorted either by size or 
by \textit{literal block distance} (\lbd, \cite{audsim09a}), 
a measure that is usually associated with the quality of a learned nogood.
We tried configurations adding the best $500$, $1000$, or $1500$ nogoods, 
according to either their nogood size or their \lbd.
The results ordering the nogoods by \lbd\ were similar but slightly better 
than those ordering by size, and here we focus on them.
%
%
We used two benchmark sets from~\cite{digelurosc18a}.
The first consists of PDDL benchmarks from planning competitions,
translated to ASP using the system \plasp\ presented in that paper.
This set contains 120 instances of 6 different domains.
The second set consists of ASP planning benchmarks from ASP competitions.
It contains 136 instances of 9 domains.
We adapted the logic programs of these benchmarks to the format of temporal logic programs 
as follows:
we deleted the facts used to specify the initial situation, 
as well as the integrity constraints used to specify the goal, 
we added some choice rules to open the initial situation, and
we fixed the initial situation and the goal using assumptions.
All benchmarks were run using the version $5.5.1$ of \clingo\
on an Intel Xeon E5-2650v4 under Debian GNU/Linux 10, 
with a memory limit of 8 GB,
and a timeout of 15 minutes per instance.

The task in the single shot experiment 
is to find a plan of a fixed length $n$ that is part of the input.
For the PDDL benchmarks we consider plan lengths varying from $5$ to $75$ in steps of $5$ units, 
for a total of $2040$ instances.
The ASP benchmarks already have a plan length, and we use it.
In a preliminary learning step,
\clingo\ is run with every instance for $10$ minutes or until $16000$ nogoods are learned, 
whatever happens first.
The actual learning time is disregarded and not taken into account in the tables.
Some PDDL instances overcome the memory limit in this phase\footnote{Since initial and goal conditions are left open, the grounding size increases and can not fit into memory}.
We leave them aside and are left with $1663$ instances of this type.
We compare the performance of \clingo\ running normally (baseline), 
versus the (learning) configurations where we add the best $500$, $1000$, or $1500$ learned nogoods
according to their \lbd\ value.
%
In this case we apply Theorem~\ref{theorem:cyclic:problem} 
and learn the nogoods using a slight variation of the original encoding, 
but use the original encoding \mbox{for the evaluation of all configurations.}
The results applying the translations from ~\ref{sec:translations} are similar.
%
They can be found in the ~\ref{sec:appendix:tables}.

Tables~\ref{tab:1shot-pddl} and~\ref{tab:1shot-asp} 
show the results for the PDDL and the ASP benchmarks, respectively.
The first columns include the name and number of instances of every domain.
The tables show the average solving times
and the number of timeouts, in parenthesis,
for every configuration and domain. 
We can observe that in general the learning configurations 
are faster than the baseline and in some domains they solve more instances.
The improvement is not huge, but is persistent among the different settings.
The only exception is the \emph{elevator} domain in PDDL, 
where the baseline is a bit faster than the other configurations.
We also analyzed the average number of conflicts per domain and configuration, 
and the results follow the same trend as the solving times.

In the Multi shot solving experiment, 
the solver first looks for a plan of length $5$.
If the solver finds no such plan, then it looks for a plan of length $10$, 
and so on until it finds a plan.
At each of these solver calls, we collect the best learned nogoods. 
Then, before the next solver call, we add the generalization of 
the best $500$, $1000$, or $1500$ of them, 
depending on the configuration.
As before, we rely on Theorem~\ref{theorem:cyclic:problem}, 
but this time we use the same original encoding, 
slightly modified, for both learning and solving.\footnote{
The results using the translations from our conference paper~\cite{roscst22a}
are similar (see~\ref{sec:appendix:tables}).}
The results for PDDL and ASP are shown in Tables~\ref{tab:mshot-pddl} and \ref{tab:mshot-asp-degm1}, 
respectively.
In both of them, the baseline and the different configurations perform similarly, 
and there is not a clear trend.
%
%
The analysis of the number of conflicts shows similar results.

\begin{table}[h!]\footnotesize
    \begin{center}
    \newcommand{\fullline}[5]{\domname#1&\hspace{5pt}\tablevalue#2 &\hspace{5pt}\tablevalue#3 &\hspace{5pt}\tablevalue#4 &\hspace{5pt}\tablevalue#5 \\}
    \newcommand{\domname}[2]{\textbf{#1} & (#2)}
    \newcommand{\tablevalue}[2]{#1\ (#2)}
    
    \( \def\arraystretch{0.2}
    \begin{array}{lr r r r r }
        \multicolumn{2}{c}{}{} &  \qquad\qquad\textbf{baseline} &  \qquad\qquad\textbf{500}  &   \qquad\qquad\textbf{1000}  &   \qquad\qquad\textbf{1500} \\
        \midrule
        \fullline {{blocks} {300}}  {{\textbf{0.5}}{0}}            {{0.6}{0}}            {{0.6}{0}}            {{0.6}{0}}            \\
        \fullline {{depots} {270}}  {{146.8}{30}}         {{140.2}{30}}         {{\textbf{124.2}}{24}}         {{135.5}{28}}         \\
        \fullline {{driverlog} {135}}  {{14.0}{1}}           {{13.5}{1}}           {{11.5}{1}}           {{\textbf{10.8}}{1}}           \\
        \fullline {{elevator} {300}} {{\textbf{3.0}}{0}}            {{5.1}{0}}            {{4.3}{0}}            {{5.2}{0}}             \\
        \fullline {{grid} {30}}  {{11.4}{0}}           {{6.0}{0}}            {{4.4}{0}}            {{\textbf{3.7}}{0}}           \\
        \fullline {{gripper} {255}}  {{381.1}{96}}         {{380.9}{90}}         {{\textbf{360.9}}{87}}         {{367.7}{90}}         \\
        \fullline {{logistics} {225}}  {{\textbf{0.5}}{0}}            {{\textbf{0.5}}{0}}            {{\textbf{0.5}}{0}}            {{0.8}{0}}            \\
        \fullline {{mystery} {130}}  {{79.6}{6}}           {{71.1}{3}}           {{\textbf{58.6}}{4}}           {{64.6}{6}}           \\
        \midrule
        \fullline {{Total} {1645}}      {{91.5}{133}}         {{90.0}{124}}         {{\textbf{83.0}}{116}}         {{86.5}{125}}         \\
    \end{array}
    \)
    \caption{Single shot solving of PDDL benchmarks.}
\label{tab:1shot-pddl}
    \end{center}
\end{table}

\vspace{-0.05cm}
\begin{table}[h!]\footnotesize
    \begin{center}
    \newcommand{\fullline}[5]{\domname#1&\hspace{5pt}\tablevalue#2 &\hspace{5pt}\tablevalue#3 &\hspace{5pt}\tablevalue#4 &\hspace{5pt}\tablevalue#5 \\}
    \newcommand{\domname}[2]{\textbf{#1} & (#2)}
    \newcommand{\tablevalue}[2]{#1\ (#2)}
    
    \( \def\arraystretch{0.2}
    \begin{array}{lr r r r r }
        \multicolumn{2}{c}{}{} &  \qquad\qquad\textbf{baseline} &  \qquad\qquad\textbf{500}  &   \qquad\qquad\textbf{1000}  &   \qquad\qquad\textbf{1500} \\
        \midrule
        \fullline {{HanoiTower} {20}}   {{160.5}{2}}          {{139.5}{0}}          {{\textbf{137.9}}{0}}          {{143.9}{1}}           \\
        \fullline {{Labyrinth} {20}}    {{\textbf{246.6}}{3}}          {{348.3}{5}}          {{284.8}{4}}          {{296.2}{5}}          \\
        \fullline {{Nomistery} {20}}    {{585.7}{12}}         {{545.4}{11}}         {{\textbf{510.2}}{9}}          {{566.7}{12}}         \\
        \fullline {{Ricochet Robots} {20}}  {{464.5}{9}}          {{\textbf{320.3}}{3}}          {{410.8}{6}}          {{404.9}{5}}          \\
        \fullline {{Sokoban} {20}}      {{458.7}{9}}          {{454.2}{9}}          {{453.8}{9}}          {{\textbf{446.3}}{9}}          \\
        \fullline {{Visit-all} {20}}    {{\textbf{559.1}}{12}}         {{562.5}{12}}         {{560.7}{12}}         {{561.5}{12}}         \\
        \midrule
        \fullline {{Total} {120}}       {{412.5}{47}}         {{395.0}{40}}         {{\textbf{393.0}}{40}}         {{403.3}{44}}         \\
    \end{array}
    \)
    \caption{Single shot solving of ASP benchmarks.}
\label{tab:1shot-asp}
    \end{center}
\end{table}

\begin{table}[h!]\footnotesize
    \begin{center}
    \newcommand{\fullline}[5]{\domname#1&\hspace{5pt}\tablevalue#2 &\hspace{5pt}\tablevalue#3 &\hspace{5pt}\tablevalue#4 &\hspace{5pt}\tablevalue#5 \\}
    \newcommand{\domname}[2]{\textbf{#1} & (#2)}
    \newcommand{\tablevalue}[2]{#1\ (#2)}
    
    \( \def\arraystretch{0.2}
    \begin{array}{lr r r r r }
        \multicolumn{2}{c}{}{} &  \qquad\qquad\textbf{baseline} &  \qquad\qquad\textbf{500}  &   \qquad\qquad\textbf{1000}  &   \qquad\qquad\textbf{1500} \\
        \midrule
        \fullline {{blocks} {20}}     {{1.3}{0}}            {{\textbf{0.7}}{0}}            {{\textbf{0.7}}{0}}            {{\textbf{0.7}}{0}}            \\
        \fullline {{depots} {18}}    {{\textbf{148.6}}{2}}          {{255.9}{3}}          {{188.7}{3}}          {{221.9}{3}}          \\
        \fullline {{driverlog} {9}}  {{108.8}{1}}          {{\textbf{102.1}}{1}}          {{104.8}{1}}          {{108.6}{1}}          \\
        \fullline {{elevator} {20}}  {{\textbf{280.4}}{5}}          {{285.6}{5}}          {{293.8}{5}}          {{304.6}{5}}           \\
        \fullline {{freecell} {16}}  {{900.0}{16}}         {{900.0}{16}}         {{900.0}{16}}         {{900.0}{16}}         \\
        \fullline {{grid} {2}}       {{5.1}{0}}            {{\textbf{3.9}}{0}}            {{4.1}{0}}            {{4.3}{0}}             \\
        \fullline {{gripper} {17}}   {{848.6}{16}}         {{\textbf{847.5}}{16}}         {{849.0}{16}}         {{847.9}{16}}          \\
        \fullline {{logistics} {20}} {{\textbf{225.2}}{5}}          {{225.3}{5}}          {{225.4}{5}}          {{225.3}{5}}          \\
        \fullline {{mystery} {14}}   {{\textbf{321.8}}{5}}          {{321.9}{5}}          {{321.9}{5}}          {{321.9}{5}}          \\
        \midrule
        \fullline {{Total} {136}}    {{\textbf{346.6}}{50}}         {{360.9}{51}}         {{353.6}{51}}         {{359.7}{51}}         \\
    \end{array}
    \)
    \caption{Multi shot solving of PDDL benchmarks.}
    \label{tab:mshot-pddl}
    \end{center}
\end{table}

\begin{table}[h!]\footnotesize
    \begin{center}
    \newcommand{\fullline}[5]{\domname#1&\hspace{5pt}\tablevalue#2 &\hspace{5pt}\tablevalue#3 &\hspace{5pt}\tablevalue#4 &\hspace{5pt}\tablevalue#5 \\}
    \newcommand{\domname}[2]{\textbf{#1} & (#2)}
    \newcommand{\tablevalue}[2]{#1\ (#2)}
    
    \( \def\arraystretch{0.2}
    \begin{array}{lr r r r r }
        \multicolumn{2}{c}{}{} &  \qquad\qquad\textbf{baseline} &  \qquad\qquad\textbf{500}  &   \qquad\qquad\textbf{1000}  &   \qquad\qquad\textbf{1500} \\
        \midrule
        \fullline {{HanoiTower} {20}}   {{\textbf{440.8}}{8}}          {{512.8}{9}}          {{489.4}{9}}          {{498.9}{9}}         \\
        \fullline {{Labyrinth} {20}}    {{633.9}{14}}         {{\textbf{633.8}}{14}}         {{\textbf{633.8}}{14}}         {{633.9}{14}}        \\
        \fullline {{Nomistery} {20}}    {{380.7}{7}}          {{\textbf{363.1}}{6}}          {{381.0}{7}}          {{384.7}{7}}         \\
        \fullline {{Ricochet Robots} {20}}  {{\textbf{521.5}}{11}}         {{523.9}{11}}         {{527.9}{11}}         {{526.0}{11}}        \\
        \fullline {{Sokoban} {20}}      {{\textbf{721.5}}{16}}         {{\textbf{721.5}}{16}}         {{721.9}{16}}         {{722.1}{16}}         \\
        \fullline {{Visit-all} {20}}    {{900.0}{20}}         {{900.0}{20}}         {{900.0}{20}}         {{900.0}{20}}         \\
        \midrule
        \fullline {{Total} {120}}       {{\textbf{599.7}}{76}}         {{609.2}{76}}         {{609.0}{77}}         {{610.9}{77}}        \\
    \end{array}
    \)
    \caption{Multi shot solving of ASP benchmarks.}
    \label{tab:mshot-asp-degm1}
    \end{center}
\end{table}

We expected similar results in the 
single-shot and the multi-shot solving experiments, 
but 
the learning configurations outperformed the baseline in the former, 
while they performed similarly in the latter.
We do not have a clear explanation for this, 
but we can suggest some hypotheses.
In the single-shot experiments 
we can select the best nogood from a larger set 
than in the multi-shot experiments,
which could influence the quality of the learned nogoods.
In addition, having the learned constraints 
from the start could 
positively influence the solver’s heuristic. 
Moreover, the fixed plan length in the single-shot experiments 
could also improve the learning process.
%
%

\section{Conclusion}\label{sec:conclusion}

Conflict-driven constraint learning (CDCL) is the key to the success of modern ASP solvers.
So far, however, ASP solvers could not exploit the temporal structure of dynamic problems.
We addressed this by elaborating upon the generalization of learned constraints in ASP solving for temporal domains.
We started with the definition of temporal logic programs and problems.
We studied the conditions under which learned constraints can be generalized, 
and we identified a class of temporal programs for 
which every learned nogood can be generalized to all time points.
It turns out that many ASP planning encodings fall into this class, 
or can be easily adapted to it. 
We complemented this with a translation 
from temporal programs in general to temporal programs of that class.
Our experimental evaluation show mixed results.
In some settings, the addition of the learned constraints 
results in a consistent improvement of performance, 
while in others the performance is similar to the baseline. 
We plan to continue this experimental investigation in the future. 
Another avenue of future work is to continue the approach sketched
at the end of Section~\ref{sec:approach}, and 
develop a dedicated implementation within an ASP solver 
based on Theorem~\ref{thm:learning:basic}.

\subsubsection*{Acknowledgments.}
This work was supported by DFG grant SCHA 550/15.

\subsubsection*{Competing interests}
The authors declare none.

\bibliographystyle{include/latex-class-tlp/acmtrans} 

\newpage
\appendix
\section{Additional rules of the Blocksworld Example} 
\label{appendix:blocks:action:description}

The following lines specify the action \texttt{unstack(X,Y)},
where `\texttt{B}' is a shorthand for the body `\texttt{block(X), block(Y)}':
\begin{lstlisting}[basicstyle=\small\ttfamily,mathescape=true]
        action(unstack(X,Y)) :- B.
pre(unstack(X,Y), handempty) :- B.
pre(unstack(X,Y),  clear(X)) :- B.
pre(unstack(X,Y),   on(X,Y)) :- B.
add(unstack(X,Y),  clear(Y)) :- B.
add(unstack(X,Y),holding(X)) :- B.  
del(unstack(X,Y),   on(X,Y)) :- B.  
del(unstack(X,Y), handempty) :- B.
del(unstack(X,Y),  clear(X)) :- B.
\end{lstlisting}

The following lines specify the actions \texttt{pick\_up(X)} and \texttt{out\_down(X)},
where `\texttt{B}' is a shorthand for the body `\texttt{block(X)}':
\begin{lstlisting}[basicstyle=\small\ttfamily,mathescape=true]
        action(pick_up(X))  :- B.
pre(pick_up(X), handempty)  :- B.
pre(pick_up(X),  clear(X))  :- B.
pre(pick_up(X),ontable(X))  :- B.
add(pick_up(X),holding(X))  :- B.
del(pick_up(X), handempty)  :- B.
del(pick_up(X),  clear(X))  :- B.
del(pick_up(X),ontable(X))  :- B.

        action(put_down(X)) :- B.
pre(put_down(X),holding(X)) :- B.
add(put_down(X),  clear(X)) :- B.
add(put_down(X), handempty) :- B.
add(put_down(X),ontable(X)) :- B.
del(put_down(X),holding(X)) :- B.
\end{lstlisting}

\section{Program translations from our conference paper}\label{sec:translations}
We present the program translations from our conference paper~\cite{roscst22a}.
We think they can still be of interest, although the 
new translation in Section~\ref{sec:cyclic:encoding} has some 
advantages, as it is  
easier to understand 
and generalizes the nogoods to the complete interval $[0,n]$ instead of $[1,n]$. 
The size of both translations is linear on the size of the input programs.

%
Given some temporal logic program $\Pi$, 
we say that the rules $r \in \Pi$ such that $\atom{r}\subseteq\mathcal{A}$ are static,
and otherwise we say that they are dynamic.

We start with a simple translation $\trb{}$ that works for temporal 
programs where all dynamic rules are integrity constraints.
Later, we show that all temporal programs
\mbox{can be translated to this form.}

We say that a temporal logic program $\Pi$ over $\mathcal{A}$ is in
\emph{previous normal form} (\pnf{}) if 
$\atom{\Pi\setminus\integr{\Pi}}\cap\mathcal{A}'=\emptyset$, 
and that a temporal logic problem \tpb{\Pi}{I}{F} over $\mathcal{A}$ is in \pnf{}
if $\Pi$ is in \pnf{}.
Given a temporal logic program $\Pi$ over $\mathcal{A}$,
let $\integrprev{\Pi}$ denote the set 
$\{ r \mid r \in \integr{\Pi}, \atom{r}\cap\mathcal{A}'\neq\emptyset\}$ 
of dynamic integrity constraints of $\Pi$.
Note that if $\Pi$ is in \pnf{}, 
then the dynamic rules of $\Pi$ 
belong to $\integrprev{\Pi}$.
The translation $\trbdoprev{\Pi}$ tags the rules in $\integrprev{\Pi}$ 
with a new atom $\trbsymbol{}$,
that does not belong to $\mathcal{A}$ or $\mathcal{A}'$,
and extends the program with a choice rule for $\trbsymbol{}$.
%
Formally, by $\trbdoprev{\Pi}$ we denote the temporal logic program:
\[
\Pi\setminus\integrprev{\Pi} \cup 
\big\{ \{ \trbsymbol{}\}\leftarrow \big\} \cup
\{ \bot \leftarrow \body{r} \cup \{\trbsymbol{}\} \mid r \in \integrprev{\Pi}\}.
\]
It is easy to see that when $\trbsymbol{}$ is chosen to be true,
$\trbdoprev{\Pi}$ generates the same transitions as $\Pi$.
Then, we can solve temporal programs $\tpb{\Pi}{I}{F}$
by solving temporal problems $\tpb{\trbdoprev{\Pi}}{I}{F}$,
if we consider only solutions that make $\trbsymbol{}$ true at all steps after the initial one.
For convenience, we consider only the case where $\trbsymbol{}$ is false.
This means that, as in Section~\ref{sec:cyclic:encoding}, 
we focus on the \trbnormal{} solutions.
%
%
The next proposition states the relation between 
these \trbnormal{} solutions and the original solutions using $\Pi$.
%

\begin{proposition}\label{prop:trb}
Let $\mathcal{T}_1=\tpb{\Pi}{I}{F}$ 
and 
let $\mathcal{T}_2=$ $\tpb{\trbdoprev{\Pi}}{I}{F}$
be temporal logic problems.
There is a one-to-one correspondence between the solutions to $\mathcal{T}_1$ and 
the \trbnormal{} solutions to $\mathcal{T}_2$.
\end{proposition}

The call 
$\cdnlasp{}(\gen{\trbdoprev{\Pi}}{n},$ $
            \at{I}{0}\cup\at{F}{n}\cup
            \{\Flit{\atomat{\trbsymbol{}}{0}},\Tlit{\atomat{\trbsymbol{}}{1}},\ldots,\Flit{\atomat{\trbsymbol{}}{n}}\})$
computes \trbnormal{} solutions to $\mathcal{T}_2$,
enforcing the correct value for $\trbsymbol$ at every time point using assumptions.
The solutions to the original problem $\mathcal{T}_1$ 
can be extracted from the \trbnormal{} solutions,
after deleting the atoms in $\btw{\{\trbsymbol{}\}}{1}{n}$.

We turn now our attention to the resolvents $\delta$ of
the set of nogoods $\btw{\Psi_{\trbdoprev{\Pi}}}{1}{n}$
used by the procedure $\cdnlasp{}$.
As we will see, just by looking at these resolvents $\delta$,
we can approximate the specific interval $[i,j]\subseteq[1,n]$ of the 
nogoods \mbox{that were used to prove them}.

To this end,
we say that the nogoods containing literals of different steps are dynamic nogoods,
and they are static nogoods otherwise.
%
All dynamic nogoods in $\btw{\Psi_{\trbdoprev{\Pi}}}{1}{n}$
come from the instantiation of some dynamic integrity constraint
$\{ \bot \leftarrow \body{r} \cup \{\trbsymbol{}\} \mid r \in \integrprev{\Pi}\}$
at some time step $i$ and, therefore,
they contain some literal of the form $\atomat{\Tlit{\trbsymbol}}{i}$.
On the other hand,
in $\btw{\Psi_{\trbdoprev{\Pi}}}{1}{n}$
there are no literals of the form $\Flit{\atomat{\trbsymbol{}}{i}}$.
Hence, the literals $\atomat{\Tlit{\trbsymbol{}}}{i}$
occurring in the dynamic nogoods can never be resolved away. 
Then, if some dynamic nogood is used to prove a learned nogood $\delta$,
the literal $\atomat{\Tlit{\trbsymbol{}}}{i}$ occurring in that dynamic nogood must belong to $\delta$.
This means that
the literals $\atomat{\Tlit{\trbsymbol{}}}{i}$ from a learned nogood $\delta$
tell us exactly the steps $i$ of the dynamic nogoods that \mbox{have been used to prove $\delta$}.

Observe now that two nogoods 
$\delta_1 \in \at{\Psi_{\trbdoprev{\Pi}}}{i}$ and
$\delta_2 \in \at{\Psi_{\trbdoprev{\Pi}}}{{i+1}}$
can only
be resolved if $\delta_2$ is a dynamic nogood.
Otherwise, the nogoods would have no opposite literals to resolve.
Applying the same reasoning,
if two nogoods
$\delta_1 \in \at{\Psi_{\trbdoprev{\Pi}}}{i}$ and
$\delta_2 \in \at{\Psi_{\trbdoprev{\Pi}}}{j}$, such that $i<j$,
are part of the same resolution proof of a learned nogood $\delta$,
then the proof must also contain some dynamic nogoods from each step in the interval $[i+1,j]$.
Therefore,
the learned nogood $\delta$ must contain the literals
$\btw{\{\Tlit{\trbsymbol{}}\}}{i+1}{j}$.

This implies that, 
given the literals $\btw{\{\Tlit{\trbsymbol{}}\}}{k}{j}$ occurring in a learned nogood $\delta$,
we can infer the following about the nogoods 
from $\btw{\Psi_{\trbdoprev{\Pi}}}{1}{n}$ used to prove $\delta$: 
dynamic nogoods from all the steps $[k,j]$ were used to prove $\delta$,
possibly some static nogoods of the step $k-1$ were used as well, and
no nogoods from other steps were used in the proof.
%
It is possible that some static nogoods at steps $[k,j]$ were also used, 
but no dynamic nogoods at $k-1$ could be used, 
since otherwise $\delta$ should contain the literal
$\atomat{\Tlit{\trbsymbol{}}}{{k-1}}$.

We formalize this with the function \step{\delta},
that approximates the specific interval $[i,j]$ of the nogoods that were used to prove $\delta$:
if $\delta$ contains some literal of the form ${\Tlit{\atomat{\trbsymbol{}}{i}}}$ for $i \in [1,n]$,
then $\step{\delta}$ is the set of steps
$\{j-1, j \mid \Tlit{\atomat{\trbsymbol}{j}}\in \delta\}$.
%
%
For example, if $\delta$ is $\{\atomat{\Tlit{a}}{3}, \atomat{\Tlit{\trbsymbol}}{3}\}$
then the value of $\step{\delta}$ is $\{2,3\}$.
It is clear that $\delta$ was derived using some dynamic nogood of step $3$,
that added the literal \atomat{\Tlit{\trbsymbol}}{3}.
And it could also happen that some static nogood of step $2$ was used, 
but we are uncertain about it.
That is why we say that $\mathit{step}$ is an approximation.
To continue, note that it can also be that $\delta$ 
has no literals of the form ${\Tlit{\atomat{\trbsymbol{}}{i}}}$.
In this case, $\delta$
must be the result of resolving some static nogoods of a single time step, 
and we can extract that time step from the
unique time step of the literals occurring in the nogood.
Hence, in this case we define 
$\step{\delta}$ as $\stepsimple{\delta}$. 
%
For example, $\step{\{\atomat{\Tlit{c}}{2}, \atomat{\Tlit{d}}{2}\}}=\{2\}$. 
With this, 
we can generalize a nogood $\delta$ to the shifted nogoods $\shift{\delta}{t}$
whose $\mathit{step}$ value fits in the interval $[1,n]$.
We state this precisely in the next theorem.
Observe that it excludes the shifted nogoods $\shift{\delta}{t}$ that contain the literal $\Tlit{\atomat{\trbsymbol}{1}}$,
since in that case $\step{\shift{\delta}{t}}$ contains the step $0 \notin [1,n]$.
This makes sense because to prove $\shift{\delta}{t}$ 
we could need some static nogoods at step $0$, 
and they do not belong to $\btw{\Psi_{\trbdoprev{\Pi}}}{1}{n}$.
%

\begin{theorem}\label{thm:learning:zero}
Let $\Pi$ be a temporal logic program in \pnf{}, 
and $\delta$ be a resolvent of
$\btw{\Psi_{\trbdoprev{\Pi}}}{1}{m}$ for some $m\geq 1$.
Then, for every $n\geq 1$, 
the set of nogoods $\btw{\Psi_{\trbdoprev{\Pi}}}{1}{n}$ entails the 
generalization \[\{ \shift{\delta}{t} \mid \step{\shift{\delta}{t}} \subseteq [1,n]\}.\]
\end{theorem}

\begin{figure}
    \centering
    \includegraphics[width=0.35\textwidth]{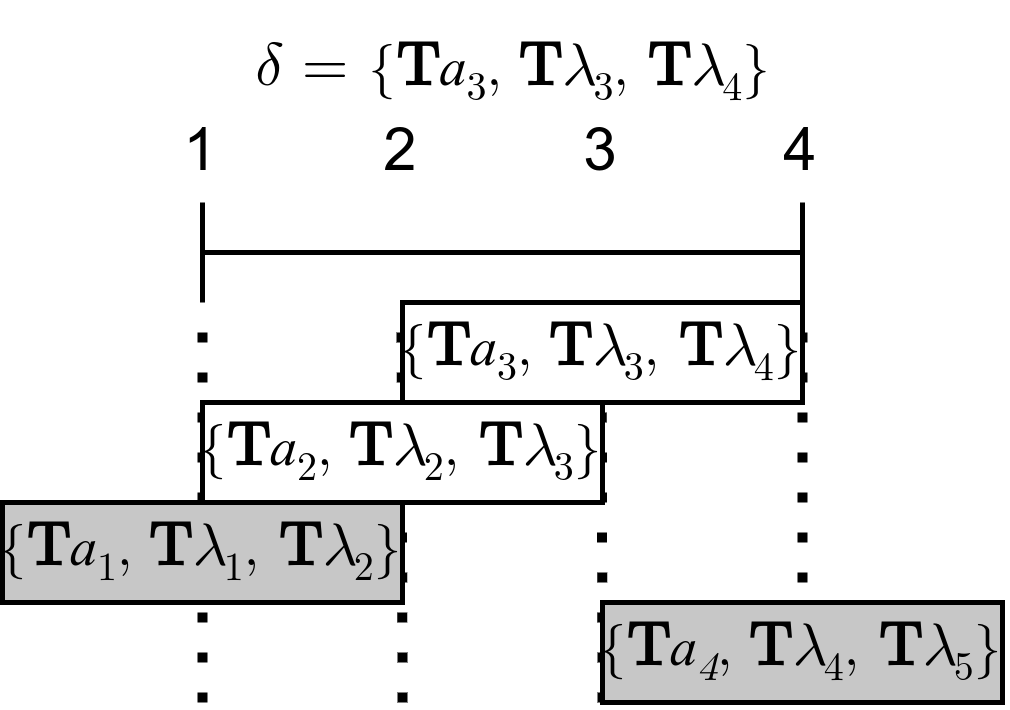}

    \caption{Representation of different shifted versions of the nogood
    $\delta=\{
    \Tlit{\atomat{a}{3}}, 
    \Tlit{\atomat{\trbsymbol{}}{3}},
    \Tlit{\atomat{\trbsymbol{}}{4}}
    \}$.
    The surrounding rectangles cover the interval of their $\mathit{step}$ value. 
    For example, the rectangle of 
    $\{
    \Tlit{\atomat{a}{2}}, 
    \Tlit{\atomat{\trbsymbol{}}{2}},
    \Tlit{\atomat{\trbsymbol{}}{3}}
    \}$
    covers the interval $[1,3]$ because $\mathit{step}(
    \{
    \Tlit{\atomat{a}{2}}, 
    \Tlit{\atomat{\trbsymbol{}}{2}},
    \Tlit{\atomat{\trbsymbol{}}{3}}
    \}
    ) = [1,3]$.
}
    \label{fig:shifting-ex-lambda}
\end{figure}

\begin{example}
Consider the call $\cdnlasp{}(\gen{\trbdoprev{\Pi_1}}{4},$ $\emptyset)$,
similar to the one that we have seen before using the original program $\Pi_1$.
The nogoods $\btw{\Psi_{\trbdoprev{\Pi_1}}}{1}{n}$ 
are the same as those in $\btw{\Psi_{{\Pi_1}}}{1}{n}$, 
except that every dynamic nogood contains one instantiation of
the literal $\Tlit{{\trbsymbol{}}}$.
Instead of learning the nogood 
$\{\Tlit{\atomat{a}{3}}\}$ 
the algorithm would learn the nogood 
$\delta=\{
\Tlit{\atomat{a}{3}}, 
\Tlit{\atomat{\trbsymbol{}}{3}},
\Tlit{\atomat{\trbsymbol{}}{4}}
\}$.
Then, applying part (i) of Theorem~\ref{thm:learning:zero} 
the nogood $\delta$ can be generalized to 
$\shift{\delta}{-1}=
\{
\Tlit{\atomat{a}{2}}, 
\Tlit{\atomat{\trbsymbol{}}{2}},
\Tlit{\atomat{\trbsymbol{}}{3}}
\}
$,
but not to 
$\shift{\delta}{1}=
\{
\Tlit{\atomat{a}{4}}, 
\Tlit{\atomat{\trbsymbol{}}{4}},
\Tlit{\atomat{\trbsymbol{}}{5}}
\}
$
or to $\shift{\delta}{-2}=
\{
\Tlit{\atomat{a}{1}}, 
\Tlit{\atomat{\trbsymbol{}}{1}},$ $
\Tlit{\atomat{\trbsymbol{}}{2}}
\}$ (see Figure~\ref{fig:shifting-ex-lambda}).
%
%
\end{example}

The next step is to show how temporal programs in general can be translated to \pnf{} form.
For this, given a temporal logic program $\Pi$ over $\mathcal{A}$,
let $\traatom{\mathcal{A}}=\{ \traatom{a} \mid a \in \mathcal{A}\}$,
and assume that this set is disjoint from $\mathcal{A}$ and $\mathcal{A}'$.
The translation \trado{\Pi} consists of two parts.
The first part consists of the result of replacing 
in $\Pi$ every atom $\prev{a}\in\mathcal{A}'$ 
by its corresponding new atom $\traatom{a}$.
The second part consists of the union of the rules
\[
\{
\{ \traatom{a} \} \leftarrow;
\bot \leftarrow a', \neg \traatom{a};
\bot \leftarrow \neg a', \traatom{a}
\}
\]
for every $a \in \mathcal{A}$.
The idea of the translation is that the atoms $a' \in \mathcal{A}'$ 
are confined to integrity constraints by replacing them 
by new atoms $\traatom{a}\in\traatom{\mathcal{A}}$, 
whose truth value is completely determined by the corresponding $a' \in \mathcal{A}'$ atoms
\mbox{by means of the last set of rules.}
%

\begin{proposition}\label{prop:trstar_pnf}
For any temporal logic program ${\Pi}$, 
the program $\trado{\Pi}$ is in \pnf{}.
\end{proposition}

The solutions to temporal problems with ${\Pi}$ are the same
as the solutions to the same temporal problems with $\trado{\Pi}$
where the atoms $\at{\traatom{a}}{i}$ are false at $i=0$
and have the truth value of $\at{a}{i-1}$ at the other time steps $i$.
Just like before, when we use this translation, 
we have to add to \cdnlasp{} the correct assumptions to fix
the value of the $\traatom{a}$ atoms at step $0$.
%

\begin{proposition}\label{prop:p_eq_pstar_solutions}
Let $\mathcal{T}_1=\tpb{\Pi}{I}{F}$ and
let $\mathcal{T}_2=\tpb{\trado{\Pi}}{I}{F}$
be temporal logic problems.
There is a one-to-one correspondence between the solutions to $\mathcal{T}_1$ and 
the solutions to $\mathcal{T}_2$ that do not contain any atom 
$\traatom{p}\in\traatom{\mathcal{P}}$ at step $0$.
\end{proposition}

This proposition allows us to replace any temporal program $\Pi$
by a temporal program $\trado{\Pi}$ in \pnf{}.
We can then apply the translation $\trb{}$ and benefit from Theorem~\ref{thm:learning:zero}.
In fact, we can go one step further, 
and apply the nogoods learned with the program 
$\trbdoprev{\trado{\Pi}}$
directly to the original problem with $\Pi$.
We make this claim precise in the next theorem. 
%
We extend our definition of the simplification of a nogood $\delta$, 
$\simp{\delta}$,
to accommodate literals over $\mathcal{A}\cup\traatom{\mathcal{A}}$. 
That is, $\simp{\delta}$ is the nogood
$\{\Vlit{\atomat{a}{i}}\mid\Vlit{\atomat{a}{i}}\in\delta, \Vlit{}\in\{\Tlit{},\Flit{}\}, a \in \mathcal{A}\}
 \cup
 \{\Vlit{\atomat{a}{{i-1}}}\mid\Vlit{\atomat{\traatom{a}}{i}}\in\delta, \Vlit{}\in\{\Tlit{},\Flit{}\}, \traatom{a} \in \traatom{\mathcal{A}}\}$
that results from skipping the $\atomat{\trbsymbol}{i}$ literals
of $\delta$,
and replacing the atoms $\atomat{\traatom{a}}{i}$  
by their corresponding atoms $\atomat{a}{{i-1}}$.
%
%

\begin{theorem}\label{thm:learning:one}
Let $\Pi$ be a temporal logic program,
and $\delta$ be a resolvent of
$\btw{\Psi_{\trbdoprev{\trado{\Pi}}}}{1}{m}$ for some $m\geq 1$.
Then, for every $n\geq 1$, 
the set of nogoods $\btw{\Psi_{{\Pi}}}{1}{n}$ entails 
the generalization 
\[\{\simp{\shift{\delta}{t}} \mid \step{\shift{\delta}{t}}\subseteq[1,n]\}.\]
\end{theorem}

%

\section{Proofs}\label{sec:proofs}
\setcounter{theorem}{1}
\setcounter{proposition}{0}


\begin{lemma}\label{lemma:timepoint-application}
    For any temporal logic program $\Pi$, $\at{\Sigma_{\Pi}}{n} = \Sigma_{\at{\Pi}{n}}$.
\end{lemma}

\begin{proof}[Proof]

\begin{enumerate}
    \item \textbf{Constraints}\label{constraints}
    
    \item[] For any constraint $c$ of the form $ \bot \leftarrow a_1, ...a_m, \nlit{b_{m+1}}, ..., \nlit{b_l}$ it holds that
    \begin{align*}
        \at{c}{n} =&\ \  \bot \leftarrow \at{a_1}{n}, ...\at{a_m}{n}, \at{\nlit{b_{m+1}}}{n}, ..., \at{\nlit{b_{l}}}{n} \\
        \Sigma_{\at{c}{n}} =&\ \   \at{\Tlit{a_1}}{n}, ...\at{\Tlit{a_m}}{n}, \at{\Flit{b_{m+1}}}{n}, ..., \at{\Flit{b_{l}}}{n} \\
        \Sigma_{c} =&\ \  \Tlit{a_1}, ...\Tlit{a_m}, \Flit{b_{m+1}}, ..., \Flit{b_{l}} \\
        \at{\Sigma_{c}}{n} =&\ \   \at{\Tlit{a_1}}{n}, ...\at{\Tlit{a_m}}{n}, \at{\Flit{b_{m+1}}}{n}, ..., \at{\Flit{b_{l}}}{n} \\
    \end{align*}
    \item[] Since all constraints have the form of $c$, we can conclude that $\Sigma_{\at{c}{n}} = \at{\Sigma_{c}}{n}$ for any constraint.

    \item \textbf{Body}\label{Body}
        \item[] For any body $B$ of the form $\{ a_1, ...a_m, \nlit{b_{m+1}}, ..., \nlit{b_l} \}$ it holds that
    \begin{align*}
        \at{B}{n} =&\ \  \at{a_1}{n}, ...\at{a_m}{n}, \at{\nlit{b_{m+1}}}{n}, ..., \at{\nlit{b_{l}}}{n} \\
        \Sigma_{\at{B}{n}} = &\ \ \{ \{ \Tlit{\at{B}{n}}, \Flit{\at{a_1}{n}} \},...,\{ \Tlit{\at{B}{n}}, \Flit{\at{b_l}{n}} \} \} \cup \{ \Flit{\at{B}{n}}, \Tlit{\at{a_1}{n}},...,\Flit{\at{b_l}{n}} \} \\
        \Sigma_{B} =&\ \  \{ \{ \Tlit{B}, \Flit{a_1} \},...,\{ \Tlit{B}, \Flit{b_l} \} \} \cup \{ \Flit{B}, \Tlit{a_1},...,\Flit{b_l} \} \\
        \at{\Sigma_{B}}{n} =&\ \  \{ \{ \Tlit{\at{B}{n}}, \Flit{\at{a_1}{n}} \},...,\{ \Tlit{\at{B}{n}}, \Flit{\at{b_l}{n}} \} \} \cup \{ \Flit{\at{B}{n}}, \Tlit{\at{a_1}{n}},...,\Flit{\at{b_l}{n}} \} \\
    \end{align*}

        \item[] Since all bodies have the form of $B$, we can then conclude that $\Sigma_{\at{B}{n}} = \at{\Sigma_{B}}{n}$.

    \item \textbf{A set of rules with the same head}\label{rules}
        \item[] For any set of rules with the same head $\Pi$ of the form $\{a \leftarrow B_1 ,..., a \leftarrow B_l\}$ 
        where $B_i$ are bodies and $\Sigma_{B}$ is the set of all body nogoods it holds that

    \begin{align*}
        \at{\Pi}{n} =&\ \  \{\at{a}{n} \leftarrow \at{B_1}{n} ,..., \at{a}{n} \leftarrow \at{B_l}{n}\} \\
        \Sigma_{\at{\Pi}{n}} =&\ \  \{ \{ \Flit{\at{B_1}{n}},...,\Flit{\at{B_l}{n}},\Tlit{\at{a}{n}} \}, \{ \Tlit{\at{B_1}{n}},\Flit{\at{a}{n}} \},...,\{ \Tlit{\at{B_l}{n}},\Flit{\at{a}{n}} \} \} \cup \Sigma_{B} \\
        \Sigma{\Pi} =&\ \  \{ \{ \Flit{B_1},...,\Flit{B_l},\Tlit{a} \}, \{ \Tlit{B_1},\Flit{a} \},...,\{ \Tlit{B_l},\Flit{a} \}  \cup \Sigma_{B} \} \\
        \at{\Sigma_{\Pi}}{n} =&\ \  \{ \{ \Flit{\at{B_1}{n}},...,\Flit{\at{B_l}{n}},\Tlit{\at{a}{n}} \}, \{ \Tlit{\at{B_1}{n}},\Flit{\at{a}{n}} \},...,\{ \Tlit{\at{B_l}{n}},\Flit{\at{a}{n}} \} \} \cup \Sigma_{B} \\
    \end{align*}

        \item[] Since body nogoods are also equal, we can conclude that $\Sigma_{\at{\Pi}{n}} = \at{\Sigma_{\Pi}}{n}$.

    \item \textbf{Choice rules}\label{choices}
        \item[] Since choice rule nogoods are a subset of normal rule nogoods, we can conclude that $\Sigma_{\at{c}{n}} = \at{\Sigma_{c}}{n}$ for any choice rule $c$.

    \item \textbf{Loops}\label{loops}
        \item[] For any set of rules $\Pi$ forming a loop of the form $a_1 \leftarrow a_2, B_1, ..., a_{n} \leftarrow a_1, B_n$
         with external Bodies for (some) $a_i$ being labeled $E_i$ and $\Sigma_{R}$ is the set of all rule nogoods it holds that

    \begin{align*}
        \at{\Pi}{n} =&\ \  \{\at{a_1}{n} \leftarrow \at{a_2}{n}, \at{B_1}{n}, ..., \at{a_{n}}{n} \leftarrow \at{a_1}{n}, \at{B_n}{n}\}  \\
        \Sigma_{\at{\Pi}{n}} =&\ \  \{ \{ \Tlit{\at{a_1}{n}}, \Flit{\at{E_{i_1}}{n}},...,\Flit{\at{E_{i_m}}{n}} \},...,\{ \Tlit{\at{a_n}{n}}, \Flit{\at{E_{i_1}}{n}},...,\Flit{\at{E_{i_m}}{n}} \} \} \cup \Sigma_{R}  \\
        \Sigma_{\Pi} =&\ \  \{ \{ \Tlit{a_1}, \Flit{E_{i_1}},...,\Flit{E_{i_m}} \},...,\{ \Tlit{a_n}, \Flit{E_{i_1}},...,\Flit{E_{i_m}} \} \} \cup \Sigma_{R} \\
        \at{\Sigma_{\Pi}}{n} =&\ \  \{ \{ \Tlit{\at{a_1}{n}}, \Flit{\at{E_{i_1}}{n}},...,\Flit{\at{E_{i_m}}{n}} \},...,\{ \Tlit{\at{a_n}{n}}, \Flit{\at{E_{i_1}}{n}},...,\Flit{\at{E_{i_m}}{n}} \} \} \cup \Sigma_{R}  \\
    \end{align*}

        \item[] Since rule nogoods are also equal, we can conclude that $\Sigma_{\at{\Pi}{n}} = \at{\Sigma_{\Pi}}{n}$.
\end{enumerate}

From items \ref{constraints}, \ref{Body}, \ref{rules}, \ref{choices}, \ref{loops} we can say that for any program $\Pi$, $\Sigma_{\at{\Pi}{n}} = \at{\Sigma_{\Pi}}{n}$.

\end{proof}

\begin{proof}[Proof]

Let $C = \{\{a'\} \leftarrow | a \in \mathcal{A}\}$ where $\mathcal{A}$ is the set of atoms ocurring in $\Pi$.
Since all the rules in $C$ are choice rules with empty bodies, $\Sigma_{\at{C}{n}}$ is comprised of nogoods of the form $\{ \Tlit{\at{\prev{a}}{n}}, \Flit{\emptyset} \}$.
Given that $\Flit{\emptyset}$ is always false the nogoods can be safely removed.
Hence, for any program $\Pi$ it holds that $\Sigma_{\at{C}{n}} \cup \Sigma_{\Pi} = \Sigma_{\Pi}$.

For a given temporal logic program $\Pi$ we can define $\trans{\Pi} = C \cup \Pi$.
Additionally,
$\gen{\Pi}{n}$ can be defined as $\at{C}{1} \cup \btw{\Pi}{1}{n}$,
which means that
\begin{align*}
    \Sigma_{\gen{\Pi}{n}} &= \Sigma_{\at{C}{1}} \cup \Sigma_{\btw{\Pi}{1}{n}} \\
                          &= \Sigma_{\at{C}{1}} \cup \Sigma_{\at{\Pi}{1}} \cup ... \cup \Sigma_{\at{\Pi}{n}} \\
                          &= \Sigma_{\at{\Pi}{1}} \cup ... \cup \Sigma_{\at{\Pi}{n}} & (\text{deleting choice nogoods})
\end{align*}

Also, 
\begin{align*}
    \btw{\Psi_{\Pi}}{1}{n} &= \btw{\Sigma_{\trans{\Pi}}}{1}{n}  \\
                           &= \at{\Sigma_{\trans{\Pi}} }{1} \cup ... \cup \at{\Sigma_{\trans{\Pi}}}{n} \\
                           &= \at{\Sigma_{C}}{1} \cup \at{\Sigma_{\Pi}}{1} \cup ... \cup \at{\Sigma_{C}}{n} \cup \at{\Sigma_{\Pi}}{n} \\
                           &= \at{\Sigma_{\Pi}}{1} \cup ... \cup \at{\Sigma_{\Pi}}{n} & (\text{deleting choice nogoods})\\
                           &= \Sigma_{\at{\Pi}{1}} \cup ... \cup \Sigma_{\at{\Pi}{n}} & (\text{lemma} \ref{lemma:timepoint-application}) \\
                           &= \Sigma_{\gen{\Pi}{n}}
\end{align*}

\end{proof}


\begin{proof}[Proof]
Let $\mathcal{A}$ be the set of atoms occurring in $\Pi$.

By Proposition~\ref{thm:nogoods:temp:eq} 
a solution for the set of nogoods $\btw{\Psi_\Pi}{1}{n}$ is a solution for $\Sigma_{\gen{\Pi}{n}}$.
A solution for $\Sigma_{\gen{\Pi}{n}}$ is a stable model for the generator program $\gen{\Pi}{n}$.
Since a stable model of $\gen{\Pi}{n}$ consistent with $I$ and $F$ is a solution of \tpb{\Pi}{I}{F},
then a solution $S$ of $\btw{\Psi_\Pi}{1}{n}$ consistent with some $I$ and $F$,
the pair $(X,n)$ where $X=\Tass{S} \cap \btw{\mathcal{A}}{1}{n}$ is a solution for \tpb{\Pi}{I}{F}.

Let $(X,n)$ be a solution to the temporal logic problem \tpb{\Pi}{I}{F}.
By definition, $X$ is a stable model of \gen{\Pi}{n} consistent with $I$ and $F$.
Since a stable model of \gen{\Pi}{n} is a solution of $\Sigma_{\gen{\Pi}{n}}$ which is a solution of $\btw{\Psi_{\Pi}}{1}{n}$ (by Proposition~\ref{thm:nogoods:temp:eq}),
it follows that $S = \{\Tlit{a} | a \in X\} \cup \{\Flit{a} | a \in \btw{\mathcal{A}}{0}{n} \setminus X\}$
is a solution for the temporal logic program $\btw{\Psi_\Pi}{1}{n}$ such that $\at{I}{0} \cup \at{F}{n} \subseteq S$.
\end{proof}

\begin{lemma}\label{lemma:shift-resolution}
      For any resolvent $\delta$ of $\btw{\Psi}{i}{j}$ it holds that $\shift{\delta}{t}$ is a resolvent of $\btw{\Psi}{i+t}{j+t}$
\end{lemma}

\begin{proof}[Proof]

Recall that if a nogood is a resolvent of $\btw{\Psi}{i}{j}$ then it must have a resolution proof $\mathcal{T}$ where every nogood $\delta_i \in \mathcal{T}$
is either entailed by $\btw{\Psi}{i}{j}$
or the result of resolving some $\delta_j$ and $\delta_k$
where $j < k < i$ and both $\delta_j$ and $\delta_k$ are entailed by $\btw{\Psi}{i}{j}$.
Additionally, for a resolution proof $\mathcal{T} = \delta_1,  \ldots \delta_n$ the result is $\delta_n$.
Finally, note that if a nogood $\delta \in \btw{\Psi}{i}{j}$ then $\shift{\delta}{t} \in \btw{\Psi}{i+t}{j+t}$

We now prove the lemma by induction. Let $\mathcal{T}$ be the resolution proof of a nogood $\delta$ that is entailed by $\btw{\Psi}{i}{j}$.

\textbf{\textit{Induction base 1:}} If $\mathcal{T} = \delta$ then $\delta \in \btw{\Psi}{i}{j}$ holds and, trivially, $\shift{\delta}{t} \in \btw{\Psi}{i+t}{j+t}$.

\smallskip

\textbf{\textit{Induction base 2:}} If $\mathcal{T} = \delta_1, \delta_2$ then, since there less than two nogoods before $\delta_1$ and $\delta_2$ then both must be in $\btw{\Psi}{i}{j}$.
Consequently, $\shift{\delta_1}{t} \in \btw{\Psi}{i+t}{j+t}$ and $\shift{\delta_2}{t} \in \btw{\Psi}{i+t}{j+t}$.

\smallskip

\textbf{\textit{Induction step n:}} Let $\mathcal{T} = \delta_1, ... , \delta_n$ be a resolution proof for nogood $\delta_n$.
If $\delta_n \in \btw{\Psi}{i}{j}$ then, trivially, $\shift{\delta_n}{t} \in \btw{\Psi}{i+t}{j+t}$.
If $\delta_n \notin \btw{\Psi}{i}{j}$ then we know by induction that all $\delta_i$ with $0 \leq i \leq n-1$ are entailed by $\btw{\Psi}{i}{j}$. Since $\delta_n \notin \btw{\Psi}{i}{j}$
then there are some $\delta_k$ and $\delta_l$ where $k < l < i$ that resolve to $\delta_n$.
By induction, $\shift{\delta_k}{t}$ and $\shift{\delta_l}{t}$ are entailed by $\btw{\Psi}{i+t}{j+t}$.
Consequently, $\shift{\delta_n}{t}$ is entailed by $\btw{\Psi}{i+t}{j+t}$.

\end{proof} 


\begin{proof}[Proof]

Let $\delta$ be a resolvent of $\btw{\Psi}{i}{j}$. 
Then the shifted nogood $\shift{\delta}{t}$ is entailed by $\btw{\Psi}{i+t}{j+t}$ (Lemma \ref{lemma:shift-resolution}).
Let $t$ be a value where $[i+t,j+t] \subseteq [1,n]$ holds, then $\shift{\delta}{t}$ is entailed by $\btw{\Psi}{1}{n}$ 
since $\btw{\Psi}{i+t}{j+t} \subseteq \btw{\Psi}{1}{n}$.

\end{proof}


\begin{proof}[Proof]

The solution to the temporal logic problem \tpb{\Pi}{I}{F} is a stable model of \gen{\Pi}{n} consistent with $I[0]$ and $F[0]$.
We can split \gen{\Pi}{n} as follows:
Let $C = \{\{a\} \leftarrow | a \in \mathcal{A}\}$
$$ C[0] \cup \Pi[1] \cup ... \cup \Pi[n]$$
where $\mathcal{A}$ is the set of atoms ocurring in $\Pi$.

From the Splitting Set Theorem \cite{liftur94a} it follows that we can build every stable model $X$ for \gen{\Pi}{n} as follows:
\begin{align*}
    X_0 & \text{ is a stable model of } C[0] \\
    X_1 & \text{ is a stable model of } \Pi[1] \cup X_0 \\
    ... \\
    X_n & \text{ is a stable model of } \Pi[n] \cup X_{n-1} \\
\end{align*}
where $X_n$ is a stable model of \gen{\Pi}{n}.

It is easy to see that every $X_{i-1} \subseteq X_i$ where $1 \leq i \leq n$. 
Let $s_i = X_i \cap \mathcal{A}[i]$ with $0 \leq i \leq n$, 
then program $\Pi[i] \cup X_{i-1}$ can be rewritten as $\Pi[i] \cup s_{i-1} \cup ... \cup s_{0}$.

$M$ is a stable model of $\Pi[i] \cup s_{i-1} \cup ... \cup s_{0}$ iff $M$ has the form $M_i \cup s_{i-2} \cup ... \cup s_{0}$ 
for some stable model $M_i$ of $\trans{\Pi}[i]$ such that $s_{i-1} = M_i \cap P[i-1]$.
This follows from the fact that $\trans{\Pi}[i] = \Pi[i] \cup C[i-1]$.
Following the Splitting Set Theorem,
we can build a stable model for $\trans{\Pi}[i]$ by first getting a model $S$ for $C[i-1]$ and then a model for $\Pi[i] \cup S$.
Since $C$ is comprised of choice rules for all atoms, then the assignment formed from $s_{i-1}$ is a stable model of $C[i-1]$.
Thus, a stable model of $\Pi[i] \cup s_{i-1}$ is a stable model of $\trans{\Pi}[i]$.

This also means that $s_i = M_i \cap P[i]$ is a state in $\graph{\Pi}$ and that $\langle s_i, s_{i-1} \rangle$ is an edge.

Consequently, we can say that the states $s_0,...,s_n$ form a path in the graph $\graph{\Pi}$.
Finally, for any stable model of \gen{\Pi}{n} consistent with $I[0]$ and $F[n]$,
then the states $s_0,...,s_n$ form a path in $\graph{\Pi}$ and $I[0]$ and $F[n]$ are consistent with $s_0$ and $s_n$ respectively.

\end{proof} 

\begin{proof}[Proof]


By Theorem \ref{thm:problemgraph} when $I$ and $F$ are empty,
the solutions to $\btw{\Psi_{\Pi}}{i}{j}$ correspond to paths of length $j-i+1$ in $\graph{\Pi}$.
This means that no path of this length violates a nogood in $\btw{\Psi_{\Pi}}{i}{j}$.

Since $\delta$ is entailed by $\btw{\Psi_{\Pi}}{i}{j}$, then no path of length $j-i+1$ in $\graph{\Pi}$ violates $\delta$.

\end{proof}

\begin{proof}[Proof]
Let $Y_i = X_0$, \ldots, and $Y_{n+i} = X_n$. 
We have that $(Y_i, \ldots, Y_{n+i})$ is a path in \graph{\Pi}. 
Since $\Pi$ is \cyclic{},
there is some edge $(Y_{i-1},Y_i)$ in \graph{\Pi}, 
and there is also some edge $(Y_{i-2},Y_{i-1})$ in \graph{\Pi}, 
and so on.
Hence, there is a path $(Y_0, \ldots, Y_{n+i})$ of length $n+i$ in \graph{\Pi}.
Similarly, since \graph{\Pi} is \cyclic{},
in \graph{\Pi}
there are edges $(Y_{n+i},Y_{n+i+1})$, $(Y_{n+i+1},Y_{n+i+2})$, and so on. 
Hence, there is a path $(Y_0, Y_{n+i+j})$ in \graph{\Pi} of length $n+i+j$
such that $(X_0,\ldots,X_n) = (Y_i,\ldots,Y_{n+i})$.%
\end{proof}

\begin{proof}[Proof]
    We prove the case where $\delta$ consist of normal atoms,
    the proof for the general case follows the same lines.
    
    Let $\Pi$ be defined over some set of atoms $\mathcal{A}$.
    Given that $\delta$ is a resolvent of $\btw{\Psi_{{\Pi}}}{i}{j}$,
    its atoms must belong to some smallest set $\btw{\mathcal{A}}{k}{l}$
    such that $0 \leq i-1 \leq k\leq l\leq j$.
    %
    Then, the integers $t$ such that 
    $\stepsimple{\shift{\delta}{t}} \subseteq [0,n]$ are exactly the $t$'s 
    such that $-k\leq t \leq n-l$.
    Hence, to prove this theorem we just have to prove that for every $t$
    such that $-k\leq t \leq n-l$ the set of nogoods
    $\btw{\Psi_{{\Pi}}}{1}{n}$ entails \shift{\delta}{t}.
    
    Since $\delta$ is a resolvent of $\btw{\Psi_{{\Pi}}}{i}{j}$,
    the shifted version \shift{\delta}{1-i} is a resolvent of $\btw{\Psi_{{\Pi}}}{1}{j+1-i}$, 
    and therefore $\btw{\Psi_{{\Pi}}}{1}{j+1-i}$ entails \shift{\delta}{1-i}.
    %
    %
    By Proposition~\ref{prop:nogoodpath}, 
    every path $( X_0, \ldots, X_{j+1-i} )$ in $\graph{\Pi}$
    does not violate $\shift{\delta}{1-i}$. 
    %
    We consider two cases: $k<l$ and $k=l$.

    \emph{Case 1 ($k<l$)}. 
    Since $\Pi$ is \cyclic, 
    we can prove by contradiction that 
    every 
    path $( Y_{0}, \ldots, Y_{l-k} )$ in $\graph{\Pi}$
    does not violate $\shift{\delta}{-k}$. 
     
    Assume that there is such a path
    $( Y_{0}, \ldots, Y_{l-k} )$.
    %
    By Proposition~\ref{prop:extendpath:program}, 
    there is a path $(X_0, \ldots, X_{j+1-i} )$ in $\graph{\Pi}$
    such that $( Y_{0}, \ldots, Y_{l-k} ) = (X_{k+1-i}, \ldots, X_{l+1-i})$. 
    Given that $( Y_{0}, \ldots, Y_{l-k} )$ violates $\shift{\delta}{-k}$,
    the path $(X_0, \ldots, X_{j+1-i} )$ would violate 
    $\shift{\delta}{-k+(k+1-i)} = \shift{\delta}{1-i}$, 
    which contradicts one of our previous statements.
    
    %
    %
    %
    If every path $( Y_{0}, \ldots, Y_{l-k} )$ in $\graph{\Pi}$
    does not violate $\shift{\delta}{-k}$, 
    then for every 
    path $( X_0, \ldots, X_n )$ in \graph{\Pi}
    and every $t$ such that $-k \leq t \leq n-l$,
    the shifted nogood \shift{\delta}{t} is not violated.
    %
    %
    Then, by Theorem~\ref{thm:problemgraph}, 
    we can conclude that 
    for every $t$ such that $-k \leq t \leq n-l$,
    the solutions to $\btw{\Psi_\Pi}{1}{n}$ 
    do not violate \shift{\delta}{t},
    and therefore $\btw{\Psi_\Pi}{1}{n}$ entails $\shift{\delta}{t}$.

    \emph{Case 2 ($k=l$)}. 
    Note that since all atoms of $\delta$ belong to $\at{\mathcal{A}}{k}$, 
    all atoms of $\shift{\delta}{-k}$ belong to \at{\mathcal{A}}{0}. 
    Given that $\Pi$ is \cyclic, 
    we can prove by contradiction that 
    every 
    path $( Y_{0}, Y_1 )$ in $\graph{\Pi}$
    does not violate $\shift{\delta}{-k}$. 
     
    Assume that there is such a path $( Y_{0}, Y_1 )$.
    Then, by Proposition~\ref{prop:extendpath:program}, 
    there is a path of the form 
    $(X_0, \ldots, X_{j+1-i}, X_{j+2-i} )$ in $\graph{\Pi}$
    such that $( Y_{0}, Y_1 ) = (X_{k+1-i}, X_{k+2-i})$.
    This path violates $\shift{\delta}{-k+(k+1-i)}=\shift{\delta}{1-i}$. 
    Since the atoms of $\shift{\delta}{1-i}$ belong to $\at{\mathcal{A}}{k+1-i}$
    and $k+1-i < j+2-i$, 
    the subpath 
    $(X_0, \ldots, X_{j+1-i})$ in $\graph{\Pi}$ also violates $\shift{\delta}{1-i}$, 
    which contradicts one of our previous statements.
    
    Given that every path $( Y_{0}, Y_1 )$ in $\graph{\Pi}$
    does not violate $\shift{\delta}{-k}$,
    it follows that for every 
    path $( X_0, \ldots, X_n, X_{n+1} )$ in \graph{\Pi}
    and every $t$ such that $-k \leq t \leq n-l$,
    the shifted nogood \shift{\delta}{t} is not violated.
    Since the atoms of $\shift{\delta}{t}$ belong to $\at{\mathcal{A}}{k+t}$
    and $k+t < n+1$ 
    for all $t$, then 
    the previous statement also holds for all paths 
    $( X_0, \ldots, X_n )$ in \graph{\Pi}.
    Finally, we can reason as in the previous case, 
    and by Theorem~\ref{thm:problemgraph} conclude that 
    for all $t$ such that $-k \leq t \leq n-l$, 
    the set of nogoods $\btw{\Psi_\Pi}{1}{n}$ entails $\shift{\delta}{t}$.
    \end{proof}

\begin{proof}[Proof]
Let $Y_i = X_0$, \ldots, and $Y_{n+i} = X_n$. 
We have that $(Y_i, \ldots, Y_{n+i})$ is a path in \graph{\Pi} 
where $Y_i$ is initial or reachable wrt $I$. 
If $Y_i$ is initial wrt $I$, 
since $\Pi$ is \cyclic{} wrt $I$,
$Y_i$ is also loop-reachable,
and therefore there is a path
$(Y_0,\ldots,Y_i)$ of length $i$ in $\graph{\Pi}$, 
that may go through a loop in $\graph{\Pi}$ as many times as necessary.
Similarly, 
if $Y_i$ is reachable wrt $I$,
since $\Pi$ is \cyclic{} wrt $I$, there is a path
$(Y_0,\ldots,Y_i)$ of length $i$ in $\graph{\Pi}$ 
that may go through a loop in $\graph{\Pi}$ 
and through some initial state wrt $I$. 
Both cases imply that there is a path $(Y_0, \ldots, Y_{n+i})$ of length $n+i$ in \graph{\Pi}.

On the other direction, 
in \graph{\Pi}
there are edges $(Y_{n+i},Y_{n+i+1})$, $(Y_{n+i+1},Y_{n+i+2})$, and so on.
These edges must exist because
the states occurring in them are reachable wrt $I$,
and therefore they must have some outgoing edge.
This gives us a path $(Y_0, \ldots,Y_{n+i+j})$ in \graph{\Pi} of length $n+i+j$
such that $(X_0,\ldots,X_n) = (Y_i,\ldots,Y_{n+i})$.%
\end{proof}

\begin{proof}[Proof]
\emph{From left to right.} 
Assume $\Pi$ is \cyclic.
To prove condition (i) of being \cyclic{} wrt $\emptyset$, 
take any initial state $X$ wrt $\emptyset$.
Since $\Pi$ is \cyclic, $X$ has some predecessor in \graph{\Pi}.
Similarly, each predecessor must have another predecessor, and so on.
Given that $\graph{\Pi}$ is finite, 
at some point one of these states must be repeated,
which implies that $X$ is loop-reachable and condition (i) holds. 
Condition (ii)
of being \cyclic{} wrt $\emptyset$ follows directly from the assumption. 

\emph{From right to left.} 
Assume that $\Pi$ is \cyclic{} wrt $\emptyset$.
The initial states wrt $\emptyset$ are the states of \graph{\Pi}
that have some outgoing edge. 
By the assumption, these states are loop-reachable. 
Hence, they 
have some incoming edge, 
which implies that they are internal.
The reachable states wrt $\emptyset$ are the states of $\graph{\Pi}$
that have some incoming edge.
Again, by the assumption, these states are also internal.
This shows that states with incoming or incoming edges are internal, 
and therefore $\Pi$ is \cyclic.
\end{proof}

\begin{proof}[Proof]
The proof is similar to the proof of Theorem~\ref{thm:cyclic:program}.
    We prove the case where $\delta$ consist of normal atoms,
    the proof for the general case follows the same lines.
    
    Let $\Pi$ be defined over some set of atoms $\mathcal{A}$.
    Given that $\delta$ is a resolvent of $\btw{\Psi_{{\Pi}}}{i}{j}$,
    its atoms must belong to some smallest set $\btw{\mathcal{A}}{k}{l}$
    such that $0 \leq i-1 \leq k\leq l\leq j$.
    Then, the integers $t$ such that 
    $\stepsimple{\shift{\delta}{t}} \subseteq [0,n]$ are exactly the $t$'s 
    such that $-k\leq t \leq n-l$.
    Hence, to prove this theorem we just have to prove that for every $t$
    such that $-k\leq t \leq n-l$ the set of nogoods
    $\btw{\Psi_{{\Pi}}}{1}{n}\cup \assignfornogood{I}$
    entails \shift{\delta}{t}.
    
    Since $\delta$ is a resolvent of $\btw{\Psi_{{\Pi}}}{i}{j}$,
    the shifted version \shift{\delta}{1-i} is a resolvent of $\btw{\Psi_{{\Pi}}}{1}{j+1-i}$, 
    and therefore $\btw{\Psi_{{\Pi}}}{1}{j+1-i}$ entails \shift{\delta}{1-i}.
    %
    %
    By Proposition~\ref{prop:nogoodpath}, 
    every path $( X_0, \ldots, X_{j+1-i} )$ in $\graph{\Pi}$
    does not violate $\shift{\delta}{1-i}$. 
    We consider two cases: $k<l$ and $k=l$.

    \emph{Case 1 ($k<l$)}. 
    Since $\Pi$ is \cyclic{} wrt $I$, 
    we can prove by contradiction that 
    every 
    path $( Y_{0}, \ldots, Y_{l-k} )$ in $\graph{\Pi}$
    where $Y_0$ is initial or reachable wrt $I$
    does not violate $\shift{\delta}{-k}$. 
     
    Assume that there is such a path
    $( Y_{0}, \ldots, Y_{l-k} )$.
    By Proposition~\ref{prop:extendpath:problem}, 
    there is a path of the form $(X_0, \ldots, X_{j+1-i} )$ in $\graph{\Pi}$
    such that $( Y_{0}, \ldots, Y_{l-k} ) = (X_{k+1-i}, \ldots, X_{l+1-i})$. 
    Given that $( Y_{0}, \ldots, Y_{l-k} )$ violates $\shift{\delta}{-k}$,
    the path $(X_0, \ldots, X_{j+1-i} )$ would violate 
    $\shift{\delta}{-k+(k+1-i)} = \shift{\delta}{1-i}$, 
    which contradicts one of our previous statements.
    
    If every path $( Y_{0}, \ldots, Y_{l-k} )$ in $\graph{\Pi}$
    where $Y_0$ is initial or reachable wrt $I$
    does not violate $\shift{\delta}{-k}$, 
    then for every 
    path $( X_0, \ldots, X_n )$ in \graph{\Pi}
    where $X_0$ is initial wrt $I$
    and every $t$ such that $-k \leq t \leq n-l$,
    the shifted nogood \shift{\delta}{t} is not violated.
    Theorem~\ref{thm:problemgraph} gives us a correspondence between 
    the paths $( X_0, \ldots, X_n )$ in \graph{\Pi}
    and the solutions to $\btw{\Psi_\Pi}{1}{n}$.
    It is easy to see that, if we add to 
    $\btw{\Psi_{{\Pi}}}{1}{n}$ the set $\assignfornogood{I}$, 
    we also have a correspondence between 
    the paths $( X_0, \ldots, X_n )$ in \graph{\Pi} where $X_0$ is initial wrt $I$ 
    and 
    the solutions to $\btw{\Psi_{{\Pi}}}{1}{n}\cup\assignfornogood{I}$.
    Hence, 
    we can conclude that 
    the solutions to $\btw{\Psi_{{\Pi}}}{1}{n}\cup\assignfornogood{I}$
    do not violate \shift{\delta}{t},
    and therefore $\btw{\Psi_\Pi}{1}{n}\cup\assignfornogood{I}$ entails $\shift{\delta}{t}$.

    \emph{Case 2 ($k=l$)}. 
    Note that since all atoms of $\delta$ belong to $\at{\mathcal{A}}{k}$, 
    all atoms of $\shift{\delta}{-k}$ belong to \at{\mathcal{A}}{0}. 
    Given that $\Pi$ is \cyclic{} wrt $I$, 
    we can prove by contradiction that 
    every 
    path $( Y_{0}, Y_1 )$ in $\graph{\Pi}$
    where $Y_0$ is initial or reachable wrt $I$
    does not violate $\shift{\delta}{-k}$. 
     
    Assume that there is such a path $( Y_{0}, Y_1 )$.
    Then, by Proposition~\ref{prop:extendpath:problem}, 
    there is a path of the form 
    $(X_0, \ldots, X_{j+1-i}, X_{j+2-i} )$ in $\graph{\Pi}$
    such that $( Y_{0}, Y_1 ) = (X_{k+1-i}, X_{k+2-i})$.
    This path violates $\shift{\delta}{-k+(k+1-i)}=\shift{\delta}{1-i}$. 
    Since the atoms of $\shift{\delta}{1-i}$ belong to $\at{\mathcal{A}}{k+1-i}$
    and $k+1-i < j+2-i$, 
    the subpath 
    $(X_0, \ldots, X_{j+1-i})$ in $\graph{\Pi}$ also violates $\shift{\delta}{1-i}$, 
    which contradicts one of our previous statements.
    
    Given that every path $( Y_{0}, Y_1 )$ in $\graph{\Pi}$
    where $Y_0$ is initial or reachable wrt $I$
    does not violate $\shift{\delta}{-k}$,
    it follows that for every 
    path $( X_0, \ldots, X_n, X_{n+1} )$ in \graph{\Pi}
    where $X_0$ is initial wrt $I$
    and every $t$ such that $-k \leq t \leq n-l$,
    the shifted nogood \shift{\delta}{t} is not violated.
    Since the atoms of $\shift{\delta}{t}$ belong to $\at{\mathcal{A}}{k+t}$
    and $k+t < n+1$ 
    for all $t$, then 
    the previous statement also holds for all paths 
    $( X_0, \ldots, X_n )$ in \graph{\Pi} where $X_0$ is initial wrt $I$.
    Finally, we can reason as in the previous case, 
    and by Theorem~\ref{thm:problemgraph} conclude that 
    for all $t$ such that $-k \leq t \leq n-l$, 
    the set of nogoods $\btw{\Psi_\Pi}{1}{n}\cup\assignfornogood{I}$ entails $\shift{\delta}{t}$.
    \end{proof}



\begin{proof}[Proof]
The proof follows the lines of the explanation of the main text, 
using the Splitting Set Theorem to split the program into 
the choice rule over $\trbsymbol$ and the rest of the program.
\end{proof}

\begin{proof}[Proof]

For any solution S of $\tpb{\Pi}{I}{F}$, by Theorem \ref{thm:problemgraph}, 
there is a path $P = (X_0, X_1, \ldots , X_n)$ in $\graph{\Pi} = (V,E)$
where $I \subset X_0$ and $F \subset X_n$.
By Proposition \ref{prop:labmda:graph}, if $(X,Y) \in E$ then $( X, Y \cup \{\lambda\} ) \in E^{\lambda}$
and $(X\cup\{\lambda\},Y\cup\{\lambda\}) \in E^{\lambda}$
where $\graph{\trbdo{\Pi}} = (V^{\lambda}, E^{\lambda})$.
So, we can transform $P$ to a unique $P^{\lambda} = ( X_0, X_1\cup\{\lambda\}, \ldots, X_n\cup\{\lambda\} )$
which is a path of $\graph{\trbdo{\Pi}}$.
By Theorem \ref{thm:problemgraph}, $P^{\lambda}$ is a solution of $\tpb{\trbdo{\Pi}}{I}{F}$.
Additionally, since $\lambda$ is true on all states but the first one, it is also a \trbnormal{} solution.

On the other hand, for a \trbnormal{} solution $S^{\lambda}$ of $\tpb{\trbdo{\Pi}}{I}{F}$,
we have a path $P^{\lambda} = ( Y_0, Y_1\cup \{\lambda\}, \ldots, Y_n\cup \{\lambda\} )$ (Theorem \ref{thm:problemgraph}).
Since a state $W\cup\{\lambda\} \in V^{\lambda}$ must satisfy the rules of $\Pi$
then $W \in V$.
Hence, $Y_1,\ldots,Y_n \in V$ and $(Y_{m-1}, Y_m) \in E$ for $2 \leq m \leq n$.
Since the pairs of the form $( X, Y \cup \{\lambda\} )$ only exist for $(X,Y) \in E$,
then $(Y_0,Y_1) \in E$. Therefore, by Theorem \ref{thm:problemgraph} $P = ( Y_0, Y_1, \ldots, Y_n )$ is a path in $\graph{\Pi}$.
Since $I \in Y_0$ and $F \in Y_n$, then there is a solution that corresponds to this path
and is a solution $\tpb{\Pi}{I}{F}$.

\end{proof} 

\begin{proof}[Proof]

Recall the condition for a temporal logic problem to be \cyclic{}:
1) Every initial state wrt $I$ is loop-reachable,
and 2) Every reachable state wrt $I$ is internal.

In this case $I = \{\Flit{\atomat{\trbsymbol}}{0}\}$.
The first condition is satisfied since the initial state has no $\lambda$. 
By Proposition \ref{prop:labmda:graph},
a state without lambda is connected to itself,
meaning that it is a loop.

For the second condition we consider that all possible states
are connected to some next state without $\lambda$ (Proposition \ref{prop:labmda:graph}),
hence all reachable states have outgoing edges.
Additionally, every state with or without $\lambda$ has some previous state connected to it (Proposition \ref{prop:labmda:graph}).

We can then say that the program $\trbdo{\Pi}$ is \cyclic{} wrt $\{\Flit{\atomat{\trbsymbol}}{0}\}$.
\end{proof}

\begin{proof}[Proof]

Proposition \ref{prop:cyclic:translation} tells us that $\trbdo{\Pi}$ is \cyclic{}.
We can also think of $\{\Flit{\atomat{\trbsymbol}{0}}\}$ as a partial assignment $I$.
Hence we can directly apply Theorem \ref{theorem:cyclic:problem} to get that
$\btw{\Psi_\trbdo{\Pi}}{1}{n} \cup \big\{\{\Tlit{\atomat{\trbsymbol}{0}}\}\big\} \models \delta$.

\end{proof} 

\begin{proof}[Proof]

For any $\delta$ and $t$ such that $\stepsimple{\shift{\delta}{t}} \subseteq [0,n], \lambda_0 \notin \shift{\delta}{t}$
it holds that $\btw{\Psi_\trbdo{\Pi}}{1}{n} \cup \{ \{ \atomat{\Tlit{\trbsymbol{}}}{0} \} \} \models \shift{\delta}{t}$ (Theorem \ref{thm:translation:generalization}).

This means that every path of length $k=j-i$ in $\graph{\trbdo{\Pi}}$ does not violate $\shift{\delta}{t}$ where $\lambda_0 \notin \shift{\delta}{t}$.
Since every path longer than $k$ has a subpath of length $k$, 
then every path longer than $k$ must also not violate $\shift{\delta}{t}$.
Therefore, for $n \geq k$, every path $( X_0, \ldots, X_n )$ in $\graph{\trbdo{\Pi}}$ must not violate $\shift{\delta}{t}$ where $\lambda_0 \notin \shift{\delta}{t}$.

Now we will prove the theorem by contradiction.
Take any path $ ( X_0, \ldots, X_n )$ in $\graph{\Pi}$ and assume that it violates $\simp{\shift{\delta}{t}}$.
By Proposition \ref{prop:labmda:graph} we have a path $( X_0, X_1 \cup \{\lambda\}\ldots, X_n \cup \{\lambda\})$ that will violate $\simp{\shift{\delta}{t}} \cup \{ \atomat{\trbsymbol{}}{1}, \ldots, \atomat{\trbsymbol{}}{n} \}$.
Since $\shift{\delta}{t} \subset \simp{\shift{\delta}{t}} \cup \{ \atomat{\trbsymbol{}}{1}, \ldots, \atomat{\trbsymbol{}}{n} \}$ the path also violates $\shift{\delta}{t}$.
This is a contradiction.
Hence, every path of length $n$ of $\graph{\Pi}$ does not violate $\simp{\shift{\delta}{t}}$.
Therefore $\btw{\Psi_\Pi}{1}{n} \models \{ \simp{\shift{\delta}{t}} \mid \stepsimple{\shift{\delta}{t}} \subseteq [0,n], \Tlit{\atomat{\trbsymbol{}}{0}} \notin \shift{\delta}{t} \}$

\end{proof} 


\begin{proof}[Proof]

For any model $I$ of $\mathcal{T}_1$ of lengh $n$ there is also a model $I^\trbsymbol = I \cup \{ \trbsymbol_1,...,\trbsymbol_n \}$ of $\mathcal{T}_2$.
Since $\trb$ only adds a $\trbsymbol$ to the dynamic constraints the only difference in the nogoods of $\mathcal{T}_1$ 
and $\mathcal{T}_2$ is that $\integrprev{\trbdoprev{\Pi}}$ have an additional $\trbsymbol$.
Hence, no nogoods of $\trbdoprev{\Pi} \setminus \integrprev{\trbdoprev{\Pi}}$ is satisfied by $I^\trbsymbol$.
Additionally, since in \trbnormal{} solutions all $\trbsymbol$ are true, the nogoods of $\integrprev{\Pi}$ can be simplified by deleting their $\trbsymbol$. 
The simplified nogoods of $\integrprev{\trbdoprev{\Pi}}$ are the same as the nogoods of $\integrprev{\Pi}$.
This means that $I^\trbsymbol$ does not satisfy any nogood in $\integrprev{\trbdoprev{\Pi}}$.
We can then conclude that $I^\trbsymbol$ does not satisfy any nogood of $\trbdoprev{\Pi}$ and is thus a model of $\mathcal{T}_2$.

For any model $I^\trbsymbol$ of $\mathcal{T}_2$ of lengh $n$
there is also a model $ I = I^\trbsymbol \setminus \{ \trbsymbol_1,...,\trbsymbol_n \}$ of $\mathcal{T}_1$.
Since $\trb$ only adds a $\trbsymbol$ to the dynamic constraints the only difference in the nogoods of $\mathcal{T}_1$ 
and $\mathcal{T}_2$ is that $\integrprev{\trbdoprev{\Pi}}$ have an additional $\trbsymbol$.
Hence, all nogoods of $\Pi \setminus \integrprev{\Pi}$ are not satisfied by $I$.
Since in \trbnormal{} solutions all $\trbsymbol$ are true, the nogoods of $\integrprev{\trbdoprev{\Pi}}$ act the same way as the nogoods of $\integrprev{\Pi}$.
Hence, the nogoods of $\integrprev{\Pi}$ are also not satisfied by $I$.
We can then conclude that $I$ does not satisfy any nogood of $\Pi$ and is thus a model of $\mathcal{T}_1$.

It follows that for every stable model of $\tpb{\Pi}{I}{F}$ there is a corresponding stable model of $\tpb{\trbdoprev{\Pi}}{I}{F}$ and vice versa

\end{proof}
\begin{lemma}\label{lemma:step_f_overapprox}
	An interval $[k,l]$ is an \textit{overaproximation} of the interval $[i,j]$ if $k\leq i$ and $j\leq l$ holds. For some resolvent $\delta$ of $\btw{\Psi_\trbdoprev{\Pi}}{i}{j}$, $\step{\delta}$ computes an \textit{overaproximation} of the interval $[i,j]$.
\end{lemma}

\begin{proof}[Proof]

\begin{itemize}
    \item \textbf{case 1}: $\delta$ is a resolvent of $\btw{\Psi_\trbdoprev{\Pi}}{i}{i}$.
    By definition, 
    $\step{\delta} = [i,i]$ since there would be no $\trbsymbol$ in $\delta$ and the only timestep in the atoms of $\delta$ would be $i$.

    \item \textbf{case 2}: $\delta$ is a resolvent of $\btw{\Psi_\trbdoprev{\Pi}}{i}{j}$ with $i<j$ and $\trbsymbol[i,j] \in \delta$.
    By definition,
    $\step{\delta} = [i-1,j]$ since the lowest timepoint in any $\trbsymbol$ is $i$.

    \item \textbf{case 3}: $\delta$ is a resolvent of $\btw{\Psi_\trbdoprev{\Pi}}{i}{j}$ with $i<j$ and $\trbsymbol[i+1,j] \in \delta$.
    By definition,
    $\step{\delta} = [i,j]$ since the lowest timepoint in any $\trbsymbol$ is $i+1$.
\end{itemize}

We can clearly see that for any resolvent $\delta$ the function $\step{\delta}$ computes the exact (cases 1 and 3) or a bigger (case 2) interval.
Hence, it is an overapproximation of the interval.

\end{proof}

\begin{proof}[Proof]

If $\delta$ is a resolvent of $\btw{\Psi_{\trbdoprev{\Pi}}}{1}{m}$ and $\step{\delta} = [i,j]$ where $0 \leq i \leq j \leq m$
then $\delta$ is a resolvent of $\btw{\Psi_{\trbdoprev{\Pi}}}{i}{j}$ (by Lemma~\ref{lemma:step_f_overapprox}).
For any $t$, $\step{\shift{\delta}{t}} = [i+t,j+t]$ which means $\step{\shift{\delta}{t}}$ is a resolvent of $\btw{\Psi_{\trbdoprev{\Pi}}}{i+t}{j+t}$ (by Lemma~\ref{lemma:shift-resolution}).
Consequently, for any $t$ where $[i+t,j+t] \subseteq [1,n]$ then $\shift{\delta}{t}$ is entailed by $\btw{\Psi_{\trbdoprev{\Pi}}}{1}{n}$ due to Theorem~\ref{thm:learning:basic}.

\end{proof}

\begin{proof}[Proof]
Let $\mathcal{A}$ be a set of atoms ocurring in a logic program $\Pi$ and $\prev{\mathcal{A}}$ be the set of atoms that reference the past.
Recall that a logic program $\Pi$ is in PNF if for any rule $r \in \normal{\Pi} \cup \choice{\Pi}$ it holds that $\body{r} \cap \prev{\mathcal{A}} = \emptyset$

For any rule $r \in \normal{\Pi} \cup \choice{\Pi}$ where $\body{r} \cap \prev{\mathcal{A}} \neq \emptyset$
it holds that $\body{\traatom{r}} \cap \prev{\mathcal{A}} = \emptyset$ since any occurrence is substituted by the corresponding $\traatom{p}$ atom.
For any rule $r \in \normal{\Pi} \cup \choice{\Pi}$ where $\body{r} \cap \prev{\mathcal{A}} = \emptyset$ 
it holds that $\body{\traatom{r}} \cap \prev{\mathcal{A}} = \emptyset$ since the translation does not change the rule.
Hence, for any rule $r \in \normal{\Pi} \cup \choice{\Pi}$ it holds that $\body{\traatom{r}} \cap \prev{\mathcal{A}} = \emptyset$.
Which means that $\trado{\Pi}$ is in PNF.

\end{proof} 

\begin{lemma}\label{lemma:pstar-eq-pprime}
    For any program $\Pi$ the truth value of $\traatom{p}$ and $\prev{p}$ always conincide in the solutions of $\trado{\Pi}$
    where $\traatom{a} \in \traatom{\mathcal{A}}$ are the atoms introduced by the \tra translation
    and $\prev{a} \in \prev{\mathcal{A}}$ are the atoms occuring in $\Pi$ referencing the past.
\end{lemma}

\begin{proof}[Proof]

We label the rules added by the translation \tra as follows:
\begin{align}
\{ \traatom{a} \} \leftarrow         \label{tra:r1} \\
\bot \leftarrow a', \neg \traatom{a} \label{tra:r2} \\
\bot \leftarrow \neg a', \traatom{a} \label{tra:r3}
\end{align}

\begin{itemize}
    \item if $\prev{a}$ is True then $\traatom{a}$ must also be True to not violate rule \ref{tra:r2}
    \item if $\prev{a}$ is False then $\traatom{a}$ must also be False to not violate rule \ref{tra:r3}
    \item if $\traatom{a}$ is True then $\prev{a}$ must also be True to not violate rule \ref{tra:r3} 
    \item if $\traatom{a}$ is False then $\prev{a}$ must also be False to not violate rule \ref{tra:r2} 
\end{itemize}

We can then conclude that the truth value of $\traatom{a}$ and $\prev{a}$ always conincide in the resulting program $\trado{\Pi}$.

\end{proof}

\begin{proof}[Proof]

We label the rules added by the translation \tra  as follows:
\begin{align}
\{ \traatom{a} \} \leftarrow         \label{tra2:r1} \\
\bot \leftarrow a', \neg \traatom{a} \label{tra2:r2} \\
\bot \leftarrow \neg a', \traatom{a} \label{tra2:r3}
\end{align}

Let $\traatom{a} \in \traatom{\mathcal{A}}$ be the atoms added by the $\tra$ translation
and $\prev{a} \in \prev{\mathcal{A}}$ be the set of atom occuring in $\Pi$ that reference the past.

\textbf{Case 1: } Let $\prevset{\mathcal{A}} \cap \atom{\Pi} = \emptyset$.
Since $\trado{\Pi} = \Pi$ then $\mathcal{T}_1 = \mathcal{T}_2$ and they have the same solutions.

\textbf{Case 2: } Let $\prevset{\mathcal{A}} \cap \atom{\Pi} \neq \emptyset$.
For any solution $\mathcal{S}_1$ of $\mathcal{T}_1$
there is a solution $\mathcal{S}_2$ of $\mathcal{T}_2$
where $\mathcal{S}_2 = \mathcal{S}_1 \cup \{ \traatom{a} | \prev{a} \in \mathcal{S}_1 \}$. 
Since $\prev{a}$ and $\traatom{a}$ always have the same truth value (Lemma \ref{lemma:pstar-eq-pprime}),
the evaluation of the nogoods induced by $\trado{\Pi}$ where $\prev{a}$ was substituted by $\traatom{a}$ will stay the same regardless of the assignment.
Also, none of the nogoods induced by the extra rules \eqref{tra2:r2} and \eqref{tra2:r3} will be satisfied since $\prev{a}$ and $\neg \traatom{a}$ always have different truth values.
We can also ignore rule \eqref{tra2:r1} since it does not induce any nogoods.
Hence, $\mathcal{S}_2$ is a stable model of $\trado{\Pi}$.
Finally, given that $\mathcal{S}_1 $ is consistent with $I$ and $F$
then $\mathcal{S}_2$ is also consistent with $I$ and $F$.
Consequently, it is also a solution to $\mathcal{T}_2$.

On the other hand, 
For any solution $\mathcal{S}_2$ of $\mathcal{T}_2$
there is a solution $\mathcal{S}_1$ of $\mathcal{T}_1$
where $\mathcal{S}_1 = \mathcal{S}_2 \setminus \{ \traatom{a} | \prev{a} \in \mathcal{S}_2 \}$.
Since $\prev{a}$ and $\traatom{a}$ always have the same truth value (Lemma \ref{lemma:pstar-eq-pprime}),
the evaluation of the nogoods induced by $\Pi$ will stay the same regardless of the assignment.
Hence, $\mathcal{S}_2$ is a stable model of $\Pi$.
Finally, given that $\mathcal{S}_2$ is consistent with $I$ and $F$,
then $\mathcal{S}_1$ is also consistent with $I$ and $F$.
Consequently, it is also a solution to $\mathcal{T}_1$.

\end{proof}

\begin{lemma} \label{lemma:simp-transform}
For a resolvent $\delta$ of $\btw{\Psi_{\trbdoprev{\trado{\Pi}}}}{1}{m}$ it holds that $\simp{\delta}$ is entailed by $\btw{\Psi_{\Pi}}{1}{m}$ for \trbnormal{} solutions of $\btw{\Psi_{\trbdoprev{\trado{\Pi}}}}{1}{m}$.
\end{lemma}

\begin{proof}[Proof]
Let $\delta^\trbsymbol$ be a resolvent of $\btw{\Psi_{\trbdoprev{\trado{\Pi}}}}{1}{m}$.
In \trbnormal{} solutions the $\trbsymbol$ atoms in the nogoods are always true
and thus have no effect in their satisfaction.
Hence, the nogood $\traatom{\delta} = \delta^\trbsymbol \setminus \trbsymbol[1,m]$
is entailed by $\btw{\Psi_{\trado{\Pi}}}{1}{m}$
since the nogoods in $\btw{\Psi_{\trado{\Pi}}}{1}{m}$ are the nogoods in $\btw{\Psi_{\trbdoprev{\trado{\Pi}}}}{1}{m}$ without $\trbsymbol$s.

Let $\traatom{\delta}$ be a resolvent of $\btw{\Psi_{\trado{\Pi}}}{1}{m}$
with its corresponding resolution proof $\mathcal{T}$.
Let $C$ be the constraints added by the $\tra$ translation.
Observe that a nogood of $\trado{\Pi} \setminus C$ can be transformed into a nogood of $\Pi$
simply by substituting all $\at{\traatom{a}}{i} \in \traatom{\delta}$
by their corresponding atom $\at{a}{i-1}$ from $\mathcal{A}$
where $\traatom{a} \in \traatom{\mathcal{A}}$ are the atoms introduced by the translation
and $\mathcal{A}$ is the set of atoms ocurring in $\Pi$.

For any nogood in $\mathcal{T}$ containing atoms $\at{\traatom{a}}{i}$,
we can substitute them by the corresponding atom $\at{a}{i-1}$ without changing the semantics of the nogoods
since they always have the same truth value (by Lemma \ref{lemma:pstar-eq-pprime}).

Next, recall that the constraints added by the $\tra$ translation have the form
\begin{align*}
\bot \leftarrow& \at{a}{i-1}, \neg \at{\traatom{a}}{i}\\
or \\
\bot \leftarrow& \neg \at{a}{i-1}, \at{\traatom{a}}{i}\\
\end{align*} 
for some integer $i$.
If we substitute $\traatom{a}$ by the corresponding atom we get the constraints
\begin{align*}
\bot \leftarrow& \at{a}{i-1}, \neg \at{a}{i-1}\\
 or \\
\bot \leftarrow& \neg \at{a}{i-1}, \at{a}{i-1}\\
\end{align*} 
It is easy to see that any nogood that resolves with the nogoods induced by these constraints
would result in the same nogood.
Hence, we can remove the nogoods induced by $C$ from $\mathcal{T}$ without affecting its result.
Note that the choice rules introduced by the translatation do not induce nogoods.
This means that any nogood left in $\mathcal{T}$ is either in or entailed by $\btw{\Psi_{\Pi}}{1}{m}$.
Hence, the result $\delta$ of the resolution proof $\mathcal{T}$ is entailed by $\btw{\Psi_{\Pi}}{1}{m}$.

It is clear that $\delta = \{\Vlit{\atomat{a}{i}} | \Vlit{\atomat{a}{i}} \in \delta^\trbsymbol, a \in \mathcal{A}\} \cup \{\Vlit{\atomat{a}{{i-1}}} | \Vlit{\traatom{\atomat{a}{i}}} \in \delta^\trbsymbol, \traatom{a} \in \traatom{\mathcal{A}} \}$.
In words, $\delta$ is the result of substitung any atom in $\traatom{\mathcal{A}}$ with the corresponding atom in $\mathcal{A}$ and ignoring any $\trbsymbol$ atoms.
Hence, $\delta = \simp{\delta^\trbsymbol}$.
Consequently, $\simp{\delta^\trbsymbol}$ is a resolvent of $\btw{\Psi_{\Pi}}{1}{m}$.
\end{proof}

\begin{proof}[Proof]

Since $\delta$ is entailed by $\btw{\Psi_{{\trbdoprev{\trado{\Pi}}}}}{1}{m}$ then $\simp{\delta}$ is also entailed by $\btw{\Psi_{\Pi}}{1}{m}$ (Lemma~\ref{lemma:simp-transform}).
By Theorem~\ref{thm:learning:zero} $\shift{\simp{\delta}}{t}$ is entailed by $\btw{\Psi_{\Pi}}{1}{n}$ for any $t$ where $\step{\shift{\delta}{t}} \subseteq [1,n]$.

\end{proof}

\section{Additional results}\label{sec:appendix:tables}
The following tables show the results 
of the experiments 
using the translations from our conference paper~(\ref{sec:translations}).
The experiments of our conference paper~\cite{roscst22a} have a bug 
in the multi-shot case. 
Here, that bug is fixed and 
the learning approach is no longer worse than 
the baseline, but it is still not better.
\begin{table}[h!]\footnotesize
    \begin{center}
    \newcommand{\fullline}[5]{\domname#1&\hspace{5pt}\tablevalue#2 &\hspace{5pt}\tablevalue#3 &\hspace{5pt}\tablevalue#4 &\hspace{5pt}\tablevalue#5 \\}
    \newcommand{\domname}[2]{\textbf{#1} & (#2)}
    \newcommand{\tablevalue}[2]{#1\ (#2)}
    
    \( \def\arraystretch{0.2}
    \begin{array}{lr r r r r }
        \multicolumn{2}{c}{}{} &  \qquad\qquad\textbf{baseline} &  \qquad\qquad\textbf{500}  &   \qquad\qquad\textbf{1000}  &   \qquad\qquad\textbf{1500} \\
        \midrule
        \fullline {{blocks} {300}}  {{0.5}{0}}            {{\textbf{0.1}}{0}}            {{\textbf{0.1}}{0}}            {{\textbf{0.1}}{0}}            \\
        \fullline {{depots} {270}}  {{146.4}{30}}         {{138.2}{29}}         {{\textbf{126.0}}{25}}         {{128.3}{30}}         \\
        \fullline {{driverlog} {135}}  {{14.1}{1}}           {{12.5}{1}}           {{12.3}{1}}           {{\textbf{10.7}}{1}}           \\
        \fullline {{elevator} {300}} {{\textbf{3.0}}{0}}            {{3.7}{0}}            {{3.8}{0}}            {{4.3}{0}}             \\
        \fullline {{grid} {30}}  {{11.4}{0}}           {{\textbf{5.2}}{0}}            {{5.3}{0}}            {{\textbf{5.2}}{0}}           \\
        \fullline {{gripper} {255}}  {{381.0}{96}}         {{368.5}{91}}         {{\textbf{359.0}}{90}}         {{370.8}{88}}         \\
        \fullline {{logistics} {225}}  {{\textbf{0.5}}{0}}            {{0.9}{0}}            {{0.9}{0}}            {{0.9}{0}}            \\
        \fullline {{mystery} {126}}  {{57.0}{3}}           {{58.5}{3}}           {{50.9}{3}}           {{\textbf{46.8}}{2}}           \\
        \midrule
        \fullline {{Total} {1663}}      {{89.7}{130}}         {{86.4}{124}}         {{\textbf{82.4}}{119}}         {{84.2}{121}}         \\
    \end{array}
    \)
    \caption{Single shot solving of PDDL benchmarks using translations.}
\label{tab:1shot-pddl-trans}
    \end{center}
\end{table}

\begin{table}[h!]\footnotesize
    \begin{center}
    \newcommand{\fullline}[5]{\domname#1&\hspace{5pt}\tablevalue#2 &\hspace{5pt}\tablevalue#3 &\hspace{5pt}\tablevalue#4 &\hspace{5pt}\tablevalue#5 \\}
    \newcommand{\domname}[2]{\textbf{#1} & (#2)}
    \newcommand{\tablevalue}[2]{#1\ (#2)}
    
    \( \def\arraystretch{0.2}
    \begin{array}{lr r r r r }
        \multicolumn{2}{c}{}{} &  \qquad\qquad\textbf{baseline} &  \qquad\qquad\textbf{500}  &   \qquad\qquad\textbf{1000}  &   \qquad\qquad\textbf{1500} \\
        \midrule
        \fullline {{HanoiTower} {20}}   {{160.6}{2}}          {{\textbf{97.7}}{0}}           {{101.0}{0}}          {{118.2}{1}}           \\
        \fullline {{Labyrinth} {20}}    {{\textbf{247.3}}{3}}          {{355.7}{4}}          {{355.7}{4}}          {{356.1}{4}}          \\
        \fullline {{Nomistery} {20}}    {{585.3}{12}}         {{575.6}{12}}         {{556.2}{12}}         {{\textbf{502.0}}{10}}         \\
        \fullline {{Ricochet Robots} {20}}  {{465.3}{9}}          {{\textbf{464.7}}{9}}          {{464.8}{8}}          {{\textbf{464.7}}{8}}          \\
        \fullline {{Sokoban} {20}}      {{458.8}{9}}          {{\textbf{441.5}}{9}}          {{458.8}{8}}          {{453.0}{8}}          \\
        \fullline {{Visit-all} {20}}    {{559.0}{12}}         {{556.5}{12}}         {{560.8}{12}}         {{\textbf{556.4}}{12}}         \\
        \midrule
        \fullline {{Total} {120}}       {{412.7}{47}}         {{415.3}{46}}         {{416.2}{44}}         {{\textbf{408.4}}{43}}         \\
    \end{array}
    \)
    \caption{Single shot solving of ASP benchmarks using translations.}
\label{tab:1shot-asp-trans}
    \end{center}
\end{table}

\begin{table}[h!]\footnotesize
    \begin{center}
    \newcommand{\fullline}[5]{\domname#1&\hspace{5pt}\tablevalue#2 &\hspace{5pt}\tablevalue#3 &\hspace{5pt}\tablevalue#4 &\hspace{5pt}\tablevalue#5 \\}
    \newcommand{\domname}[2]{\textbf{#1} & (#2)}
    \newcommand{\tablevalue}[2]{#1\ (#2)}
    
    \( \def\arraystretch{0.2}
    \begin{array}{lr r r r r }
        \multicolumn{2}{c}{}{} &  \qquad\qquad\textbf{baseline} &  \qquad\qquad\textbf{500}  &   \qquad\qquad\textbf{1000}  &   \qquad\qquad\textbf{1500} \\
        \midrule
        \fullline {{blocks} {20}}    {{1.3}{0}}            {{\textbf{0.7}}{0}}            {{\textbf{0.7}}{0}}            {{\textbf{0.7}}{0}}            \\
        \fullline {{depots} {18}}    {{\textbf{148.6}}{2}}          {{257.0}{3}}          {{189.4}{3}}          {{221.7}{3}}          \\
        \fullline {{driverlog} {9}}  {{108.9}{1}}          {{\textbf{102.0}}{1}}          {{104.9}{1}}          {{108.5}{1}}          \\
        \fullline {{elevator} {20}}  {{\textbf{280.3}}{5}}          {{285.7}{5}}          {{295.0}{5}}          {{305.4}{5}}          \\
        \fullline {{freecell} {16}}  {{900.0}{16}}         {{900.0}{16}}         {{900.0}{16}}         {{900.0}{16}}         \\
        \fullline {{grid} {2}}       {{5.2}{0}}            {{\textbf{4.1}}{0}}            {{4.2}{0}}            {{4.3}{0}}            \\
        \fullline {{gripper} {17}}   {{848.6}{16}}         {{\textbf{847.5}}{16}}         {{849.1}{16}}         {{847.9}{16}}         \\
        \fullline {{logistics} {20}} {{\textbf{225.2}}{5}}          {{225.3}{5}}          {{225.3}{5}}          {{225.3}{5}}          \\
        \fullline {{mystery} {14}}   {{\textbf{321.8}}{5}}          {{\textbf{321.8}}{5}}          {{321.9}{5}}          {{321.9}{5}}          \\
        \midrule
        \fullline {{Total} {136}}       {{\textbf{346.6}}{50}}         {{361.0}{51}}         {{353.8}{51}}         {{359.7}{51}}        \\
    \end{array}
    \)
    \caption{Multi shot solving of PDDL benchmarks using translations.}
    \label{tab:mshot-pddl-trans}
    \end{center}
\end{table}

\begin{table}[h!]\footnotesize
    \begin{center}
    \newcommand{\fullline}[5]{\domname#1&\hspace{5pt}\tablevalue#2 &\hspace{5pt}\tablevalue#3 &\hspace{5pt}\tablevalue#4 &\hspace{5pt}\tablevalue#5 \\}
    \newcommand{\domname}[2]{\textbf{#1} & (#2)}
    \newcommand{\tablevalue}[2]{#1\ (#2)}
    
    \( \def\arraystretch{0.2}
    \begin{array}{lr r r r r }
        \multicolumn{2}{c}{}{} &  \qquad\qquad\textbf{baseline} &  \qquad\qquad\textbf{500}  &   \qquad\qquad\textbf{1000}  &   \qquad\qquad\textbf{1500} \\
        \midrule
        \fullline {{HanoiTower} {20}}   {{\textbf{554.1}}{10}}         {{601.4}{11}}         {{593.7}{10}}         {{646.7}{11}}         \\
        \fullline {{Labyrinth} {20}}    {{\textbf{647.7}}{14}}         {{647.8}{14}}         {{647.8}{14}}         {{647.9}{14}}         \\
        \fullline {{Nomistery} {20}}    {{\textbf{64.2}}{1}}           {{77.0}{1}}           {{81.0}{1}}           {{69.3}{1}}         \\
        \fullline {{Ricochet Robots} {20}}  {{527.3}{11}}         {{\textbf{518.1}}{11}}         {{519.3}{11}}         {{521.3}{11}}         \\
        \fullline {{Sokoban} {20}}      {{\textbf{721.5}}{16}}         {{722.6}{16}}         {{722.3}{16}}         {{722.0}{16}}         \\
        \fullline {{Visit-all} {20}}    {{\textbf{677.5}}{13}}         {{704.0}{13}}         {{774.6}{15}}         {{801.6}{16}}         \\
        \midrule
        \fullline {{Total} {120}}       {{\textbf{532.1}}{65}}         {{545.2}{66}}         {{556.5}{67}}         {{568.1}{69}}         \\
    \end{array}
    \)
    \caption{Multi shot solving of ASP benchmarks using translations.}
    \label{tab:mshot-asp-degm1-trans}
    \end{center}
\end{table}

\end{document}